%% file: P4.tex
\documentclass[10pt]{article}
\usepackage[preprint]{tmlr}
\usepackage{hyperref}
\usepackage{url}
\usepackage{xurl}
\usepackage{booktabs}
\usepackage{longtable}
\usepackage{array}

\usepackage{amsmath,amssymb}
\usepackage{graphicx}
\usepackage{xcolor}
\usepackage{microtype}
\usepackage{seqsplit}
\usepackage{etoolbox}

\usepackage{calc}
\newcommand{\cv}[1]{\makebox[\widthof{+00.0}][r]{#1}}
\newcommand{\cb}[2]{[\cv{#1}, \cv{#2}]}
\newcommand{\ci}[3]{\cv{#1}~\cb{#2}{#3}}
\def\stripplus+#1\relax{#1}
\newcommand{\unsigned}[1]{\expandafter\stripplus#1\relax}
\newcommand{\LLMLtwo}{\mbox{LLMLingua-2}}
\AtBeginEnvironment{table}{\setlength{\abovecaptionskip}{0pt}\setlength{\belowcaptionskip}{6pt}}
\AtBeginEnvironment{table*}{\setlength{\abovecaptionskip}{0pt}\setlength{\belowcaptionskip}{6pt}}

\input{generated/numbers3}

\input{generated/restore}

\input{generated/p4_extra}

\input{generated/compress_fever}

\input{generated/p4_overlap}

\input{generated/compress_third}

\input{generated/compress_third_tfidf}
\input{generated/compress_fever_tfidf}
\input{generated/p4_retention}

\input{generated/p4_retention_labels}

\input{generated/locomo}

\input{generated/locomo_hard}
\input{generated/p4_audit}
\input{generated/p4new}

\input{generated/p4fu}

\input{generated/p4_sensitivity}
\input{generated/p4_refusal}

\input{generated/p4_granularity}

\input{generated/p4_known}

\input{generated/p4_cost}

\input{generated/p4_reader}

\input{generated/p4_tau}

\input{generated/p4_hybrid}

\input{generated/p4_whywords}

\input{generated/p4_fever_support}

\input{generated/p4_summary}

\hypersetup{hidelinks,pdfauthor={Junwei Yao, Yongxi Zhou, Zike Yang, Zhijun Wang, Jianan Liu, Yanxian Chen, Wenbo Ye},pdftitle={Evaluating Context Compaction on the Questions That Come Later}}
\title{Evaluating Context Compaction on the Questions That Come Later}
\author{\name Junwei Yao \email yao.junw@northeastern.edu \\ \addr Independent Researcher
\AND
\name Yongxi Zhou \email zhou.yongx@northeastern.edu \\ \addr Independent Researcher
\AND
\name Zike Yang \email zikeyang@alumni.usc.edu \\ \addr Independent Researcher
\AND
\name Zhijun Wang \email g674901087@gmail.com \\ \addr Independent Researcher
\AND
\name Jianan Liu \email jiananliu2408@gmail.com \\ \addr Independent Researcher
\AND
\name Yanxian Chen \email chen.yanx@northeastern.edu \\ \addr Independent Researcher
\AND
\name Wenbo Ye \email ywb6113@gmail.com \\ \addr Independent Researcher}

\begin{document}
\maketitle
{\let\thefootnote\relax\footnotetext{AI assistance: The authors used Claude Code (Anthropic) during the preparation of this work.}}

\begin{abstract}
Language-model agents compact their context for the question at hand, but the compacted context must also serve
questions that come later. We evaluate compaction on these later questions, using four corpora that annotate the
evidence of several questions per text and recorded agent trajectories, and compare every method with random
deletion at the same budget. On MultiRC, keeping the sentences that \texttt{gpt-6-luna} (through the OpenAI Decisions API) judges
necessary for the current question raises its accuracy on that question by \unsigned\fuMrcOaiTypedVsRandomTFKnownDelta{} points at 25\% of the tokens. On the later questions,
however, no method that selects sentences for the current question does detectably better than deleting sentences at
random. Deleting words instead of sentences does better: \LLMLtwo{}, which ignores the question, beats random sentence
deletion at the same token count by \unsigned\wwMrcOaiLinguaRandSentTFHiddenRawGap{} points on the later MultiRC
questions. It keeps about twice random deletion's share of the evidence's content words, and at equal content-word
retention its gain shrinks to \unsigned\wwMrcOaiLinguaRandSentTFHiddenEvContentGap{} points (exploratory). On
FEVER, label accuracy can overstate what the kept text supports: the gain of sentence selection is no longer detected
when answers are scored against the kept evidence, and the \unsigned\wwFeverLinguaRandSentTFHiddenRawGap-point gain of
\LLMLtwo{} is not shown to rest on the kept text. The ranking also
depends on the setting: on Evidence Inference sentence selection beats \LLMLtwo{} at a nominal 25\% budget, and on
$\tau$-bench trajectories, whose later tool calls need exact identifiers, it beats random deletion while \LLMLtwo{}
shows no detected gain. Scores on the question a compaction was made for do not reliably show how it serves later
ones.
\end{abstract}

\section{Introduction}

Applications that call a language model many times, such as agents, assistants and multi-step pipelines, carry
text from one call to the next: documents read so far, conversation turns, tool outputs. We call this text the
\emph{state}. Every call pays for it in tokens and the context window bounds it, so applications \emph{compact} it:
they delete parts of it to fit a budget. At compaction time the application usually knows its current question, and
the natural choice is to keep what that question needs. The state, however, outlives that question. Later
questions, unknown when the deletion was made, must be answered from whatever survived, and a deleted sentence is
gone for all of them. We call the question available at compaction time \emph{known} and the later ones
\emph{hidden}. Query-aware compressors are usually evaluated on the question they compress for; we evaluate them on the hidden ones.

\begin{figure}[t]
\centering\small
\fbox{\parbox{0.97\linewidth}{%
\textbf{State} (a MultiRC paragraph; ``\ldots'' marks words omitted here)\\[2pt]
\begin{tabular}{@{}r@{\hspace{5pt}}p{0.9\linewidth}@{}}
1 & Born in Moscow, Pushkin published his first poem at the age of fifteen.\\
2 & By the time he finished school \ldots\ his talent was \ldots\ recognized within the Russian literary scene.\\
3 & After school, Pushkin plunged into the \ldots\ intellectual youth culture of the capital, Saint Petersburg.\\
4 & In 1820 he published his first long poem, Ruslan and Lyudmila, amidst much controversy \ldots\\
5 & Pushkin gradually became committed to social reform and emerged as a spokesman for literary radicals.\\
6 & This angered the government, and led to his transfer from the capital (1820).\\
7 & He went to the Caucasus and to the Crimea, then to Kamenka and Chisinau \ldots\\
\end{tabular}\\[3pt]
\textbf{Known question:} ``Was Pushkin a published author and recognized by the Russian literary scene?''
(evidence: sentences 1, 2; correct answer ``Yes'').\\
\textbf{Hidden question:} ``What was going on with Pushkin in 1820?'' (evidence: sentences 4, 6; correct
answer ``He was being transferred from the capital'').\\[3pt]
\begin{tabular}{@{}lcccc@{}}
\toprule
text the reader sees & sentences kept & known answer & hidden answer & hidden sufficiency \\
\midrule
full paragraph & 1--7 & 1.00 & 0.99 & 1.00 \\
typed selection, 25\% budget & 1, 2 & 1.00 & 0.00 & 0.00 \\
random deletion, 25\% budget & 2, 6 & 0.76 & 1.00 & 1.00 \\
\bottomrule
\end{tabular}}}
\caption{Typed selection keeps the known question's evidence and deletes the hidden question's (one MultiRC
paragraph, chosen for illustration; Section~\ref{sec:results} gives averages). At a 25\%
budget two of seven sentences are kept. Typed selection asks \texttt{gpt-6-luna} whether each sentence is needed for
the known question and keeps the two most probable; random deletion happens to keep sentence 6. The columns give
the probabilities the same model assigns after reading each text: that the gold answer is correct, and that the
text contains what the hidden question needs (the sufficiency check of Section~\ref{sec:gate}).}
\label{fig:example}
\end{figure}

We compare query-aware selection, which keeps what the known question needs, with question-agnostic compressors.
Our main instance uses \emph{typed decision models}, which take a state and typed questions about it (a Boolean, a
Choice among options or a Score) and return a probability for each answer \citep{svc2026docs,openai2026decisions}.
\emph{Typed selection} asks one Boolean per sentence, ``is this sentence needed to answer the question?'', and keeps
the most probable sentences; the same model then answers the questions, and in this role we call it the
\emph{reader}. Figure~\ref{fig:example} shows the risk. We compare it with random deletion, question-agnostic
selection by the same model, TF-IDF (term frequency--inverse document frequency) selection and lead (the first sentences). We also run \LLMLtwo{}
\citep{pan2024llmlingua2}, which deletes individual words without any question, and, at token budgets, LongLLMLingua
\citep{jiang2024longllmlingua} and maximal marginal relevance (MMR; \citealp{carbonell1998mmr}), two established
compressors that use the question. The corpora are MultiRC \citep{khashabi2018multirc, wang2019superglue}, ERASER's
FEVER \citep{thorne2018fever,deyoung2020eraser}, Evidence Inference \citep{lehman2019inferring} and LoCoMo
conversation memory \citep{maharana2024locomo}, in which several questions share one text and each question's
\emph{evidence} is annotated, and recorded $\tau$-bench agent trajectories \citep{yao2024taubench}, whose questions
ask for the identifiers that later tool calls pass. Our findings are:
\begin{itemize}
\item \textbf{Selection for the known question beats random deletion on that question.} On MultiRC, typed selection, MMR and
LongLLMLingua raise known-question accuracy over random deletion at 25\%, but none shows a detected advantage on hidden
questions on average. On MultiRC, hidden questions gain only where they share the known question's evidence. On FEVER, a gain
in recovering a claim's label is not detected when the kept text is scored for what it supports
(Section~\ref{sec:results}). The label-accuracy gain of word-level compression there is not shown to rest on the
kept text either (Section~\ref{sec:compressors}).
\item \textbf{On MultiRC, picking words beats picking sentences by the same scores.} With the same importance scores,
picking words instead of sentences raises hidden-question accuracy, and \LLMLtwo{} and a hybrid of typed sentences and
\LLMLtwo{} words lead on hidden questions. They keep about as many evidence words as random deletion but far
more of the evidence's content words, which partly explains the lead: at equal content-word retention it shrinks from
\unsigned\wwMrcOaiLinguaRandSentTFHiddenRawGap{} to \unsigned\wwMrcOaiLinguaRandSentTFHiddenEvContentGap{} points with \texttt{gpt-6-luna}
(Section~\ref{sec:compressors}).
\item \textbf{The ordering is not stable across settings.} On Evidence Inference typed selection beats \LLMLtwo{} on
hidden claims at a nominal 25\% budget. On agent trajectories, whose later calls need exact identifiers, typed selection beats random deletion
and \LLMLtwo{} shows no detected gain (Section~\ref{sec:boundary}).
\end{itemize}
Concurrent work shows that compaction can serve the current query and fail later ones
\citep{arxiv2608_00902,arxiv2609_20045}. We add an evaluation on annotated evidence that compares sentence- and
word-level compression and locates where each serves later questions.

\section{Evaluation setup}
\label{sec:setup}

This section describes the MultiRC experiment; Sections~\ref{sec:fever} and~\ref{sec:boundary} give the changes for
the other settings.

\paragraph{Data and models.} Each MultiRC question has several candidate answers that annotators labeled correct or
incorrect, and a question can have more than one correct answer. We use \CpNParas{} paragraphs with at least three
questions and \CpNSentLo--\CpNSentHi{} sentences, with \CpNQuestions{} questions and \CpNAnswers{} candidate
answers, from the validation and training splits (validation alone has too few such paragraphs). The first listed
question of each paragraph is known and the others are hidden. \texttt{gpt-6-luna} reads every corpus. Every
\texttt{gpt-6-luna} read and selection in this paper goes through the OpenAI Decisions API \citep{openai2026decisions},
the typed decision interface: a request poses typed questions and returns a probability for each answer, with no
generated text. Three open-weight models built for the typed decision
interface, \texttt{open-decider} (2B), \texttt{open-bosun} (1.7B) and \texttt{open-lumma} (0.6B)
\citep{mapika2026decider,hanno2026bosun,frontiersmind2026lumma}, also read MultiRC from their released weights, and
\texttt{qwen3.8-flash} is a general instruction-tuned reader (Appendix~\ref{app:reader}). We drop refused answers;
counting them as wrong instead changes neither the sign of any typed $-$ random contrast nor whether its interval
excludes zero (Appendix~\ref{app:runs}).

\paragraph{Selectors and budgets.} Both model-based selectors ask one Boolean per sentence, with the sentences
numbered in the state. Typed selection puts the known question in the state and asks ``is sentence $i$ needed to
answer the question?''. Question-agnostic selection gives no question and asks ``does sentence $i$ state a fact that a
reader is likely to be asked about later?''. A \emph{policy} keeps a \emph{budget} $b \in \{25, 50, 75\}\%$ of a
paragraph's sentences (rounded half to even, at least one). Typed and agnostic selection keep the most probable sentences, TF-IDF
selection those closest to the known question, lead the first ones, and random deletion one seeded uniform sample.
Kept sentences stay in document order. Token-matched tests cut every text, \LLMLtwo{}'s included, to the same o200k
token count (Appendix~\ref{app:followup}).

\paragraph{Reading and measures.} One request asks one Boolean per candidate answer: ``given the passage, is this a
correct answer to the question?'', where ``no'' is defined as ``not supported or incorrect''. Once a compaction
deletes the evidence for a correct answer, the reader is therefore asked to reject that answer. We score each
candidate answer against the MultiRC labels, average within each paragraph, then over paragraphs, with
paragraph-bootstrap 95\% intervals; a difference is \emph{detected} when its interval excludes zero. Answering ``no'' to
every candidate already scores \auRdNo\%, and of the four typed readers only \texttt{gpt-6-luna} (\auRdOaiAcc\%) and \texttt{open-decider}
(\auRdDecAcc\%) are detectably above it on the full text, so the main tables report these two. A hidden question is
\emph{overlapping} if its annotated evidence shares a sentence with the known question's, and \emph{disjoint}
otherwise.

\paragraph{Prespecified analyses.} Hypotheses, tests and decision rules were specified before any model request for
the Evidence Inference overlap test, the KEEP--DROP contrast, the LoCoMo test and four follow-ups under one plan
(token-matched budgets, query-aware baselines, three-way labels and harder LoCoMo options). The same holds for the \LLMLtwo{} granularity
test, the hybrid compactor, a test with a different known question, a general-reader test, the $\tau$-bench test and the abstractive-summary test.
Appendix~\ref{app:audit} counts the outcomes and lists every prediction that was not fully met. The other analyses,
including the content-word analysis and the FEVER evidence-support check of Section~\ref{sec:compressors},
are exploratory. The ancillary files contain the summary tables and the values of all reported numbers.

\section{Selection for the known question beats random deletion on that question}
\label{sec:results}

\begin{table}[t]
\centering\footnotesize
\setlength{\tabcolsep}{2.7pt}
\caption{Typed selection keeps known-question accuracy above random deletion, while hidden-question accuracy
stays near random. Readers: \texttt{open-decider} and \texttt{gpt-6-luna}, the typed readers whose full-text accuracy
exceeds the all-``no'' rate; agnostic: question-agnostic selection.
Paragraph-macro candidate-level accuracy (\%) on the known and hidden questions after compaction, and policy
$-$ random hidden accuracy (points, paragraph-bootstrap 95\% interval). Table~\ref{tab:gate} gives the
sufficiency check for the same rows.}
\label{tab:open}
\input{generated/compress_open_main.tex}
\end{table}

\paragraph{The known question gains.} At a 25\% budget, \texttt{open-decider} answers the known question at
\CpDecKnownTFKnown\% after typed selection, against \CpDecRandomTFKnown\% after random deletion and
\CpDecOrigKnown\% on the full text; for \texttt{gpt-6-luna} the same accuracies are \CpOaiKnownTFKnown\%,
\CpOaiRandomTFKnown\% and \CpOaiOrigKnown\% (compacted texts in Table~\ref{tab:open}).

\paragraph{Averaged over hidden questions, selection stays near random deletion.} Typed $-$ random hidden accuracy
is slightly below zero for both readers at 25\% and 50\%, and question-agnostic selection is similarly close to random
(Table~\ref{tab:open}). The split holds when texts are cut to the same token count: typed $-$ random at 25\% with
\texttt{gpt-6-luna} is \ci{\fuMrcOaiTypedVsRandomTFKnownDelta}{\fuMrcOaiTypedVsRandomTFKnownLo}{\fuMrcOaiTypedVsRandomTFKnownHi}
points on the known question and \ci{\fuMrcOaiTypedVsRandomTFHiddenDelta}{\fuMrcOaiTypedVsRandomTFHiddenLo}{\fuMrcOaiTypedVsRandomTFHiddenHi}
on hidden ones. LongLLMLingua and MMR show the same split with both readers, and on hidden questions LongLLMLingua
falls below random deletion (\ci{\fuMrcOaiLongVsRandomTFHiddenDelta}{\fuMrcOaiLongVsRandomTFHiddenLo}{\fuMrcOaiLongVsRandomTFHiddenHi}
with \texttt{gpt-6-luna}; Appendix~\ref{app:followup}). With \texttt{qwen3.8-flash} as reader, typed $-$ random is
\ci{\rdMrcTypedRandTFKnownDelta}{\rdMrcTypedRandTFKnownLo}{\rdMrcTypedRandTFKnownHi} on the known question and
\ci{\rdMrcTypedRandTFHiddenDelta}{\rdMrcTypedRandTFHiddenLo}{\rdMrcTypedRandTFHiddenHi} on hidden ones.

\begin{figure}[t]
\centering
\includegraphics[width=\linewidth]{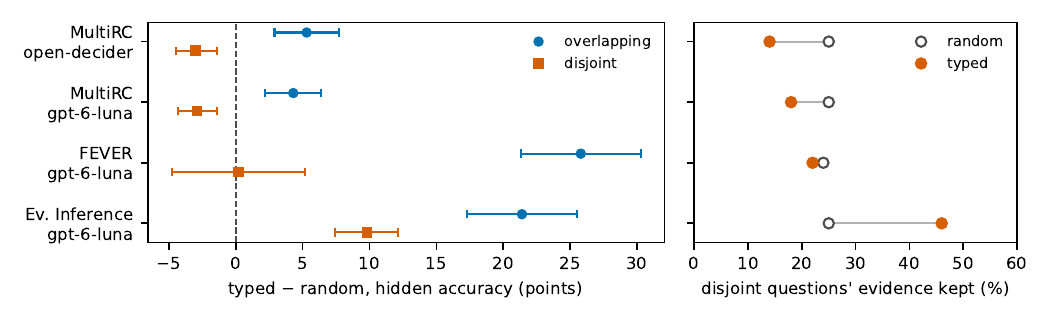}
\caption{Hidden questions that share evidence with the known question gain from typed selection; on Evidence
Inference, disjoint claims gain too, and more of their evidence is kept. 25\% budget. Left: typed $-$ random hidden
accuracy with bootstrap 95\% intervals, for questions whose evidence overlaps the known question's and for disjoint
ones. Right: share of disjoint questions' evidence kept by typed selection and by random deletion (for MultiRC and
FEVER the exact expectation for a random subset of the same size). Values and overlap shares in
Table~\ref{tab:third}.}
\label{fig:overlap}
\end{figure}

\paragraph{Hidden questions gain where they share the known question's evidence.} ERASER \citep{deyoung2020eraser}
marks evidence sentences for \ErNQuestions{} of the \CpNQuestions{} MultiRC questions. Typed selection at 25\% keeps
\ErOaiKnownTFKnownRecall\% of the known question's evidence (\texttt{gpt-6-luna}) against \ErDecRandExpKnownRecall\%
expected for a random subset, and about the random share for hidden questions. Hidden questions that share evidence with the
known one gain \ci{\poOaiTFOverlapDelta}{\poOaiTFOverlapLo}{\poOaiTFOverlapHi} points
(\texttt{gpt-6-luna}) and \ci{\poDecTFOverlapDelta}{\poDecTFOverlapLo}{\poDecTFOverlapHi} (\texttt{open-decider}).
Disjoint ones lose \ci{\poOaiTFDisjointDelta}{\poOaiTFDisjointLo}{\poOaiTFDisjointHi} and
\ci{\poDecTFDisjointDelta}{\poDecTFDisjointLo}{\poDecTFDisjointHi}, with less of their evidence kept than under random
deletion (\poOaiTFDisjointRecall\% against \poOaiTFDisjointRandRecall\%; Figure~\ref{fig:overlap}).

\paragraph{Deleted evidence drives the losses.}\label{sec:eraser} A hidden answer that was right on the full text flips more often when its evidence
was deleted. With \texttt{gpt-6-luna}, adding some of the deleted evidence back undoes more flips than adding as many sentences that are evidence
for no question (\ci{+\RsConcDiffDec}{+\RsConcDiffLoDec}{+\RsConcDiffHiDec} points; Appendix~\ref{app:restore}).
\label{sec:gate}A sufficiency check (``does the passage contain the information needed to answer this question?'')
ranks questions with broken evidence first. Yet re-reading a random quarter of hidden questions on the full text
recovers most of what re-reading the lowest-sufficiency quarter does (Appendix~\ref{app:gate}).

\paragraph{FEVER and Evidence Inference: hidden claims gain in label recovery.}\label{sec:fever} Each question here is
a claim, and the reader judges whether the text supports it. We sample \cfNDocs{} FEVER documents with
\cfNHidden{} hidden claims and \ctNDocs{} Evidence Inference clinical-trial articles (Appendix~\ref{app:fever}); the first claim of each document is known. On FEVER at 25\% typed $-$ random hidden-claim
accuracy is \cfKnownRandomTFHiddenDelta{} points, and on Evidence Inference disjoint claims gain too, with more of their
evidence kept than random deletion keeps (Figure~\ref{fig:overlap}).

\paragraph{Label recovery and support scoring disagree.} Label accuracy counts the document's label as
right even when a claim's evidence was deleted. A prespecified three-way read therefore also scores each answer
against what the kept text supports: the document's label if all of the claim's annotated evidence was kept and ``not
enough info'' if all of it was deleted. In this read, with true and false claims weighted equally, typed selection on FEVER
recovers the label \unsigned\fuCFeverTypedVsRandomTFRecoveryDelta{} points more often than random deletion, while its
support-target accuracy, with the three targets weighted equally, differs from random deletion by
\ci{\fuCFeverTypedVsRandomTFSupportDelta}{\fuCFeverTypedVsRandomTFSupportLo}{\fuCFeverTypedVsRandomTFSupportHi}
points (Appendix~\ref{app:support}). Prespecified KEEP--DROP tests give a hidden claim two texts with the same number of
sentences, one with its evidence and one without; KEEP beats DROP by
\ci{\prtFeverKeepDropAllDelta}{\prtFeverKeepDropAllLo}{\prtFeverKeepDropAllHi} points on FEVER and
\ci{\prtEiKeepDropAllDelta}{\prtEiKeepDropAllLo}{\prtEiKeepDropAllHi} on Evidence Inference (Appendix~\ref{app:retention}).

\paragraph{For sentence selection, the gain follows the extra evidence kept.}\label{sec:mechanism} Across the eight
25\% groups of Figure~\ref{fig:overlap} (corpus, reader, overlapping or disjoint), the typed $-$ random hidden gain
rises with how much more of the hidden questions' evidence typed selection keeps than random deletion, at
$0.37$ points of gain per point of extra evidence. We fixed this line before a second experiment that repeats typed
selection with a different, randomly drawn known question in every document. On the eight new groups it misses the
observed gains by \kqMAE{} points on average. The largest misses are \kqMissMax{} points (overlapping FEVER claims,
under-predicted) and \kqMissSecond{} (overlapping MultiRC questions read by \texttt{open-decider}, over-predicted);
the other six are within \kqMissThird{} points (Appendix~\ref{app:known}).

\section{On MultiRC, word-level compression serves later questions}
\label{sec:compressors}

\LLMLtwo{} scores every word of the text without any question and keeps the highest-scoring words. A prespecified
hybrid keeps typed whole sentences up to half the token budget and fills the rest with the highest-scoring \LLMLtwo{}
words of the other sentences (Appendix~\ref{app:hybrid}). Table~\ref{tab:words} compares them with sentence selection
at 25\% of the tokens.

\begin{table}[t]
\centering\small
\setlength{\tabcolsep}{3.5pt}
\caption{Word-level compression leads on hidden MultiRC and FEVER questions and keeps more of their evidence's
content words than random sentence deletion; on MultiRC, about the same share of its words. 25\% of the tokens; FEVER accuracy is label accuracy. Accuracy (\%, document-macro) on hidden questions with annotated evidence, and share
of the hidden questions' annotated evidence kept (\%), counted over all words and over content words, answer-weighted (content words: words outside
a standard English stop list). Random words: a uniform sample of words. Hybrid: typed whole sentences up to half the
budget, the rest \LLMLtwo{} words (Appendix~\ref{app:hybrid}).}
\label{tab:words}
\begin{tabular}{lrrrrrrrrr}
\toprule
& \multicolumn{3}{c}{MultiRC, \texttt{gpt-6-luna}} & \multicolumn{3}{c}{MultiRC, \texttt{open-decider}} & \multicolumn{3}{c}{FEVER, \texttt{gpt-6-luna}} \\
\cmidrule(lr){2-4}\cmidrule(lr){5-7}\cmidrule(lr){8-10}
text & acc. & words & content & acc. & words & content & acc. & words & content \\
\midrule
random sentences & \wwMrcOaiRandSentTFHiddenAcc & \wwMrcOaiRandSentTFHiddenEvWords & \wwMrcOaiRandSentTFHiddenEvContent & \wwMrcDecRandSentTFHiddenAcc & \wwMrcDecRandSentTFHiddenEvWords & \wwMrcDecRandSentTFHiddenEvContent & \wwFeverRandSentTFHiddenAcc & \wwFeverRandSentTFHiddenEvWords & \wwFeverRandSentTFHiddenEvContent \\
typed sentences & \wwMrcOaiTypedTFHiddenAcc & \wwMrcOaiTypedTFHiddenEvWords & \wwMrcOaiTypedTFHiddenEvContent & \wwMrcDecTypedTFHiddenAcc & \wwMrcDecTypedTFHiddenEvWords & \wwMrcDecTypedTFHiddenEvContent & \wwFeverTypedTFHiddenAcc & \wwFeverTypedTFHiddenEvWords & \wwFeverTypedTFHiddenEvContent \\
random words & \wwMrcOaiRandWordTFHiddenAcc & \wwMrcOaiRandWordTFHiddenEvWords & \wwMrcOaiRandWordTFHiddenEvContent & \wwMrcDecRandWordTFHiddenAcc & \wwMrcDecRandWordTFHiddenEvWords & \wwMrcDecRandWordTFHiddenEvContent & \wwFeverRandWordTFHiddenAcc & \wwFeverRandWordTFHiddenEvWords & \wwFeverRandWordTFHiddenEvContent \\
\LLMLtwo{} & \wwMrcOaiLinguaTFHiddenAcc & \wwMrcOaiLinguaTFHiddenEvWords & \wwMrcOaiLinguaTFHiddenEvContent & \wwMrcDecLinguaTFHiddenAcc & \wwMrcDecLinguaTFHiddenEvWords & \wwMrcDecLinguaTFHiddenEvContent & \wwFeverLinguaTFHiddenAcc & \wwFeverLinguaTFHiddenEvWords & \wwFeverLinguaTFHiddenEvContent \\
hybrid & \wwMrcOaiHybHalfTFHiddenAcc & \wwMrcOaiHybHalfTFHiddenEvWords & \wwMrcOaiHybHalfTFHiddenEvContent & \wwMrcDecHybHalfTFHiddenAcc & \wwMrcDecHybHalfTFHiddenEvWords & \wwMrcDecHybHalfTFHiddenEvContent & \wwFeverHybHalfTFHiddenAcc & \wwFeverHybHalfTFHiddenEvWords & \wwFeverHybHalfTFHiddenEvContent \\
\bottomrule
\end{tabular}
\end{table}

\paragraph{Word-level compression leads on hidden questions.} \LLMLtwo{} $-$ random sentence deletion on hidden
questions with annotated evidence, from the model of Appendix~\ref{app:whywords} without retention terms, is
\ci{\wwMrcOaiLinguaRandSentTFHiddenRawGap}{\wwMrcOaiLinguaRandSentTFHiddenRawLo}{\wwMrcOaiLinguaRandSentTFHiddenRawHi}
points on MultiRC with \texttt{gpt-6-luna},
\ci{\wwMrcDecLinguaRandSentTFHiddenRawGap}{\wwMrcDecLinguaRandSentTFHiddenRawLo}{\wwMrcDecLinguaRandSentTFHiddenRawHi}
with \texttt{open-decider} and
\ci{\wwFeverLinguaRandSentTFHiddenRawGap}{\wwFeverLinguaRandSentTFHiddenRawLo}{\wwFeverLinguaRandSentTFHiddenRawHi}
in label accuracy on FEVER; the paired contrast over all hidden questions, dropping refusals pairwise, gives
\fuMrcOaiLinguaVsRandomTFHiddenDelta{} and \fuFeverLinguaVsRandomTFHiddenDelta{} for the two \texttt{gpt-6-luna} cells (Appendix~\ref{app:followup}). \LLMLtwo{} is also ahead of typed selection on hidden questions in all three
cells (Table~\ref{tab:words}). The hybrid beats typed selection on hidden questions in all three cells, by
\unsigned\hyMrcDecHybHalfTypedTFHiddenDelta{} to \unsigned\hyMrcOaiHybHalfTypedTFHiddenDelta{} points with every
interval above zero; it beats \LLMLtwo{} on FEVER and stays below it on MultiRC (Appendix~\ref{app:hybrid}).

\paragraph{The gain comes from picking words.} A prespecified test keeps the \LLMLtwo{} scores and changes only the
unit kept. Picking words instead of whole sentences raises hidden-question accuracy at 25\% by
\grMrcOaiGranScorerTFDelta{} points (MultiRC, \texttt{gpt-6-luna}), \grMrcDecGranScorerTFDelta{} (MultiRC,
\texttt{open-decider}) and \grFeverGranScorerTFDelta{} (FEVER). Whole sentences picked by the same scores are not
detectably better than random sentence deletion in five of six cells (Appendix~\ref{app:granularity}). On the known
MultiRC question \LLMLtwo{} is ahead of typed selection with \texttt{gpt-6-luna} in all four comparisons. The two
intervals at sentence budgets exclude zero, and the one at 50\% of the tokens just does (upper end $-0.02$ points) (Appendices~\ref{app:lingua} and~\ref{app:granularity}).

\paragraph{Word-level compression keeps the evidence's content words.} Counted over all words, \LLMLtwo{}, the hybrid
and random sentence deletion keep similar shares of the hidden questions' evidence on MultiRC, yet their accuracies
differ by up to \unsigned\wwMrcOaiLinguaRandSentTFHiddenRawGap{} points. Sentence deletion keeps a sentence's content
and function words together. \LLMLtwo{} drops function words first, so it keeps \wwMrcOaiLinguaTFHiddenEvContent\% of
the evidence's content words against \wwMrcOaiRandSentTFHiddenEvContent\% for random sentence deletion
(Table~\ref{tab:words}). For sentence policies the two counts agree within half a point, so the relation of
Section~\ref{sec:mechanism} is unchanged.

\paragraph{Content-word retention accounts for part of the lead.} A model of each answer's correctness,
with document fixed effects, compares policies at equal content-word retention. It describes an association, since
the policies produce the retention they are compared at. At equal retention the MultiRC \LLMLtwo{} $-$
random-sentence gap shrinks from \unsigned\wwMrcOaiLinguaRandSentTFHiddenRawGap{} to
\ci{\wwMrcOaiLinguaRandSentTFHiddenEvContentGap}{\wwMrcOaiLinguaRandSentTFHiddenEvContentLo}{\wwMrcOaiLinguaRandSentTFHiddenEvContentHi}
points with \texttt{gpt-6-luna} and from \unsigned\wwMrcDecLinguaRandSentTFHiddenRawGap{} to
\ci{\wwMrcDecLinguaRandSentTFHiddenEvContentGap}{\wwMrcDecLinguaRandSentTFHiddenEvContentLo}{\wwMrcDecLinguaRandSentTFHiddenEvContentHi}
with \texttt{open-decider}. Also conditioning on how many of each candidate answer's content words survive leaves
\wwMrcOaiLinguaRandSentTFHiddenContentAnsGap{} points with \texttt{gpt-6-luna} and
\ci{\wwMrcDecLinguaRandSentTFHiddenContentAnsGap}{\wwMrcDecLinguaRandSentTFHiddenContentAnsLo}{\wwMrcDecLinguaRandSentTFHiddenContentAnsHi}
with \texttt{open-decider}. The hybrid $-$
typed gaps shrink the same way (Table~\ref{tab:whywords}). On FEVER binned retention leaves
\unsigned\wwFeverLinguaRandSentTFHiddenEvContentGap{} points; a linear term leaves
\unsigned\wwFeverLinguaRandSentTFHiddenEvContentLinGap, because accuracy rises non-linearly with retention there.

\paragraph{Which words are kept matters too.} On FEVER typed selection keeps more of the hidden claims' content words
than \LLMLtwo{} yet scores lower (Table~\ref{tab:words}). At the same content-word retention \LLMLtwo{} still beats
random word deletion by
\ci{\wwMrcOaiLinguaRandWordTFHiddenEvContentGap}{\wwMrcOaiLinguaRandWordTFHiddenEvContentLo}{\wwMrcOaiLinguaRandWordTFHiddenEvContentHi}
points on MultiRC with \texttt{gpt-6-luna},
\ci{\wwMrcDecLinguaRandWordTFHiddenEvContentGap}{\wwMrcDecLinguaRandWordTFHiddenEvContentLo}{\wwMrcDecLinguaRandWordTFHiddenEvContentHi}
with \texttt{open-decider} and
\ci{\wwFeverLinguaRandWordTFHiddenEvContentGap}{\wwFeverLinguaRandWordTFHiddenEvContentLo}{\wwFeverLinguaRandWordTFHiddenEvContentHi}
on FEVER (Appendix~\ref{app:whywords}).

\paragraph{LLM-written summaries stay below word-level compression.} A compaction can also have an LLM rewrite the
text. Two summarizers, \texttt{qwen3.8-flash} and \texttt{gpt-6.1-sol}, summarized each MultiRC paragraph within the
same token budget, once for unknown later questions and once for the known question (Appendix~\ref{app:summary}). On
hidden questions the generic summaries beat random sentence deletion by \unsigned\smSumGenRandSentHiddenDelta{} and
\unsigned\smGptGenRandSentHiddenDelta{} points but stay below \LLMLtwo{}: summary $-$ \LLMLtwo{} is
\ci{\smSumGenLinguaHiddenDelta}{\smSumGenLinguaHiddenLo}{\smSumGenLinguaHiddenHi} and
\ci{\smGptGenLinguaHiddenDelta}{\smGptGenLinguaHiddenLo}{\smGptGenLinguaHiddenHi} points. The summaries written for the known
question answer it better than typed selection and \LLMLtwo{} (\smSumFocKnown\% and \smGptFocKnown\%), but on hidden questions they fall
below random deletion (\texttt{qwen3.8-flash}, \ci{\smSumFocRandSentHiddenDelta}{\smSumFocRandSentHiddenLo}{\smSumFocRandSentHiddenHi}) or show no detected gain
(\texttt{gpt-6.1-sol}, \ci{\smGptFocRandSentHiddenDelta}{\smGptFocRandSentHiddenLo}{\smGptFocRandSentHiddenHi}).

\paragraph{On FEVER, label accuracy overstates the word-level gain.} FEVER accuracy counts the document's label,
which Section~\ref{sec:fever} shows can be right without textual support. The \LLMLtwo{} gain over random sentence
deletion lies on true claims (\ci{\fsHyLabTrueLinguaRandSentDelta}{\fsHyLabTrueLinguaRandSentLo}{\fsHyLabTrueLinguaRandSentHi}
points; false claims \ci{\fsHyLabFalseLinguaRandSentDelta}{\fsHyLabFalseLinguaRandSentLo}{\fsHyLabFalseLinguaRandSentHi}).
When at most a fifth of a true claim's evidence content words survive, the reader still calls it supported for
\fsHyYesRLeRelaxLinguaAcc\% of claims after \LLMLtwo{} and \fsHyYesRLeRelaxRandSentAcc\% after random sentence deletion. An
evidence-support target, scored on the binary reads, reproduces the three-way result for typed $-$ random
(\ci{\fsOrigStrictTypedRandSentBalDelta}{\fsOrigStrictTypedRandSentBalLo}{\fsOrigStrictTypedRandSentBalHi} points,
label-balanced) and removes the hybrid's gain over typed selection
(\ci{\fsHyStrictHybHalfTypedBalDelta}{\fsHyStrictHybHalfTypedBalLo}{\fsHyStrictHybHalfTypedBalHi}). Under \LLMLtwo{} it is
undefined for \fsHyPartialShareLingua\% of hidden claims, whose evidence is partly kept, so its FEVER gain
is not shown to rest on the kept text (Appendix~\ref{app:whywords}).

\section{The ordering is not stable: Evidence Inference and agent states}
\label{sec:boundary}

\paragraph{Evidence Inference.} On Evidence Inference typed selection keeps more of even the disjoint claims'
evidence than random deletion (Figure~\ref{fig:overlap}). It beats \LLMLtwo{} on hidden claims when \LLMLtwo{} is
set to the compression rate of 25\% sentence deletion
(\ci{\llEiKnownLinguaTFHiddenDelta}{\llEiKnownLinguaTFHiddenLo}{\llEiKnownLinguaTFHiddenHi} points); at the 50\% rate the
difference is \ci{\llEiKnownLinguaFHiddenDelta}{\llEiKnownLinguaFHiddenLo}{\llEiKnownLinguaFHiddenHi}. This comparison is not
token-matched: at 25\% the typed texts keep \csEiTypedTFTokShare\% of the article's o200k tokens and the \LLMLtwo{} texts
\csEiLinguaTFTokShare\% (Appendices~\ref{app:lingua} and~\ref{app:cost}).

\paragraph{Agent trajectories.}\label{sec:tau} In the recorded $\tau$-bench retail trajectories an agent calls tools
for a simulated user. The state is a trajectory up to its middle tool output, split into turns, tool calls and tool
outputs. The known question asks which identifier (an order, item, product or payment-method ID, or a zip code) the next tool call passes;
each hidden question asks the same for a later call, among same-shaped identifiers from the state. On
\taNStates{} states with \taNHidden{} hidden questions at 25\% of the tokens, typed $-$ random is
\ci{\taTypedVsRandomHDelta}{\taTypedVsRandomHLo}{\taTypedVsRandomHHi} points with \texttt{gpt-6-luna} and
\ci{\taQwenTypedVsRandomHDelta}{\taQwenTypedVsRandomHLo}{\taQwenTypedVsRandomHHi} with \texttt{qwen3.8-flash}, while
\LLMLtwo{} shows no detected gain (\ci{\taLinguaVsRandomHDelta}{\taLinguaVsRandomHLo}{\taLinguaVsRandomHHi}); token-level
extractive compressors are known to strip the action syntax that agents need \citep{arxiv2605_26596}. Typed
selection keeps the hidden question's identifier for \taGoldTypedH\% of hidden questions against \taGoldRandomH\%
(random) and \taGoldLinguaH\% (\LLMLtwo{}).

\paragraph{Anticipating later calls costs the known one.} Selectors that anticipate later calls trade the known call for the hidden
ones: on a \taNECovStates-state subsample, expected-coverage selection gains
\ci{\taECovTVsTypedHDelta}{\taECovTVsTypedHLo}{\taECovTVsTypedHHi} points over typed selection on hidden questions and
\taECovTVsTypedKDelta{} (90\% interval [\taECovTVsTypedKLoNinety, \taECovTVsTypedKHiNinety]) on the known one.
About half of the hidden gain comes from fewer refusals: scored on answered questions only, it is
\ci{\taAnsECovTVsTypedHDelta}{\taAnsECovTVsTypedHLo}{\taAnsECovTVsTypedHHi} (Appendix~\ref{app:tau}).

\paragraph{Conversation memory.}\label{sec:locomo} LoCoMo's ten long conversations are compacted four times as they
grow, so a dropped turn never returns. At the last checkpoint, a typed 25\% memory with a hidden question's evidence
beats one with the same number of turns without it by \unsigned\lcmHTwoADelta{} points (\texttt{gpt-6-luna}), in all
\lcmHTwoANConvPos{} conversations. Typed versus random deletion is \lcmHOneVerdict{} for both readers, and on the
four-choice format empty memory already scores \lcmAccNoneKD\% (Appendix~\ref{app:locomo}).

\section{Discussion}
\label{sec:discussion}

Query-aware sentence selection raises the known question's accuracy over random deletion. On later questions it
helps those that share the known question's evidence. On MultiRC, where many do not, it does no better than random
deletion on average.

On MultiRC the compressors that lead on later questions delete words. They drop function words first, so
the same token budget holds more of the evidence's content words. At equal content-word retention two fifths to two
thirds of \LLMLtwo{}'s MultiRC lead over random sentence deletion disappears, and \LLMLtwo{} still beats random words, so
which words are kept also matters. On FEVER, typed selection's label gain is not detected under support scoring, and the
word-level gain is not shown to rest on the kept text. The ordering
is not general. On Evidence Inference, where even disjoint claims keep more evidence under typed selection, it beats \LLMLtwo{}
at a nominal 25\% budget. On agent trajectories, whose later calls need exact identifiers, \LLMLtwo{}
keeps fewer of them than typed selection and shows no detected gain.

For evaluation, the results suggest scoring hidden questions as well as the known one, comparing with random
deletion at matched tokens, measuring the hidden questions' evidence kept in content words, and reporting
support-based scoring next to label recovery. Compaction also costs tokens to build: for typed selection at 25\%
with \texttt{gpt-6-luna}, compacting once pays for itself in input tokens only with one request per later question,
after about \csEiCompactBreakEven{} to \csLocomoCompactBreakEven{} of them on FEVER, Evidence Inference and LoCoMo.
On Evidence Inference this is more than its mean of \csEiKbar{} later questions, and on MultiRC the extrapolated
break-even, \csMrcCompactBreakEven, exceeds its mean of \csMrcKbar{} (Appendix~\ref{app:cost}). Keeping the
full state allows a re-read for a later question, but neither recovery we tested matches it. On MultiRC, re-reading
the lowest-sufficiency quarter of hidden questions on the full text lifts accuracy over all hidden questions from \GtOaiKnownTFAccCompact\%
to \GtOaiKnownTFGatedTwentyFive\%, against \GtOaiKnownTFAccOrig\% on the full text (Appendix~\ref{app:gate}). On agent
trajectories, fetching dropped units on demand reaches \taRecGistPtrAccB\% against \taFullItemAcc\% for the full state
at \taRecGistPtrTokShareB\% of its passage tokens, and the prespecified prediction that it would come close is not met
(Appendix~\ref{app:tau}).

\section{Related work}
\label{sec:related}

Prompt compression removes tokens or sentences by perplexity \citep{jiang2023llmlingua}, by query-aware
importance \citep{jiang2024longllmlingua} or by a learned task-agnostic importance
\citep{pan2024llmlingua2}, and key--value (KV) cache compression has the same split \citep{kim2025kvzip,li2024scbench}. Earlier
agent-memory work studied the same write-before-query asymmetry. TierMem \citep{zhu2026tiermem} falls back to a raw
log when a router judges pre-query summaries insufficient, and EMBER \citep{li2026ember} learns which evidence to
retain before the query. \citet{chen2026contrastkv} find that query-aware KV eviction does not generalize across
queries, and \citet{arxiv2605_24579} separate write-time from retrieval-time loss. We delete sentences (turns
on LoCoMo) for one known question and test the questions hidden at deletion time with annotated evidence and
KEEP--DROP substitutions. The closest work is concurrent. \citet{arxiv2608_01326} formalize compaction against later
queries, with a set-membership case study, and \citet{arxiv2608_00902} find that compacting agent KV caches before
later queries arrive hurts, and that deferring compaction recovers much of the cost; RestoreKV
\citep{arxiv2608_01247} adds learned restoration to query-agnostic eviction, and \citet{arxiv2607_11942} find that hiding the
question at compression time reorders KV-cache compressors, with most no better than trivial baselines. These work on
KV caches rather than on evidence-annotated text. \citet{arxiv2608_16370} counts the extra retrieval calls an agent makes to re-acquire state
that compression dropped, and \citet{arxiv2609_20045} tests whether compressed memory keeps what a later update needs
on synthetic histories; we measure answer accuracy on later questions of existing corpora and locate the loss with
evidence annotations. \citet{arxiv2608_11242} find that context compactors keep about a sixth of the constraints a user sets for
the agent's later behavior. \citet{arxiv2610_02897} and \citet{arxiv2609_32805} optimize what is written when the later
goal is known. Mnemon \citep{arxiv2609_36059} uses the same per-record ``is this needed'' Boolean but keeps its raw
records recoverable. \citet{arxiv2610_01006} find that asking whether the evidence suffices is more informative
than answer confidence; we ask it after compaction has deleted part of the evidence. \citet{arxiv2608_01631} find that
KV-cache eviction can keep final-answer accuracy while the reasoning that supports the answer degrades; on FEVER we
find the analogous gap between label accuracy and the evidence the kept text retains.

\section{Limitations}

The experiments use public text and conversation benchmarks and recorded agent trajectories, not live agents. Abstractive summaries are tested only on MultiRC at a 25\% budget. In
the $\tau$-bench test the actions and tool outputs are fixed by the recording, the questions ask only for
identifiers, retail is the only domain with enough trajectories for inference, and the only budget is 25\%. The relation between gain and evidence kept is tested by consistency across settings,
KEEP--DROP interventions and one out-of-sample check on the three document corpora, for sentence
selection; it could not be assessed on LoCoMo and holds only qualitatively on $\tau$-bench. The content-word analysis of word-level compression is exploratory,
leaves part of the MultiRC gap unexplained and, on FEVER, does not separate content words from all words. The model-based sentence
selectors are typed models, which usually also answer; \texttt{qwen3.8-flash} is the
only general instruction-tuned reader, and it reads the texts that \texttt{gpt-6-luna} selected. Only two typed readers perform above the all-``no'' baseline on MultiRC. On the
three document corpora the known question is the first listed one or, in one test, one other randomly drawn question
per document. Evidence annotations are incomplete, so neither the retention tests nor the evidence-support target can
account for unannotated evidence.

Budgets count sentences or conversation turns, except in the token-matched tests and the $\tau$-bench test; Appendix~\ref{app:cost} counts
input tokens. The KEEP--DROP comparison covers one budget and does not include MultiRC. LoCoMo has ten conversations
and a four-choice format on which empty memory already scores \lcmAccNoneKD\%. Hosted reads taken at different times
can differ slightly (Appendix~\ref{app:masks}). Intervals are unadjusted for multiple comparisons. The benchmark
texts predate the models, so contamination cannot be ruled out.

\section*{Broader Impact Statement}

Compaction that deletes a question's evidence can make supported answers ``unsupported''. Without
provenance, a downstream consumer may take them for answers that are genuinely unsupported or false, and the
sufficiency check does not flag every such case. Recoverable full-state designs should be evaluated before compaction is used in consequential
settings. Keeping the full state in turn carries the privacy and retention obligations of whatever it contains.
The Evidence Inference articles are clinical-trial reports used only as a benchmark, and compaction and the
sufficiency check remain unvalidated for clinical decision support.

\bibliography{P4}
\bibliographystyle{tmlr}

\appendix
\section{Prespecified predictions and their outcomes}
\label{app:audit}

Ten analyses were specified before their first model request, with 51 predictions and decision rules in all.
Table~\ref{tab:audit} counts the outcomes by analysis; each prediction, its rule and its estimate are reported in
the appendix named in the table. Intervals are 95\% percentile bootstrap intervals, unadjusted for multiple
comparisons. The predictions that were not fully met are listed below the table.

\begin{table}[ht]
\centering\small
\setlength{\tabcolsep}{4pt}
\caption{Outcomes of the prespecified predictions. Partly met: met in some of the cells the prediction covers.}
\label{tab:audit}
\begin{tabular}{llrrrrr}
\toprule
analysis & appendix & predictions & met & partly met & not met & unresolved \\
\midrule
Evidence Inference overlap test & \ref{app:third} & 2 & 1 & --- & 1 & --- \\
KEEP--DROP contrast & \ref{app:retention} & 4 & 4 & --- & --- & --- \\
LoCoMo test & \ref{app:locomo} & 6 & 3 & --- & --- & 3 \\
Follow-ups A--D & \ref{app:followup}, \ref{app:support}, \ref{app:locomo} & 9 & 6 & 1 & 1 & 1 \\
\LLMLtwo{} granularity test & \ref{app:granularity} & 2 & 1 & 1 & --- & --- \\
Hybrid compactor & \ref{app:hybrid} & 3 & 1 & 2 & --- & --- \\
A different known question & \ref{app:known} & 4 & 4 & --- & --- & --- \\
General instruction-tuned reader & \ref{app:reader} & 2 & 2 & --- & --- & --- \\
$\tau$-bench test & \ref{app:tau} & 13 & 8 & --- & 5 & --- \\
Abstractive summaries & \ref{app:summary} & 6 & 6 & --- & --- & --- \\
\midrule
total & & 51 & 36 & 4 & 7 & 4 \\
\bottomrule
\end{tabular}
\end{table}

\paragraph{Not met.}
\begin{itemize}\setlength\itemsep{1pt}
\item Evidence Inference, H2: disjoint hidden claims were predicted not to gain from typed selection; they gain
\ci{\ctTFDisjointDelta}{\ctTFDisjointLo}{\ctTFDisjointHi} points.
\item Follow-up D0: harder LoCoMo options were to bring \texttt{gpt-6-luna}'s empty-memory accuracy to at most 60\%; it
is \fuDPooledHardAcc\%.
\item $\tau$-bench M2: expected-coverage selection was to lose at most 3 points on the known call by the 90\% lower end; ECov
$-$ typed is \taECovTVsTypedKDelta{} with lower end \taECovTVsTypedKLoNinety.
\item $\tau$-bench M3: expected-coverage selection was to close at least 0.70 of the random--oracle gap ($G \geq 0.70$); it closes
$G = \taGECovT$.
\item $\tau$-bench M4: embedding-based expected coverage was to stay within $\pm 3$ points of the Boolean version; it is
\taECovEVsECovTHDelta{} [\taECovEVsECovTHLoNinety, \taECovEVsECovTHHiNinety] (90\%).
\item $\tau$-bench M6: the recoverable gist arm was to reach 95\% of full-state accuracy at no more than 60\% of its
tokens; it reaches \taRecGistPtrAccB\% against \taFullItemAcc\% at \taRecGistPtrTokShareB\% of the tokens.
\item $\tau$-bench T7: the hybrid was to stay within 2 points of typed selection by the 90\% lower end; the lower end is
\taHybridVsTypedHLoNinety.
\end{itemize}

\paragraph{Partly met.}
\begin{itemize}\setlength\itemsep{1pt}
\item Follow-up B: LongLLMLingua and MMR beat random deletion on the known question in every cell except
LongLLMLingua with \texttt{open-decider} at 50\% (\fuMrcDecLongVsRandomFKnownDelta).
\item Granularity G2: random words beat random sentences with \texttt{gpt-6-luna} on both corpora but not with
\texttt{open-decider} (\ci{\grMrcDecGranRandomTFDelta}{\grMrcDecGranRandomTFLo}{\grMrcDecGranRandomTFHi}).
\item Hybrid H2 and H3: the hybrid beats \LLMLtwo{} on the known question, and stays within 2 points of it on hidden
questions, on FEVER only.
\end{itemize}

\paragraph{Unresolved.}
\begin{itemize}\setlength\itemsep{1pt}
\item LoCoMo H1: typed $-$ random at 25\% is \ci{\lcmHOneDelta}{\lcmHOneLo}{\lcmHOneHi} with \texttt{gpt-6-luna}
and \ci{\lcmHOneQwenDelta}{\lcmHOneQwenLo}{\lcmHOneQwenHi} with \texttt{qwen3.8-flash}; H3, which needs both
differences resolved, cannot be assessed.
\item Follow-up D1: with harder options typed $-$ random is
\ci{\fuDLunaTypedTFVsRandomTFAllDelta}{\fuDLunaTypedTFVsRandomTFAllLo}{\fuDLunaTypedTFVsRandomTFAllHi} and
\ci{\fuDQwenTypedTFVsRandomTFAllDelta}{\fuDQwenTypedTFVsRandomTFAllLo}{\fuDQwenTypedTFVsRandomTFAllHi}.
\end{itemize}

\section{Run details}
\label{app:runs}

\paragraph{Texts.} The analyses use, for each MultiRC paragraph, the original text and fifteen compactions
(the five sentence policies of Section~\ref{sec:setup} at three budgets), plus the additional random masks of Appendix~\ref{app:masks}.
\LLMLtwo{} texts, at 25\% and 50\%, are read separately (Appendix~\ref{app:lingua}).

\paragraph{Coverage and request size.} \texttt{open-decider}'s 75\% rows cover
\CpDecSFNParasLo--\CpDecSFNParasHi{} paragraphs and its other rows, like \texttt{gpt-6-luna}'s, cover \CpDecNParas; \texttt{open-bosun} reads
the first 100 paragraphs. The open-weight models are released as decider-2b, bosun-v3.1-1.7b and Lumma-fev-0.6b.
The Decisions API accepts at most 200 questions per request, so the \CpOaiNChunkedReq{} read requests above that
limit are split into chunks of \CpOaiChunkSize{} questions.

\paragraph{Repeat reads.} \texttt{open-decider} reads deterministically up to floating point. Each read is one forward pass in evaluation mode with a softmax over
option logits, no sampling and a fixed option order. Each question has its own row, unaffected by the others in the request. Repeated runs of hosted language models
can differ \citep{ouyang2025nondeterminism,atil2025nondeterminism}. For \texttt{gpt-6-luna}, across \GtOaiRepNItems{} repeated full-text reads, all \GtOaiRepNAnswers{} answers were identical. Its sufficiency checks take few distinct values, which is
why ties at the cut are re-read in proportion.

\paragraph{Averages.} Paragraph-macro averages weigh each paragraph equally, with answers averaged within it;
bootstrap intervals use 2000 resamples (macro-averaged accuracies and contrasts). The pooled flip-rate risk
differences of Section~\ref{sec:eraser} use 1000. The intervals of the sufficiency AUROC against flipped answers
and of the re-read gain over a random quarter (MultiRC and FEVER) use 500.

\paragraph{Selection details.} A budget $b$ keeps $\max(1, \mathrm{round}(bn))$ of a paragraph's $n$ sentences.
The score-based policies keep the highest-scoring sentences, ties going to the earlier sentence. TF-IDF selection
fits scikit-learn's \texttt{TfidfVectorizer} (English stop words, other settings default) on the paragraph's
sentences plus the known question and ranks sentences by cosine similarity to the question.

\paragraph{Refusals.} In the sentence-budget MultiRC reads, \CpOaiNRefused{} of \texttt{gpt-6-luna}'s answers came back as refusals
rather than probabilities; they came from \CpOaiNRefusedReq{} requests and were all candidate-answer Booleans.
We drop each refused answer from every statistic that needs it. Refusals are rare: under the five sentence policies at
25\% and 50\%, at most \rfMaxRate\% of a policy's known or hidden answers are refused (Table~\ref{tab:refusal}a). At
token-matched budgets the highest share in the follow-ups is LongLLMLingua's, \rfMaxRateTokens\% of hidden answers at 25\%; in
the granularity test random word deletion draws \grMrcOaiRandWordTFRefused\% at 25\%, so its contrasts rest on the answers
it receives. The open-weight readers never refuse.

\paragraph{Counting refusals as wrong changes no contrast's sign and one detection.} Table~\ref{tab:refusal}b
recomputes \texttt{gpt-6-luna}'s MultiRC contrasts with each refused answer scored as wrong, with the same paired
paragraph bootstrap and seeds. None of the \rfNContrasts{} contrasts changes sign. One changes detection: at a
token-matched 50\% budget, LongLLMLingua falls detectably below random deletion on hidden questions
(\rfTokensLongVsRandomFHiddenDelta{} [\rfTokensLongVsRandomFHiddenLo, \rfTokensLongVsRandomFHiddenHi] points,
against \rfTokensLongVsRandomFHiddenDropDelta{} [\rfTokensLongVsRandomFHiddenDropLo,
\rfTokensLongVsRandomFHiddenDropHi] with refusals dropped). The largest shift, \rfMaxShift{} points, is
LongLLMLingua's hidden-question contrast at 25\%, where it refuses most.

Typed $-$ random hidden accuracy at 25\% becomes \rfSingleTypedVsRandomTFHiddenDelta{} points with one random mask
and \rfMasksTypedVsRandomTFHiddenDelta{} averaged over masks (refusals dropped: \CpOaiKnownVsrandomTFHiddenDelta{}
and \rmMrcOaiKnownTFHiddenDelta). Known-question accuracy at 25\% is \rfRandomTFKnownWrong\% after random deletion
and \rfTypedTFKnownWrong\% after typed selection.
The same recomputation for FEVER and Evidence
Inference covers \rfOtherNContrasts{} contrasts: typed $-$ random with one mask and averaged over masks, and FEVER's
token-matched contrasts. None changes sign or detection, and the largest shift rounds to \rfOtherMaxShift{} points.

\begin{table}[tp]
\centering\footnotesize
\caption{Counting \texttt{gpt-6-luna}'s refused MultiRC answers as wrong changes no contrast's sign. (a) Refused
share of each policy's answers (\%) on the known and hidden questions. Token-matched texts are those of
Appendix~\ref{app:followup}; random deletion pools its masks. (b) The contrasts whose estimate moves by at least 0.5 points or whose detection changes when a refused answer
is scored as wrong instead of dropped from both policies (points, paired paragraph-bootstrap 95\% interval); no
contrast changes sign.}
\label{tab:refusal}
\input{generated/p4_refusal_table.tex}
\end{table}

\paragraph{Evidence details.} ERASER's evidence sentences were mapped to our segmentation by token overlap
(threshold 0.3). At 25\% on MultiRC typed selection keeps the full evidence set for
\ErDecKnownTFHiddenSurvive\% (\texttt{open-decider}) and \ErOaiKnownTFHiddenSurvive\% (\texttt{gpt-6-luna}) of
hidden questions (paragraph-macro; random \ErDecKnownTFRandHiddenSurvive\%). Under random deletion the right-to-wrong rates with
evidence intact and broken are \ErDecRandomTFFlipSurvived\% and \ErDecRandomTFFlipBroken\%
(\texttt{open-decider}) and \ErOaiRandomTFFlipSurvived\% and \ErOaiRandomTFFlipBroken\% (\texttt{gpt-6-luna}).

\paragraph{General instruction-tuned reader.} \label{app:reader}

\paragraph{Reader: plan.} The plan fixed the reader, texts and estimators and two predictions. \texttt{qwen3.8-flash}'s full-text MultiRC accuracy is above the all-``no'' rate, with the interval
above it (R1), and typed $-$ random on the known question at 25\% is positive on each corpus (R2); hidden contrasts and the overlap split are reported without predictions.

\paragraph{Reader: design.} \texttt{qwen3.8-flash} answers one yes/no question per call, with the paper's read
instructions; the probability of ``yes'' is taken from the first-token
log-probabilities of ``yes'' and ``no'' (thinking off) and thresholded at 0.5. It reads the paper's texts: the full
text and typed selection (\texttt{gpt-6-luna}'s selection), question-agnostic selection (MultiRC and
FEVER; Evidence Inference has no question-agnostic texts) and random deletion (the paper's mask) at 25\% and 50\% of the sentences, and on MultiRC and FEVER the
token-matched \LLMLtwo{} text at 25\%. In all, \rdQwenCalls{} calls; the provider's content filter refused the
answers on 18 MultiRC paragraphs and one FEVER document (0.9--2.8\% of MultiRC answers per text), which are dropped as refusals.
\texttt{gpt-6-luna} reads the empty text with the paper's read requests. Estimators are the paper's.

\paragraph{Reader: results.} Both predictions are met. The full-text accuracy of \texttt{qwen3.8-flash} is
\rdMrcQwenFullAcc\% [\rdMrcQwenFullLo, \rdMrcQwenFullHi] on MultiRC (all-``no'' \rdMrcAllNo\%), \rdFeverQwenFullAcc\%
on FEVER and \rdEiQwenFullAcc\% on Evidence Inference. Its contrasts follow those of \texttt{gpt-6-luna}
(Table~\ref{tab:reader}): on MultiRC typed selection helps the known question but not hidden questions on average, as overlapping
hidden questions gain and disjoint ones lose; on FEVER and Evidence Inference hidden claims gain, and on Evidence
Inference disjoint claims gain too. Question-agnostic $-$ random on hidden questions at 25\% is
\rdMrcAgnRandTFHiddenDelta{} [\rdMrcAgnRandTFHiddenLo, \rdMrcAgnRandTFHiddenHi] on MultiRC and
\rdFeverAgnRandTFHiddenDelta{} [\rdFeverAgnRandTFHiddenLo, \rdFeverAgnRandTFHiddenHi] on FEVER. The token-matched
\LLMLtwo{} text scores \rdMrcQwenLinguaTFHidden\% and \rdFeverQwenLinguaTFHidden\% on hidden questions, against
\rdMrcQwenTypedTFHidden\% and \rdFeverQwenTypedTFHidden\% for typed selection at 25\% of the sentences.

\begin{table}[t]
\centering\scriptsize
\setlength{\tabcolsep}{2pt}
\caption{\texttt{qwen3.8-flash} as reader: typed $-$ random accuracy (points, document-bootstrap 95\% interval) on
the known and hidden questions, with the hidden questions split by evidence overlap, at 25\% and (last column) 50\%.}
\label{tab:reader}
\begin{tabular}{lccccc}
\toprule
 & & \multicolumn{3}{c}{hidden questions, 25\%} & hidden \\
\cmidrule(lr){3-5}
corpus & known, 25\% & all & overlapping & disjoint & 50\% \\
\midrule
MultiRC & \ci{\rdMrcTypedRandTFKnownDelta}{\rdMrcTypedRandTFKnownLo}{\rdMrcTypedRandTFKnownHi} & \ci{\rdMrcTypedRandTFHiddenDelta}{\rdMrcTypedRandTFHiddenLo}{\rdMrcTypedRandTFHiddenHi} & \ci{\rdMrcTypedRandTFOverlapDelta}{\rdMrcTypedRandTFOverlapLo}{\rdMrcTypedRandTFOverlapHi} & \ci{\rdMrcTypedRandTFDisjointDelta}{\rdMrcTypedRandTFDisjointLo}{\rdMrcTypedRandTFDisjointHi} & \ci{\rdMrcTypedRandFHiddenDelta}{\rdMrcTypedRandFHiddenLo}{\rdMrcTypedRandFHiddenHi} \\
FEVER & \ci{\rdFeverTypedRandTFKnownDelta}{\rdFeverTypedRandTFKnownLo}{\rdFeverTypedRandTFKnownHi} & \ci{\rdFeverTypedRandTFHiddenDelta}{\rdFeverTypedRandTFHiddenLo}{\rdFeverTypedRandTFHiddenHi} & \ci{\rdFeverTypedRandTFOverlapDelta}{\rdFeverTypedRandTFOverlapLo}{\rdFeverTypedRandTFOverlapHi} & \ci{\rdFeverTypedRandTFDisjointDelta}{\rdFeverTypedRandTFDisjointLo}{\rdFeverTypedRandTFDisjointHi} & \ci{\rdFeverTypedRandFHiddenDelta}{\rdFeverTypedRandFHiddenLo}{\rdFeverTypedRandFHiddenHi} \\
Ev.\ Inference & \ci{\rdEiTypedRandTFKnownDelta}{\rdEiTypedRandTFKnownLo}{\rdEiTypedRandTFKnownHi} & \ci{\rdEiTypedRandTFHiddenDelta}{\rdEiTypedRandTFHiddenLo}{\rdEiTypedRandTFHiddenHi} & \ci{\rdEiTypedRandTFOverlapDelta}{\rdEiTypedRandTFOverlapLo}{\rdEiTypedRandTFOverlapHi} & \ci{\rdEiTypedRandTFDisjointDelta}{\rdEiTypedRandTFDisjointLo}{\rdEiTypedRandTFDisjointHi} & \ci{\rdEiTypedRandFHiddenDelta}{\rdEiTypedRandFHiddenLo}{\rdEiTypedRandFHiddenHi} \\
\bottomrule
\end{tabular}
\end{table}

\section{Random deletion over several masks}
\label{app:masks}

The main text compares each policy with one seeded random mask per paragraph and budget. Here every contrast
with random is recomputed against the mean over several seeded masks: \rmMrcOaiMasks{} per paragraph for
\texttt{gpt-6-luna} on MultiRC and FEVER, \rmMrcDecMasks{} for \texttt{open-decider} and \rmEiMasks{} on
Evidence Inference, the main mask included. A document's contrast is the mean over its masks of the paired
policy $-$ random difference, and the bootstrap resamples documents (2000 resamples) and keeps each document's
mask average. Table~\ref{tab:masks} gives the hidden-question contrasts. On MultiRC the 90\% intervals of all \rmMrcNTests{}
hidden-question contrasts with random (the eight of Table~\ref{tab:masks} and four from \texttt{open-decider}'s
re-read) have a largest absolute end of \rmMrcMaxBound{} points, and typed selection still beats random on the known question
(\texttt{gpt-6-luna} \rmMrcOaiKnownTFKnownDelta{} [\rmMrcOaiKnownTFKnownLo, \rmMrcOaiKnownTFKnownHi] and
\texttt{open-decider} \rmMrcDecKnownTFKnownDelta{} [\rmMrcDecKnownTFKnownLo, \rmMrcDecKnownTFKnownHi] points at
25\%). Across single
masks, \texttt{gpt-6-luna}'s typed $-$ random on hidden MultiRC questions at 25\% ranges from
\rmMrcOaiKnownSpreadTFMin{} to \rmMrcOaiKnownSpreadTFMax{} points, and the main-text mask gives the lowest
value. Two cells differ from the one-mask results: the MultiRC disjoint contrast excludes zero in all four
reader-budget cells, and the FEVER disjoint contrast at 50\% is below random. FEVER agnostic $-$ random still spans zero at both budgets,
and on Evidence Inference H1 remains supported and H2 contradicted. Repeated reads are stable: re-reading the main-text items puts \rmDriftDecSameSide\% of \texttt{open-decider}'s
answers on the same side of 0.5, with contrasts on the re-read items within 0.4 points of Table~\ref{tab:masks}, and a
repeat of 60 MultiRC paragraphs puts \rmDriftOaiSameSide\% of \texttt{gpt-6-luna}'s answers (to one decimal) on the same side.

\begin{table}[t]
\centering\footnotesize
\setlength{\tabcolsep}{3pt}
\caption{Hidden-question contrasts against random deletion averaged over seeded masks (points, document
bootstrap 95\% interval), at budget $b$ (\%), over all hidden questions and split by evidence overlap with the
known question. Compare Table~\ref{tab:overlap}, which uses one mask.}
\label{tab:masks}
\begin{tabular}{llrlll}
\toprule
corpus, reader & contrast & $b$ & all hidden & overlapping & disjoint \\
\midrule
MultiRC, \texttt{open-decider} & typed $-$ random & 25 & \ci{\rmMrcDecKnownTFHiddenDelta}{\rmMrcDecKnownTFHiddenLo}{\rmMrcDecKnownTFHiddenHi} & \ci{\rmMrcDecKnownTFOverlapDelta}{\rmMrcDecKnownTFOverlapLo}{\rmMrcDecKnownTFOverlapHi} & \ci{\rmMrcDecKnownTFDisjointDelta}{\rmMrcDecKnownTFDisjointLo}{\rmMrcDecKnownTFDisjointHi} \\
 & & 50 & \ci{\rmMrcDecKnownFHiddenDelta}{\rmMrcDecKnownFHiddenLo}{\rmMrcDecKnownFHiddenHi} & \ci{\rmMrcDecKnownFOverlapDelta}{\rmMrcDecKnownFOverlapLo}{\rmMrcDecKnownFOverlapHi} & \ci{\rmMrcDecKnownFDisjointDelta}{\rmMrcDecKnownFDisjointLo}{\rmMrcDecKnownFDisjointHi} \\
 & agnostic $-$ random & 25 & \ci{\rmMrcDecAgnosticTFHiddenDelta}{\rmMrcDecAgnosticTFHiddenLo}{\rmMrcDecAgnosticTFHiddenHi} & & \\
 & & 50 & \ci{\rmMrcDecAgnosticFHiddenDelta}{\rmMrcDecAgnosticFHiddenLo}{\rmMrcDecAgnosticFHiddenHi} & & \\
MultiRC, \texttt{gpt-6-luna} & typed $-$ random & 25 & \ci{\rmMrcOaiKnownTFHiddenDelta}{\rmMrcOaiKnownTFHiddenLo}{\rmMrcOaiKnownTFHiddenHi} & \ci{\rmMrcOaiKnownTFOverlapDelta}{\rmMrcOaiKnownTFOverlapLo}{\rmMrcOaiKnownTFOverlapHi} & \ci{\rmMrcOaiKnownTFDisjointDelta}{\rmMrcOaiKnownTFDisjointLo}{\rmMrcOaiKnownTFDisjointHi} \\
 & & 50 & \ci{\rmMrcOaiKnownFHiddenDelta}{\rmMrcOaiKnownFHiddenLo}{\rmMrcOaiKnownFHiddenHi} & \ci{\rmMrcOaiKnownFOverlapDelta}{\rmMrcOaiKnownFOverlapLo}{\rmMrcOaiKnownFOverlapHi} & \ci{\rmMrcOaiKnownFDisjointDelta}{\rmMrcOaiKnownFDisjointLo}{\rmMrcOaiKnownFDisjointHi} \\
 & agnostic $-$ random & 25 & \ci{\rmMrcOaiAgnosticTFHiddenDelta}{\rmMrcOaiAgnosticTFHiddenLo}{\rmMrcOaiAgnosticTFHiddenHi} & & \\
 & & 50 & \ci{\rmMrcOaiAgnosticFHiddenDelta}{\rmMrcOaiAgnosticFHiddenLo}{\rmMrcOaiAgnosticFHiddenHi} & & \\
FEVER, \texttt{gpt-6-luna} & typed $-$ random & 25 & \ci{\rmFeverKnownTFHiddenDelta}{\rmFeverKnownTFHiddenLo}{\rmFeverKnownTFHiddenHi} & \ci{\rmFeverKnownTFOverlapDelta}{\rmFeverKnownTFOverlapLo}{\rmFeverKnownTFOverlapHi} & \ci{\rmFeverKnownTFDisjointDelta}{\rmFeverKnownTFDisjointLo}{\rmFeverKnownTFDisjointHi} \\
 & & 50 & \ci{\rmFeverKnownFHiddenDelta}{\rmFeverKnownFHiddenLo}{\rmFeverKnownFHiddenHi} & \ci{\rmFeverKnownFOverlapDelta}{\rmFeverKnownFOverlapLo}{\rmFeverKnownFOverlapHi} & \ci{\rmFeverKnownFDisjointDelta}{\rmFeverKnownFDisjointLo}{\rmFeverKnownFDisjointHi} \\
 & agnostic $-$ random & 25 & \ci{\rmFeverAgnosticTFHiddenDelta}{\rmFeverAgnosticTFHiddenLo}{\rmFeverAgnosticTFHiddenHi} & & \\
 & & 50 & \ci{\rmFeverAgnosticFHiddenDelta}{\rmFeverAgnosticFHiddenLo}{\rmFeverAgnosticFHiddenHi} & & \\
Ev.\ Inference, \texttt{gpt-6-luna} & typed $-$ random & 25 & \ci{\rmEiKnownTFHiddenDelta}{\rmEiKnownTFHiddenLo}{\rmEiKnownTFHiddenHi} & \ci{\rmEiKnownTFOverlapDelta}{\rmEiKnownTFOverlapLo}{\rmEiKnownTFOverlapHi} & \ci{\rmEiKnownTFDisjointDelta}{\rmEiKnownTFDisjointLo}{\rmEiKnownTFDisjointHi} \\
 & & 50 & \ci{\rmEiKnownFHiddenDelta}{\rmEiKnownFHiddenLo}{\rmEiKnownFHiddenHi} & \ci{\rmEiKnownFOverlapDelta}{\rmEiKnownFOverlapLo}{\rmEiKnownFOverlapHi} & \ci{\rmEiKnownFDisjointDelta}{\rmEiKnownFDisjointLo}{\rmEiKnownFDisjointHi} \\
\bottomrule
\end{tabular}
\end{table}

\section{Deleted evidence: the restore experiment and the sufficiency check}
\label{app:restore}

\paragraph{Answers flip more often when evidence was deleted.} Of the hidden answers the reader got right on the
full text, under typed selection at 25\% \texttt{open-decider} flips \ErDecKnownTFFlipSurvived\% when their evidence
is intact and \ErDecKnownTFFlipBroken\% when it is broken, a difference of \ErDecKnownTFRiskDiff{}
[\ErDecKnownTFRiskLo, \ErDecKnownTFRiskHi] points; for \texttt{gpt-6-luna} the rates are \ErOaiKnownTFFlipSurvived\%
and \ErOaiKnownTFFlipBroken\% (\ErOaiKnownTFRiskDiff{} [\ErOaiKnownTFRiskLo, \ErOaiKnownTFRiskHi]).

\texttt{gpt-6-luna} reads all texts of this experiment. The restore and placebo texts keep the compacted sentences and add the chosen ones, in document order. Each is read with the paragraph's original question set, split into requests of at most 125 questions as in that
read, so only the state differs from the typed-selection 25\% read. Recovery rates and accuracies are pooled over
candidate answers of broken-evidence hidden questions, not paragraph-macro averages; intervals are paragraph
bootstraps (2000 resamples) of the restore $-$ placebo recovery difference.

\paragraph{Design and results.} On each eligible paragraph let $m$ be the number of deleted sentences that are
evidence for no question and $U$ the deleted evidence of the broken hidden questions; the dose is $d=\min(m,|U|)$,
and paragraphs with $m=0$ are skipped (\RsConcSkipped{} of \RsConcEligible). Restore adds $d$ sentences drawn from
$U$ (seeded per paragraph) and the placebo $d$ non-evidence sentences, so in \RsDoseNewParas{} paragraphs restore adds
only part of the deleted evidence. Restore and placebo reads are interleaved in time. Over \RsConcAnswers{} answers
(the \RsConcExcluded{} answers refused in any of the full, compacted, restore and placebo reads are excluded),
accuracy goes from \RsConcAccCompact{} compacted to \RsConcAccRestore{} with restore and \RsConcAccPlacebo{} with the
placebo (full text \RsConcAccFull). Of the \RsConcFlips{} broken-evidence flips, restore undoes \RsConcRecovered{}
and the placebo \RsConcRecoveredPlacebo{} (+\RsConcDiff{} [+\RsConcDiffLo, +\RsConcDiffHi]). On the
\RsConcAllBackFlips{} flips whose question got all its deleted evidence back, the rates are \RsConcAllBackRecovered{}
and \RsConcAllBackRecoveredPlacebo{} (+\RsConcAllBackDiff{} [+\RsConcAllBackDiffLo, +\RsConcAllBackDiffHi]), and on
the other \RsConcPartialFlips{}, \RsConcPartialRecovered{} and \RsConcPartialRecoveredPlacebo. Adding all of the
deleted evidence (\RsMeanAdded{} sentences on average) against a placebo of as many non-evidence sentences where
available (\RsPlaceboMeanAdded{} on average) gives +\RsRecoveredDiff{} [+\RsDiffLo, +\RsDiffHi] points over
\RsParas{} paragraphs.

\paragraph{The sufficiency check.} \label{app:gate}

\begin{table}[t]
\centering\footnotesize
\setlength{\tabcolsep}{2.7pt}
\caption{How well the sufficiency check finds damaged hidden questions, and what re-reading them recovers (MultiRC,
the rows of Table~\ref{tab:open}). The AUROC (area under the receiver operating characteristic curve) columns rank hidden questions by low sufficiency and score that ranking
against two question-level outcomes: at least one annotated evidence sentence deleted (evidence lost), and at least
one candidate answer turned from right to wrong (answer flipped). The last two columns give hidden candidate-level
accuracy averaged over questions (not exact match) after re-reading the full text for 25\% of hidden questions:
those with the lowest sufficiency, or a random 25\%.}
\label{tab:gate}
\input{generated/compress_open_gate.tex}
\end{table}

Every read asks the sufficiency check. We rank the hidden questions by low sufficiency and score the ranking
against two outcomes: at least one of the question's annotated evidence sentences was deleted, and at least one
of its candidate answers that was correct on the full text is wrong on the compacted one. These are
question-level averages, unlike the paragraph-macro accuracies of the main tables. The re-read simulation
gives the lowest-scoring fraction of questions the answers the same reader gave on the full text and compares
hidden accuracy with substituting a random share of the same size (paragraph-bootstrap interval on the
difference). This equals a real re-read when a second full-text read returns the recorded answer
(Appendix~\ref{app:runs} reports repeated reads). Questions tied at the cut are re-read in proportion (the
expectation of a random tie-break).

Under typed selection at 25\%, the AUROCs against deleted evidence and against flipped answers are \GtDecKnownTFAurocBroken{} and \GtDecKnownTFAurocFlip{} for \texttt{open-decider}, and \GtOaiKnownTFAurocBroken{} and \GtOaiKnownTFAurocFlip{} for \texttt{gpt-6-luna}. The gap between the two AUROCs comes from the reader: many questions whose evidence is broken are still answered
correctly. Most of the recovered accuracy comes from re-reading at all: for \texttt{gpt-6-luna}, re-reading the
lowest-sufficiency quarter lifts hidden accuracy (candidate-level, averaged over questions) from
\GtOaiKnownTFAccCompact\% to \GtOaiKnownTFGatedTwentyFive\%, and a random quarter to \GtOaiKnownTFRandomTwentyFive\%
(\texttt{open-decider}: \GtDecKnownTFAccCompact\%, \GtDecKnownTFGatedTwentyFive\% and \GtDecKnownTFRandomTwentyFive\%).
Choosing by the check thus adds \GtOaiKnownTFGainTwentyFive{} [\GtOaiKnownTFGainTwentyFiveLo,
\GtOaiKnownTFGainTwentyFiveHi] and \GtDecKnownTFGainTwentyFive{} [\GtDecKnownTFGainTwentyFiveLo,
\GtDecKnownTFGainTwentyFiveHi] points over a random quarter. The 95\% intervals of the AUROCs against flipped answers are [\GtDecKnownTFAurocFlipLo, \GtDecKnownTFAurocFlipHi]
(\texttt{open-decider}) and [\GtOaiKnownTFAurocFlipLo, \GtOaiKnownTFAurocFlipHi] (\texttt{gpt-6-luna}). Under typed selection at 25\%,
\GtDecKnownTFBrokenRate\% (\texttt{open-decider}) and \GtOaiKnownTFBrokenRate\% (\texttt{gpt-6-luna}) of hidden questions lost at
least one evidence sentence (question-pooled). Re-reading half the questions gains \GtDecKnownTFGainFifty{} and
\GtOaiKnownTFGainFifty{} points over a random half and lifts that accuracy to \GtDecKnownTFGatedFifty\% and
\GtOaiKnownTFGatedFifty\% (full text \GtDecKnownTFAccOrig\% and \GtOaiKnownTFAccOrig\%).

\section{FEVER and Evidence Inference: data and the overlap split}
\paragraph{FEVER.} \label{app:fever}

At 25\% typed selection's known-claim accuracy is \cfKnownTFKnown\% (full text \cfOrigKnown\%, random
\cfRandomTFKnown\%). Agnostic $-$ random on the hidden claims is \cfAgnosticRandomTFHiddenDelta{}
[\cfAgnosticRandomTFHiddenLo, \cfAgnosticRandomTFHiddenHi] points at 25\% and \cfAgnosticRandomFHiddenDelta{}
[\cfAgnosticRandomFHiddenLo, \cfAgnosticRandomFHiddenHi] at 50\%. At 25\%, hidden claims judged correctly on the full text flip at \cfKnownTFFlipBroken\% when their evidence
was broken and \cfKnownTFFlipSurvived\% when it survived. Unlike on MultiRC, the sufficiency check ranks flipped
claims better (AUROC \cfKnownTFAurocFlip{} [\cfKnownTFAurocFlipLo, \cfKnownTFAurocFlipHi]) than deleted evidence
(\cfKnownTFAurocBroken), and re-reading its lowest quarter lifts question-level accuracy from
\cfKnownTFAccCompact\% to \cfKnownTFGatedTwentyFive\% (random \cfKnownTFRandomTwentyFive\%), a gain of
\cfKnownTFGainTwentyFive{} [\cfKnownTFGainTwentyFiveLo, \cfKnownTFGainTwentyFiveHi] points over a random quarter.

Each read asks, per claim, ``given the passage, is this claim supported?'' and the sufficiency check.
The \cfNDocs{} documents (\cfMeanSent{} sentences and \cfMeanClaims{} claims on average, one sentence per line)
are a seeded sample of the ERASER FEVER documents with
\poFeverMinSent--\poFeverMaxSent{} sentences and no empty line that carry at least \poFeverMinClaims{} distinct
claims. All but \prtLabFeverNConflicting{} claims have a single label (SUPPORTS or REFUTES) and sentence-level evidence lying entirely in that
document (\cfSupportsShare\% of claims are SUPPORTS). A claim's evidence is the union of its annotated rationale groups.
Claims keep the dataset's file order. Read requests carry at most \poFeverClaimsPerRequest{} claims (\poFeverMaxQuestions{} questions, the API limit).
For the reported conditions \texttt{gpt-6-luna} answered \cfNScoreRequests{} scoring requests
(\cfNScoreAnswers{} sentence Booleans, \cfNScoreRefused{} refused) and \cfNCoreReadRequests{} read requests
(\cfNCoreClaimAnswers{} claim answers and \cfNCoreSuffAnswers{} sufficiency checks, \cfNCoreRefused{}
refused), with \cfCoreInputTokens{} input tokens. The data are available
under the license terms of the Wikipedia article pages or, where unavailable, CC BY-SA 3.0.

\paragraph{Evidence Inference.} \label{app:third}

Before any request of this experiment, we specified the corpus,
the eligibility and sampling rules, a sample-size rule (the longest prefix of a seeded article order whose
expected spend fits a fixed budget), the prompts, policies and budgets, the overlap definition, both hypotheses
with their directions, tests and level, the estimand and the bootstrap seed, and which outcomes are exploratory, among them
the evidence retention of Table~\ref{tab:third}.

The data are ERASER's release \citep{deyoung2020eraser} of Evidence Inference \citep{lehman2019inferring}:
full-text clinical-trial articles, one sentence per line, with prompts naming an intervention $I$, a comparator
$C$ and an outcome $O$, a three-way label and annotated evidence sentences. Each prompt becomes the claim
``Compared with $C$, $I$ produced a statistically significant difference in $O$.'' The claim is true when the label is a
significant increase or decrease and false for no significant difference (\ctTrueShare\% of the \ctNClaims{}
claims are true; \ctMeanSent{} sentences and \ctMeanClaims{} claims per article on average). The first prompt in file order is known; of the
\ctNHidden{} hidden claims, \ctNOverlap{} overlap the known claim's evidence and \ctNDisjoint{} do not. Scoring
and reading prompts are those of FEVER with the claim text substituted. \texttt{gpt-6-luna} answered
\ctNRequests{} requests (\ctInputTokens{} input tokens, \$\ctCostUSD) with \ctNAnswers{} read answers, of which
\ctNRefused{} was a refusal.

Both hypotheses are one-sided tests at nominal level 0.05 on 2000 article-bootstrap means of the paired typed $-$
random difference, each decided by the share of means on the wrong side of zero, with 50\% as a secondary
replication. H1 (overlapping claims, difference $> 0$) is \ctHOneTFVerdict{} at both budgets (\ctTFOverlapDelta{}
points at 25\%, \ctFOverlapDelta{} at 50\%). H2 (disjoint claims, difference $\le 0$) is \ctHTwoTFVerdict{} at both
budgets: disjoint claims gain \ctTFDisjointDelta{} [\ctTFDisjointLo, \ctTFDisjointHi] points at 25\% and
\ctFDisjointDelta{} [\ctFDisjointLo, \ctFDisjointHi] at 50\%, and the share of bootstrap means at or below zero is
$\ctTFDisjointPPos$ at each. Table~\ref{tab:overlap} gives the exploratory unsplit contrast. Hidden accuracy at 25\% is
\ctKnownTFHiddenOverlap\% (overlapping) and \ctKnownTFHiddenDisjoint\% (disjoint) under typed selection,
\ctRandomTFHiddenOverlap\% and \ctRandomTFHiddenDisjoint\% under random deletion, and \ctOrigHiddenOverlap\% and
\ctOrigHiddenDisjoint\% on the full text; the known claim is answered at \ctKnownTFKnown\%, \ctRandomTFKnown\% and
\ctOrigKnown\%.

\begin{table}[t]
\centering\footnotesize
\setlength{\tabcolsep}{2.5pt}
\caption{Hidden questions whose evidence overlaps the known question's gain more from typed selection. Typed $-$
random hidden accuracy (points, paragraph or document bootstrap 95\% interval) at budget $b$ (\%), over all hidden
questions and split by evidence overlap with the known question. Overlap share: evidence-classified hidden
questions sharing an evidence sentence with the known question (\%). Disjoint kept: share of disjoint questions'
evidence kept by typed selection and by random deletion (\%; for MultiRC and FEVER the exact expectation for a
random subset of the same size, for Evidence Inference the kept set of the random-deletion read). Each accuracy
cell averages within documents, over the documents that have such questions, so the two cells need not combine to
the all-question contrast by the overlap share. The overlap split is post hoc on MultiRC and FEVER;
$^\dagger$on Evidence Inference the overlapping and disjoint contrasts are prespecified, and the all-hidden contrast
and the kept column exploratory.}
\label{tab:overlap}\label{tab:third}
\begin{tabular}{lrlllrr}
\toprule
 & & \multicolumn{3}{c}{typed $-$ random} & overlap & disjoint kept \\
\cmidrule(lr){3-5}
corpus, reader & $b$ & all hidden & overlapping & disjoint & share & typed/random \\
\midrule
MultiRC, \texttt{open-decider} & 25 & \ci{\CpDecKnownVsrandomTFHiddenDelta}{\CpDecKnownVsrandomTFHiddenLo}{\CpDecKnownVsrandomTFHiddenHi} & \ci{\poDecTFOverlapDelta}{\poDecTFOverlapLo}{\poDecTFOverlapHi} & \ci{\poDecTFDisjointDelta}{\poDecTFDisjointLo}{\poDecTFDisjointHi} & \poOverlapShare & \poDecTFDisjointRecall/\poDecTFDisjointRandRecall \\
 & 50 & \ci{\CpDecKnownVsrandomFHiddenDelta}{\CpDecKnownVsrandomFHiddenLo}{\CpDecKnownVsrandomFHiddenHi} & \ci{\poDecFOverlapDelta}{\poDecFOverlapLo}{\poDecFOverlapHi} & \ci{\poDecFDisjointDelta}{\poDecFDisjointLo}{\poDecFDisjointHi} &  & --- \\
MultiRC, \texttt{gpt-6-luna} & 25 & \ci{\CpOaiKnownVsrandomTFHiddenDelta}{\CpOaiKnownVsrandomTFHiddenLo}{\CpOaiKnownVsrandomTFHiddenHi} & \ci{\poOaiTFOverlapDelta}{\poOaiTFOverlapLo}{\poOaiTFOverlapHi} & \ci{\poOaiTFDisjointDelta}{\poOaiTFDisjointLo}{\poOaiTFDisjointHi} & \poOverlapShare & \poOaiTFDisjointRecall/\poOaiTFDisjointRandRecall \\
 & 50 & \ci{\CpOaiKnownVsrandomFHiddenDelta}{\CpOaiKnownVsrandomFHiddenLo}{\CpOaiKnownVsrandomFHiddenHi} & \ci{\poOaiFOverlapDelta}{\poOaiFOverlapLo}{\poOaiFOverlapHi} & \ci{\poOaiFDisjointDelta}{\poOaiFDisjointLo}{\poOaiFDisjointHi} &  & --- \\
FEVER, \texttt{gpt-6-luna} & 25 & \ci{\cfKnownRandomTFHiddenDelta}{\cfKnownRandomTFHiddenLo}{\cfKnownRandomTFHiddenHi} & \ci{\poFeverTFOverlapDelta}{\poFeverTFOverlapLo}{\poFeverTFOverlapHi} & \ci{\cfKnownRandomTFHiddenDisjointDelta}{\cfKnownRandomTFHiddenDisjointLo}{\cfKnownRandomTFHiddenDisjointHi} & \poFeverOverlapShare & \poFeverTFDisjointRecall/\poFeverTFDisjointRandRecall \\
 & 50 & \ci{\cfKnownRandomFHiddenDelta}{\cfKnownRandomFHiddenLo}{\cfKnownRandomFHiddenHi} & \ci{\poFeverFOverlapDelta}{\poFeverFOverlapLo}{\poFeverFOverlapHi} & \ci{\cfKnownRandomFHiddenDisjointDelta}{\cfKnownRandomFHiddenDisjointLo}{\cfKnownRandomFHiddenDisjointHi} & & --- \\
Ev.\ Inference, \texttt{gpt-6-luna}$^\dagger$ & 25 & \ci{\ctTFAllDelta}{\ctTFAllLo}{\ctTFAllHi} & \ci{\ctTFOverlapDelta}{\ctTFOverlapLo}{\ctTFOverlapHi} & \ci{\ctTFDisjointDelta}{\ctTFDisjointLo}{\ctTFDisjointHi} & \ctOverlapShare & \ctKnownTFRecallDisjoint/\ctRandomTFRecallDisjoint \\
 & 50 & \ci{\ctFAllDelta}{\ctFAllLo}{\ctFAllHi} & \ci{\ctFOverlapDelta}{\ctFOverlapLo}{\ctFOverlapHi} & \ci{\ctFDisjointDelta}{\ctFDisjointLo}{\ctFDisjointHi} & & \ctKnownFRecallDisjoint/\ctRandomFRecallDisjoint \\
\bottomrule
\end{tabular}
\end{table}

\paragraph{Evidence kept (exploratory).} On the overlapping questions typed selection keeps more evidence than
random, and they gain, on every corpus. Of the disjoint questions' evidence (Table~\ref{tab:third}, last column) it
keeps far more than random on Evidence Inference, and they gain; about as much on FEVER, with no detectable gain at
25\%; and less on MultiRC, where they lose accuracy. The gain on overlapping questions is also larger on FEVER, so the
overlap share is not the only difference behind the main contrast between MultiRC and FEVER. At 25\% typed selection
keeps, of the overlapping questions' evidence, \poDecTFOverlapRecall\% and \poOaiTFOverlapRecall\% on MultiRC
(\texttt{open-decider}, \texttt{gpt-6-luna}), \poFeverTFOverlapRecall\% on FEVER and \ctKnownTFRecallOverlap\%
on Evidence Inference, against \poDecTFOverlapRandRecall\%, \poFeverTFOverlapRandRecall\% and
\ctRandomTFRecallOverlap\% for random; of the disjoint questions' evidence it keeps the shares in
Table~\ref{tab:third}. Random is the expectation for a uniform subset of the same size on MultiRC and FEVER,
and the random policy's kept set on Evidence Inference. At 50\% on Evidence Inference it keeps \ctKnownFRecallOverlap\% and
\ctKnownFRecallDisjoint\% (random \ctRandomFRecallOverlap\% and \ctRandomFRecallDisjoint\%).

\section{A different known question}
\label{app:known}

\begin{figure}[t]
\centering
\includegraphics[width=0.5\linewidth]{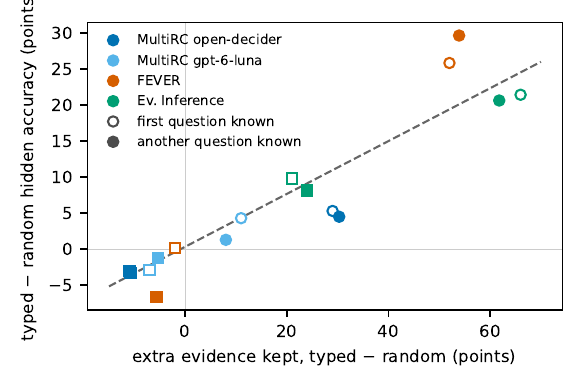}
\caption{For typed sentence selection, the hidden-question gain follows the extra evidence kept. Typed $-$ random hidden accuracy against typed
minus random evidence kept (points), at 25\%, on overlapping (circles) and disjoint (squares) hidden questions.
Open markers: the paper's known question (Figure~\ref{fig:overlap}); filled: a different known question per document. Dashed: the line specified before the second experiment, $0.366 \times$ extra evidence $+\,0.343$.}
\label{fig:line}
\end{figure}

\paragraph{Plan.} The plan fixed the choice of the new known question, the policies, readers, budgets and estimators, and four predictions at 25\%. Typed beats random deletion on the known question in every corpus--reader cell, with
the interval above 0 (K1). On hidden MultiRC questions typed $-$ random is within $\pm$\pqMargin{} points by the 90\%
interval, for each reader (K2). Overlapping hidden questions gain more than disjoint ones in every cell, by the
estimate (K3). Finally, the line $0.366 \times$ extra evidence $+\,0.343$ fits the new groups: a line fitted
to them has slope above 0, and the fixed line's mean absolute error is below 4 points (K4).

\paragraph{Design.} In each document the new known question is drawn uniformly, with a seed fixed per document, from
the eligible questions other than the first listed one; all other questions, the first one included, are hidden. On
FEVER the \prtLabFeverNConflicting{} claims listed with both labels are not eligible, as for hidden claims; the plan does
not state this exclusion, and every document has an alternative.
This gives \kqMrcNDocs{} MultiRC paragraphs, \kqFeverNDocs{} FEVER documents and \kqEiNDocs{} Evidence Inference
articles. Typed selection is scored again with the new known question and read again at 25\% and 50\%, with each
MultiRC reader scoring for itself. Random deletion keeps the paper's masks and reads, since its texts do not depend
on the known question and each read answers every question; only the overlap labels change. Estimators are the
paper's, with 2000 document resamples. Repeated reads keep \rmDriftDecSameSide\% of \texttt{open-decider}'s and \rmDriftOaiSameSide\% of \texttt{gpt-6-luna}'s
answers on the same side of 0.5 (Appendix~\ref{app:masks}).

\paragraph{Results.} All four predictions are met (Table~\ref{tab:known}); the 90\% intervals of K2 are
[\kqMrcOaiTFHiddenLoNinety, \kqMrcOaiTFHiddenHiNinety] (\texttt{gpt-6-luna}) and [\kqMrcDecTFHiddenLoNinety, \kqMrcDecTFHiddenHiNinety] (\texttt{open-decider}). The line fitted to the new cells has slope
\kqSlope{} [\kqSlopeLo, \kqSlopeHi] and intercept \kqIntercept. The largest misses are MultiRC overlapping questions with \texttt{open-decider}
(predicted \kqLineMrcDecOverlapPred, observed \kqLineMrcDecOverlapObs) and FEVER overlapping claims (predicted
\kqLineFeverOverlapPred, observed \kqLineFeverOverlapObs). On the hidden questions that are hidden under both known
questions, typed $-$ random at 25\% is \kqMrcOaiTFCommonDelta{} (MultiRC, \texttt{gpt-6-luna}),
\kqMrcDecTFCommonDelta{} (MultiRC, \texttt{open-decider}), \kqFeverTFCommonDelta{} (FEVER) and
\kqEiTFCommonDelta{} (Evidence Inference) points. At 50\%, typed $-$ random on the known question is
\kqMrcOaiFKnownDelta{} [\kqMrcOaiFKnownLo, \kqMrcOaiFKnownHi] and on hidden questions
\kqMrcOaiFHiddenDelta{} [\kqMrcOaiFHiddenLo, \kqMrcOaiFHiddenHi] on MultiRC (\texttt{gpt-6-luna}).

\begin{table}[t]
\centering\scriptsize
\setlength{\tabcolsep}{3pt}
\caption{With a different known question, typed selection still helps the known question, hidden MultiRC questions
stay near random deletion, and overlapping hidden questions gain more than disjoint ones. Typed $-$ random accuracy
at 25\% (points, document-bootstrap 95\% interval). Overlapping $-$ disjoint is the paired contrast within the documents that have
both kinds of hidden question. Bottom: extra evidence kept (typed $-$ random, points) and the gain predicted by the line specified before this
experiment, $0.366 \times$ extra evidence $+\,0.343$; the observed gains are the overlapping and disjoint columns above.}
\label{tab:known}
\setlength{\tabcolsep}{1.6pt}
\begin{tabular}{lccccc}
\toprule
 & & \multicolumn{3}{c}{hidden questions} & overlapping \\
\cmidrule(lr){3-5}
corpus, reader & known & all & overlapping & disjoint & $-$ disjoint \\
\midrule
MultiRC, \texttt{open-decider} & \ci{\kqMrcDecTFKnownDelta}{\kqMrcDecTFKnownLo}{\kqMrcDecTFKnownHi} & \ci{\kqMrcDecTFHiddenDelta}{\kqMrcDecTFHiddenLo}{\kqMrcDecTFHiddenHi} & \ci{\kqMrcDecTFOverlapDelta}{\kqMrcDecTFOverlapLo}{\kqMrcDecTFOverlapHi} & \ci{\kqMrcDecTFDisjointDelta}{\kqMrcDecTFDisjointLo}{\kqMrcDecTFDisjointHi} & \ci{\kqMrcDecTFInteractionDelta}{\kqMrcDecTFInteractionLo}{\kqMrcDecTFInteractionHi} \\
MultiRC, \texttt{gpt-6-luna} & \ci{\kqMrcOaiTFKnownDelta}{\kqMrcOaiTFKnownLo}{\kqMrcOaiTFKnownHi} & \ci{\kqMrcOaiTFHiddenDelta}{\kqMrcOaiTFHiddenLo}{\kqMrcOaiTFHiddenHi} & \ci{\kqMrcOaiTFOverlapDelta}{\kqMrcOaiTFOverlapLo}{\kqMrcOaiTFOverlapHi} & \ci{\kqMrcOaiTFDisjointDelta}{\kqMrcOaiTFDisjointLo}{\kqMrcOaiTFDisjointHi} & \ci{\kqMrcOaiTFInteractionDelta}{\kqMrcOaiTFInteractionLo}{\kqMrcOaiTFInteractionHi} \\
FEVER, \texttt{gpt-6-luna} & \ci{\kqFeverTFKnownDelta}{\kqFeverTFKnownLo}{\kqFeverTFKnownHi} & \ci{\kqFeverTFHiddenDelta}{\kqFeverTFHiddenLo}{\kqFeverTFHiddenHi} & \ci{\kqFeverTFOverlapDelta}{\kqFeverTFOverlapLo}{\kqFeverTFOverlapHi} & \ci{\kqFeverTFDisjointDelta}{\kqFeverTFDisjointLo}{\kqFeverTFDisjointHi} & \ci{\kqFeverTFInteractionDelta}{\kqFeverTFInteractionLo}{\kqFeverTFInteractionHi} \\
Ev.\ Inference, \texttt{gpt-6-luna} & \ci{\kqEiTFKnownDelta}{\kqEiTFKnownLo}{\kqEiTFKnownHi} & \ci{\kqEiTFHiddenDelta}{\kqEiTFHiddenLo}{\kqEiTFHiddenHi} & \ci{\kqEiTFOverlapDelta}{\kqEiTFOverlapLo}{\kqEiTFOverlapHi} & \ci{\kqEiTFDisjointDelta}{\kqEiTFDisjointLo}{\kqEiTFDisjointHi} & \ci{\kqEiTFInteractionDelta}{\kqEiTFInteractionLo}{\kqEiTFInteractionHi} \\
\bottomrule
\end{tabular}
\medskip
{\footnotesize
\label{tab:known-line}
\begin{tabular}{lrrrr}
\toprule
& \multicolumn{2}{c}{extra evidence kept} & \multicolumn{2}{c}{predicted gain} \\
\cmidrule(lr){2-3}\cmidrule(lr){4-5}
corpus, reader & overlapping & disjoint & overlapping & disjoint \\
\midrule
MultiRC, \texttt{open-decider} & \kqLineMrcDecOverlapDKept & \kqLineMrcDecDisjointDKept & \kqLineMrcDecOverlapPred & \kqLineMrcDecDisjointPred \\
MultiRC, \texttt{gpt-6-luna} & \kqLineMrcOaiOverlapDKept & \kqLineMrcOaiDisjointDKept & \kqLineMrcOaiOverlapPred & \kqLineMrcOaiDisjointPred \\
FEVER, \texttt{gpt-6-luna} & \kqLineFeverOverlapDKept & \kqLineFeverDisjointDKept & \kqLineFeverOverlapPred & \kqLineFeverDisjointPred \\
Ev.\ Inference, \texttt{gpt-6-luna} & \kqLineEiOverlapDKept & \kqLineEiDisjointDKept & \kqLineEiOverlapPred & \kqLineEiDisjointPred \\
\bottomrule
\end{tabular}
}
\end{table}

\section{\LLMLtwo{} and other compressors at sentence and token budgets}
\label{app:lingua}

\paragraph{Setup.} We use \texttt{microsoft/\allowbreak llmlingua-2-\allowbreak xlm-roberta-\allowbreak large-\allowbreak meetingbank} with \texttt{llmlingua}
0.2.2, calling \texttt{compress\_prompt} on each document with no question, at rate $\mathrm{round}(bn)/n$ for a
document of $n$ sentences, the sentence share of a random $b$-budget subset, as a nominal target. Realized token shares are below. The compressed text is read with the same prompts, readers and documents as the
sentence policies. \LLMLtwo{} $-$ random contrasts
pair it with random deletion averaged over the seeded masks of Appendix~\ref{app:masks}; typed $-$
\LLMLtwo{} contrasts pair it with the main-text typed-selection read. At 25\% it keeps \llMrcTFLinguaKeptTokens\%,
\llFeverTFLinguaKeptTokens\% and \llEiTFLinguaKeptTokens\% of the tokens on MultiRC, FEVER and Evidence
Inference by its own tokenizer (\csMrcLinguaTFTokShare, \csFeverLinguaTFTokShare{} and \csEiLinguaTFTokShare\% in o200k tokens), and \llMrcTFLinguaKeptChars\%, \llFeverTFLinguaKeptChars\% and \llEiTFLinguaKeptChars\% of the
characters (random deletion \llMrcTFRandomKeptChars\%, \llFeverTFRandomKeptChars\% and
\llEiTFRandomKeptChars\%). The two are
matched in nominal retained share, not in tokens; the follow-up below repeats the comparison at exactly
matched token counts. It keeps \llMrcTFLinguaKeptWords\%,
\llFeverTFLinguaKeptWords\% and \llEiTFLinguaKeptWords\% of the words. Word shares count \LLMLtwo{}'s own word
units, and whitespace-separated words for the sentence policies, so comparisons between the two are
approximate.

\paragraph{Evidence kept.} At 25\% \LLMLtwo{} keeps \llMrcTFLinguaHiddenEv\%, \llFeverTFLinguaHiddenEv\% and
\llEiTFLinguaHiddenEv\% of the words of hidden questions' evidence sentences (MultiRC, FEVER, Evidence
Inference), against \llMrcTFRandomHiddenEv\%, \llFeverTFRandomHiddenEv\% and \llEiTFRandomHiddenEv\% for random
deletion and \llMrcTFKnownHiddenEv\%, \llFeverTFKnownHiddenEv\% and \llEiTFKnownHiddenEv\% for typed
selection. For the known question it keeps \llMrcTFLinguaKnownEv\%, \llFeverTFLinguaKnownEv\% and
\llEiTFLinguaKnownEv\%, where typed selection keeps \llMrcTFKnownKnownEv\%, \llFeverTFKnownKnownEv\% and
\llEiTFKnownKnownEv\%. It thus keeps evidence words at roughly the random rate, but spread across most sentences:
\llMrcTFLinguaSentCov\%, \llFeverTFLinguaSentCov\% and \llEiTFLinguaSentCov\% of sentences keep at least one word,
with each kept word placed in its sentence. Appendix~\ref{app:granularity} separates this spread from the scores.

\begin{table}[t]
\centering\footnotesize
\setlength{\tabcolsep}{3pt}
\caption{\LLMLtwo{} (L2), at a rate targeting random deletion's share, compared with random deletion
averaged over the seeded masks and with typed selection, on hidden and known questions (points,
paragraph or document bootstrap 95\% interval) at budget $b$ (\%; exploratory).}
\label{tab:lingua}
\begin{tabular}{lrllll}
\toprule
 & & \multicolumn{2}{c}{hidden questions} & \multicolumn{2}{c}{known question} \\
\cmidrule(lr){3-4}\cmidrule(lr){5-6}
corpus, reader & $b$ & L2 $-$ random & typed $-$ L2 & L2 $-$ random & typed $-$ L2 \\
\midrule
MultiRC, \texttt{open-decider} & 25 & \ci{\llMrcDecLinguaRandomTFHiddenDelta}{\llMrcDecLinguaRandomTFHiddenLo}{\llMrcDecLinguaRandomTFHiddenHi} & \ci{\llMrcDecKnownLinguaTFHiddenDelta}{\llMrcDecKnownLinguaTFHiddenLo}{\llMrcDecKnownLinguaTFHiddenHi} & \ci{\llMrcDecLinguaRandomTFKnownDelta}{\llMrcDecLinguaRandomTFKnownLo}{\llMrcDecLinguaRandomTFKnownHi} & \ci{\llMrcDecKnownLinguaTFKnownDelta}{\llMrcDecKnownLinguaTFKnownLo}{\llMrcDecKnownLinguaTFKnownHi} \\
 & 50 & \ci{\llMrcDecLinguaRandomFHiddenDelta}{\llMrcDecLinguaRandomFHiddenLo}{\llMrcDecLinguaRandomFHiddenHi} & \ci{\llMrcDecKnownLinguaFHiddenDelta}{\llMrcDecKnownLinguaFHiddenLo}{\llMrcDecKnownLinguaFHiddenHi} & \ci{\llMrcDecLinguaRandomFKnownDelta}{\llMrcDecLinguaRandomFKnownLo}{\llMrcDecLinguaRandomFKnownHi} & \ci{\llMrcDecKnownLinguaFKnownDelta}{\llMrcDecKnownLinguaFKnownLo}{\llMrcDecKnownLinguaFKnownHi} \\
MultiRC, \texttt{gpt-6-luna} & 25 & \ci{\llMrcOaiLinguaRandomTFHiddenDelta}{\llMrcOaiLinguaRandomTFHiddenLo}{\llMrcOaiLinguaRandomTFHiddenHi} & \ci{\llMrcOaiKnownLinguaTFHiddenDelta}{\llMrcOaiKnownLinguaTFHiddenLo}{\llMrcOaiKnownLinguaTFHiddenHi} & \ci{\llMrcOaiLinguaRandomTFKnownDelta}{\llMrcOaiLinguaRandomTFKnownLo}{\llMrcOaiLinguaRandomTFKnownHi} & \ci{\llMrcOaiKnownLinguaTFKnownDelta}{\llMrcOaiKnownLinguaTFKnownLo}{\llMrcOaiKnownLinguaTFKnownHi} \\
 & 50 & \ci{\llMrcOaiLinguaRandomFHiddenDelta}{\llMrcOaiLinguaRandomFHiddenLo}{\llMrcOaiLinguaRandomFHiddenHi} & \ci{\llMrcOaiKnownLinguaFHiddenDelta}{\llMrcOaiKnownLinguaFHiddenLo}{\llMrcOaiKnownLinguaFHiddenHi} & \ci{\llMrcOaiLinguaRandomFKnownDelta}{\llMrcOaiLinguaRandomFKnownLo}{\llMrcOaiLinguaRandomFKnownHi} & \ci{\llMrcOaiKnownLinguaFKnownDelta}{\llMrcOaiKnownLinguaFKnownLo}{\llMrcOaiKnownLinguaFKnownHi} \\
FEVER, \texttt{gpt-6-luna} & 25 & \ci{\llFeverLinguaRandomTFHiddenDelta}{\llFeverLinguaRandomTFHiddenLo}{\llFeverLinguaRandomTFHiddenHi} & \ci{\llFeverKnownLinguaTFHiddenDelta}{\llFeverKnownLinguaTFHiddenLo}{\llFeverKnownLinguaTFHiddenHi} & \ci{\llFeverLinguaRandomTFKnownDelta}{\llFeverLinguaRandomTFKnownLo}{\llFeverLinguaRandomTFKnownHi} & \ci{\llFeverKnownLinguaTFKnownDelta}{\llFeverKnownLinguaTFKnownLo}{\llFeverKnownLinguaTFKnownHi} \\
 & 50 & \ci{\llFeverLinguaRandomFHiddenDelta}{\llFeverLinguaRandomFHiddenLo}{\llFeverLinguaRandomFHiddenHi} & \ci{\llFeverKnownLinguaFHiddenDelta}{\llFeverKnownLinguaFHiddenLo}{\llFeverKnownLinguaFHiddenHi} & \ci{\llFeverLinguaRandomFKnownDelta}{\llFeverLinguaRandomFKnownLo}{\llFeverLinguaRandomFKnownHi} & \ci{\llFeverKnownLinguaFKnownDelta}{\llFeverKnownLinguaFKnownLo}{\llFeverKnownLinguaFKnownHi} \\
Ev.\ Inference, \texttt{gpt-6-luna} & 25 & \ci{\llEiLinguaRandomTFHiddenDelta}{\llEiLinguaRandomTFHiddenLo}{\llEiLinguaRandomTFHiddenHi} & \ci{\llEiKnownLinguaTFHiddenDelta}{\llEiKnownLinguaTFHiddenLo}{\llEiKnownLinguaTFHiddenHi} & \ci{\llEiLinguaRandomTFKnownDelta}{\llEiLinguaRandomTFKnownLo}{\llEiLinguaRandomTFKnownHi} & \ci{\llEiKnownLinguaTFKnownDelta}{\llEiKnownLinguaTFKnownLo}{\llEiKnownLinguaTFKnownHi} \\
 & 50 & \ci{\llEiLinguaRandomFHiddenDelta}{\llEiLinguaRandomFHiddenLo}{\llEiLinguaRandomFHiddenHi} & \ci{\llEiKnownLinguaFHiddenDelta}{\llEiKnownLinguaFHiddenLo}{\llEiKnownLinguaFHiddenHi} & \ci{\llEiLinguaRandomFKnownDelta}{\llEiLinguaRandomFKnownLo}{\llEiLinguaRandomFKnownHi} & \ci{\llEiKnownLinguaFKnownDelta}{\llEiKnownLinguaFKnownLo}{\llEiKnownLinguaFKnownHi} \\
\bottomrule
\end{tabular}
\end{table}

On the known claim at 25\% typed selection beats \LLMLtwo{} on FEVER and Evidence Inference (Table~\ref{tab:lingua}).
On MultiRC the difference is not detected with \texttt{open-decider}
(\llMrcDecKnownLinguaTFKnownDelta{} [\llMrcDecKnownLinguaTFKnownLo, \llMrcDecKnownLinguaTFKnownHi]), and with
\texttt{gpt-6-luna} \LLMLtwo{} is better
(\llMrcOaiKnownLinguaTFKnownDelta{} [\llMrcOaiKnownLinguaTFKnownLo, \llMrcOaiKnownLinguaTFKnownHi]). At 25\% on the hidden questions \LLMLtwo{} answers \llMrcOaiAccTFHidden\% (MultiRC, \texttt{gpt-6-luna}),
\llFeverAccTFHidden\% (FEVER) and \llEiAccTFHidden\% (Evidence Inference).

\label{app:followup}\paragraph{Follow-up plan.} One plan fixed four follow-ups with their comparisons, decision rules and budgets (A1--D2b). A: typed beats random deletion on the known question in every corpus, reader and budget (A1); on hidden
MultiRC questions it is within $\pm$\pqMargin{} points of random by the 90\% interval, for each reader and budget (A2a);
on hidden FEVER claims it beats random (A2b). B: if LongLLMLingua and MMR beat random on the known question and stay
within the margin or below it on hidden questions, the MultiRC hidden result is attributed to conditioning on the
known question. C1 and D0--D2b are stated with their results in Appendices~\ref{app:support} and~\ref{app:locomo}. Estimators and bootstrap units are the paper's: document-macro accuracy,
paired contrasts that drop an answer refused in either arm (on LoCoMo a refusal counts as wrong, as in
Appendix~\ref{app:locomo}), 2000 document resamples (conversations on LoCoMo). Random deletion at token budgets
averages three seeded masks per document, computed as in Appendix~\ref{app:masks}. Follow-ups A and B are reported here, C in Appendix~\ref{app:support} and D in Appendix~\ref{app:locomo}.

\paragraph{Token-matched budgets.} A document's budget is $T = \mathrm{round}(b\,n)$ tokens, where $n$ is the
length of its full text under the tiktoken \texttt{o200k\_base} encoding and $b \in \{25, 50\}\%$. A sentence
policy adds whole sentences in its rank order while the count stays at most $T$. It then cuts the first sentence that does
not fit to the remaining tokens and stops; the pieces stay in document order. Typed and question-agnostic selection
rank by the paper's scores (on MultiRC, each reader by its own), TF-IDF selection by similarity to the known
question, random deletion by a seeded permutation. For \LLMLtwo{} and LongLLMLingua, the compression rate (or
target length) is bisected to the smallest setting whose output reaches $T$ tokens, and the output is cut to $T$.
Every text is within \fuTokMaxDev{} token of its target. LongLLMLingua \citep{jiang2024longllmlingua} runs
llmlingua 0.2.2 with Llama-2-7B as its scoring model, the sentences as contexts and the known question (or claim)
as the query, with document order kept and the library's settings otherwise. MMR
\citep{carbonell1998mmr} ranks sentences greedily by $0.5$ times the cosine to the query minus $0.5$ times the
largest cosine to a sentence already chosen, with bge-large-en-v1.5 embeddings \citep{xiao2024cpack}. Reads use
the paper's prompts: \texttt{gpt-6-luna} on MultiRC and FEVER and \texttt{open-decider} on MultiRC.

\paragraph{Outcomes A and B.} On the known question typed $-$ random is
\fuMrcOaiTypedVsRandomTFKnownDelta{} and \fuMrcOaiTypedVsRandomFKnownDelta{} (MultiRC, \texttt{gpt-6-luna}, 25\% and
50\%), \fuMrcDecTypedVsRandomTFKnownDelta{} and \fuMrcDecTypedVsRandomFKnownDelta{} (\texttt{open-decider}) and
\fuFeverTypedVsRandomTFKnownDelta{} and \fuFeverTypedVsRandomFKnownDelta{} (FEVER), every interval above zero (A1). On
hidden FEVER claims it is \ci{\fuFeverTypedVsRandomTFHiddenDelta}{\fuFeverTypedVsRandomTFHiddenLo}{\fuFeverTypedVsRandomTFHiddenHi}
and \ci{\fuFeverTypedVsRandomFHiddenDelta}{\fuFeverTypedVsRandomFHiddenLo}{\fuFeverTypedVsRandomFHiddenHi} (A2b).
Table~\ref{tab:fu-ab} gives the accuracies on both question types and Table~\ref{tab:fu-abc} the contrasts with
random deletion on hidden questions. As predicted, typed selection
beats random deletion on the known question in every corpus, reader and budget. On hidden questions it is
within $\pm$\pqMargin{} points of random on MultiRC (90\% interval within that margin in \fuMrcTypedNEquiv{} of \fuMrcTypedNTests{} contrasts; at 25\%,
[\fuMrcOaiTypedVsRandomTFHiddenLoNinety, \fuMrcOaiTypedVsRandomTFHiddenHiNinety] with \texttt{gpt-6-luna} and [\fuMrcDecTypedVsRandomTFHiddenLoNinety, \fuMrcDecTypedVsRandomTFHiddenHiNinety] with \texttt{open-decider}) and above it on FEVER. With \texttt{open-decider} at 25\% it is also
detectably below random. MMR is within the margin in \fuMrcMmrNEquiv{} of \fuMrcMmrNTests{} contrasts and
LongLLMLingua in \fuMrcLongNEquiv{}, with the other cells below random. Both beat random deletion on the known
question except LongLLMLingua with \texttt{open-decider} at 50\% (\fuMrcDecLongVsRandomFKnownDelta{}
[\fuMrcDecLongVsRandomFKnownLo, \fuMrcDecLongVsRandomFKnownHi]).
On FEVER, all query-aware selectors gain on hidden
claims, as typed selection does at sentence budgets (Section~\ref{sec:fever}).

\begin{table}[t]
\centering\footnotesize
\setlength{\tabcolsep}{2.6pt}
\caption{At token-matched budgets, typed selection beats random deletion on the known question, and \LLMLtwo{} is
highest on hidden questions. Accuracy (\%) at 25\% and 50\% of each
document's \texttt{o200k\_base} tokens, on the known and hidden questions. MultiRC: candidate-level, paragraph-macro; FEVER: claim accuracy,
document-macro. Random deletion averages three masks. Each level uses every answer its policy gave; differences of
these levels are within 1 point of the paired contrasts of Table~\ref{tab:fu-abc}.}
\label{tab:fu-ab}
\input{generated/p4fu_ab_levels.tex}
\end{table}

\begin{table}[t]
\centering\scriptsize
\setlength{\tabcolsep}{1.8pt}
\caption{On hidden MultiRC questions, every policy except \LLMLtwo{} is near or below random deletion; on
FEVER, every policy except question-agnostic selection is above it. Hidden questions at token-matched budgets,
paired: an answer refused in either read is dropped from both. Policy/random: the two accuracies (\%) on the
remaining answers, document-macro, with random deletion averaged over its three masks within each document.
Difference: policy $-$ random, which equals the difference of the two levels up to rounding (points,
document-bootstrap 95\% interval).}
\label{tab:fu-abc}
\input{generated/p4fu_ab_contrasts.tex}
\end{table}

\section{Word-level compression: the granularity test and the hybrid}
\label{app:granularity}

\LLMLtwo{} differs from random sentence deletion in two ways: it deletes words, so most sentences keep some words,
and it chooses them by its scores. This experiment changes one at a time.

\paragraph{The known question.} On the known MultiRC question with \texttt{gpt-6-luna}, typed $-$ \LLMLtwo{} is
\fuMrcOaiTypedVsLinguaTFKnownDelta{} [\fuMrcOaiTypedVsLinguaTFKnownLo, \fuMrcOaiTypedVsLinguaTFKnownHi] and
\fuMrcOaiTypedVsLinguaFKnownDelta{} [\fuMrcOaiTypedVsLinguaFKnownLo, \ensuremath{-}0.02] points at matched tokens
(25\% and 50\%), and \llMrcOaiKnownLinguaTFKnownDelta{} [\llMrcOaiKnownLinguaTFKnownLo,
\llMrcOaiKnownLinguaTFKnownHi] and \llMrcOaiKnownLinguaFKnownDelta{} [\llMrcOaiKnownLinguaFKnownLo,
\llMrcOaiKnownLinguaFKnownHi] at the sentence share (Table~\ref{tab:lingua}): \LLMLtwo{} is ahead in all four
comparisons; the two intervals at the sentence share exclude zero, and the one at 50\% of the tokens just excludes it (upper end $-0.02$ points). With \texttt{open-decider}, typed selection is ahead at 50\% (\fuMrcDecTypedVsLinguaFKnownDelta{}
[\fuMrcDecTypedVsLinguaFKnownLo, \fuMrcDecTypedVsLinguaFKnownHi] at matched tokens) and the 25\% intervals span
zero. On the known FEVER claim typed selection is ahead (\fuFeverTypedVsLinguaTFKnownDelta{}
[\fuFeverTypedVsLinguaTFKnownLo, \fuFeverTypedVsLinguaTFKnownHi]).

\paragraph{Plan.} The plan fixed the policies, contrasts, readers and budgets below and predicted at 25\% that contrasts 1 (G1) and 2 (G2) are positive in each cell, with the interval above 0.

\paragraph{Policies.} Every text has the token budget $T$ of Appendix~\ref{app:followup}, to within
\grTokMaxDev{} token. Four policies fill it:
\begin{itemize}\itemsep0pt
\item \emph{Word scores}: the \LLMLtwo{} text of Appendix~\ref{app:followup}.
\item \emph{Sentence scores}: the same model, with no question. A word's score is its keep probability in
\texttt{compress\_prompt} (the ``preserve'' probability averaged over its tokens), and a sentence's score is the
mean over its words. Sentences are added in descending score, ties in document order, as for the sentence policies
of Appendix~\ref{app:followup}.
\item \emph{Random words}: whitespace-separated words are added in a seeded random order until the budget is
reached, then kept in document order. Each document has three masks.
\item \emph{Random sentences}: the random deletion of Appendix~\ref{app:followup}, with its three masks.
\end{itemize}
The four contrasts are (1) word scores $-$ sentence scores, (2) random words $-$ random sentences, (3) sentence
scores $-$ random sentences and (4) word scores $-$ random words. Contrasts 1 and 2 change only the unit of
deletion, contrasts 3 and 4 only whether the scores choose. Readers, prompts and estimators are those of
Appendix~\ref{app:followup}: paired, document-macro accuracy on hidden questions, with 2000 document resamples. A
random policy's value for a document is the mean over its three masks.

\paragraph{Results.} Prediction 1 is met in all three cells: at 25\%, word scores $-$ sentence
scores is \grMrcDecGranScorerTFDelta{} points (MultiRC, \texttt{open-decider}), \grMrcOaiGranScorerTFDelta{}
(MultiRC, \texttt{gpt-6-luna}) and \grFeverGranScorerTFDelta{} (FEVER; Table~\ref{tab:granularity}).
Prediction 2 is met with \texttt{gpt-6-luna}, with random words $-$ random sentences at
\grMrcOaiGranRandomTFDelta{} points on MultiRC and \grFeverGranRandomTFDelta{} on FEVER, but not with
\texttt{open-decider}
(\grMrcDecGranRandomTFDelta{} [\grMrcDecGranRandomTFLo, \grMrcDecGranRandomTFHi]). Applied to whole sentences,
the scores are not detectably better than random sentences in five of the six cells; the exception is
\texttt{open-decider} at 25\% (\grMrcDecScorerSentTFDelta{} [\grMrcDecScorerSentTFLo, \grMrcDecScorerSentTFHi]).
Applied to words, they beat random words in every cell. The \LLMLtwo{} advantage thus comes mainly from its scores
choosing words. Keeping words spread over nearly every sentence without the scores adds less (contrast 2 against contrast 4), and
with \texttt{open-decider} no gain is detected.

\begin{table}[t]
\centering\footnotesize
\setlength{\tabcolsep}{3pt}
\caption{The \LLMLtwo{} scores help when they choose words, and in five of six cells not when they choose
sentences. Top: contrasts on hidden questions at budget $b$ (\% of the tokens; points, document-bootstrap 95\%
interval), paired on the answers both policies gave. Bottom: hidden-question accuracy (\%) of sentence scores and random words; Table~\ref{tab:fu-ab} gives word scores
(\LLMLtwo{}) and random sentences. Random policies average three masks. luna: \texttt{gpt-6-luna}; dec: \texttt{open-decider}.}
\label{tab:granularity}
\begin{tabular}{lrllll}
\toprule
& & \multicolumn{2}{c}{words $-$ sentences} & \multicolumn{2}{c}{scores $-$ random} \\
\cmidrule(lr){3-4}\cmidrule(lr){5-6}
corpus, reader & $b$ & by scores (1) & at random (2) & sentences (3) & words (4) \\
\midrule
MultiRC, dec & 25 & \ci{\grMrcDecGranScorerTFDelta}{\grMrcDecGranScorerTFLo}{\grMrcDecGranScorerTFHi} & \ci{\grMrcDecGranRandomTFDelta}{\grMrcDecGranRandomTFLo}{\grMrcDecGranRandomTFHi} & \ci{\grMrcDecScorerSentTFDelta}{\grMrcDecScorerSentTFLo}{\grMrcDecScorerSentTFHi} & \ci{\grMrcDecScorerWordTFDelta}{\grMrcDecScorerWordTFLo}{\grMrcDecScorerWordTFHi} \\
 & 50 & \ci{\grMrcDecGranScorerFDelta}{\grMrcDecGranScorerFLo}{\grMrcDecGranScorerFHi} & \ci{\grMrcDecGranRandomFDelta}{\grMrcDecGranRandomFLo}{\grMrcDecGranRandomFHi} & \ci{\grMrcDecScorerSentFDelta}{\grMrcDecScorerSentFLo}{\grMrcDecScorerSentFHi} & \ci{\grMrcDecScorerWordFDelta}{\grMrcDecScorerWordFLo}{\grMrcDecScorerWordFHi} \\
MultiRC, luna & 25 & \ci{\grMrcOaiGranScorerTFDelta}{\grMrcOaiGranScorerTFLo}{\grMrcOaiGranScorerTFHi} & \ci{\grMrcOaiGranRandomTFDelta}{\grMrcOaiGranRandomTFLo}{\grMrcOaiGranRandomTFHi} & \ci{\grMrcOaiScorerSentTFDelta}{\grMrcOaiScorerSentTFLo}{\grMrcOaiScorerSentTFHi} & \ci{\grMrcOaiScorerWordTFDelta}{\grMrcOaiScorerWordTFLo}{\grMrcOaiScorerWordTFHi} \\
 & 50 & \ci{\grMrcOaiGranScorerFDelta}{\grMrcOaiGranScorerFLo}{\grMrcOaiGranScorerFHi} & \ci{\grMrcOaiGranRandomFDelta}{\grMrcOaiGranRandomFLo}{\grMrcOaiGranRandomFHi} & \ci{\grMrcOaiScorerSentFDelta}{\grMrcOaiScorerSentFLo}{\grMrcOaiScorerSentFHi} & \ci{\grMrcOaiScorerWordFDelta}{\grMrcOaiScorerWordFLo}{\grMrcOaiScorerWordFHi} \\
FEVER, luna & 25 & \ci{\grFeverGranScorerTFDelta}{\grFeverGranScorerTFLo}{\grFeverGranScorerTFHi} & \ci{\grFeverGranRandomTFDelta}{\grFeverGranRandomTFLo}{\grFeverGranRandomTFHi} & \ci{\grFeverScorerSentTFDelta}{\grFeverScorerSentTFLo}{\grFeverScorerSentTFHi} & \ci{\grFeverScorerWordTFDelta}{\grFeverScorerWordTFLo}{\grFeverScorerWordTFHi} \\
 & 50 & \ci{\grFeverGranScorerFDelta}{\grFeverGranScorerFLo}{\grFeverGranScorerFHi} & \ci{\grFeverGranRandomFDelta}{\grFeverGranRandomFLo}{\grFeverGranRandomFHi} & \ci{\grFeverScorerSentFDelta}{\grFeverScorerSentFLo}{\grFeverScorerSentFHi} & \ci{\grFeverScorerWordFDelta}{\grFeverScorerWordFLo}{\grFeverScorerWordFHi} \\
\bottomrule
\end{tabular}
\medskip

\begin{tabular}{lrrr}
\toprule
corpus, reader & $b$ & sentence scores & random words \\
\midrule
MultiRC, dec & 25 & \grMrcDecLinguaSentTFHidden & \grMrcDecRandWordTFHidden \\
 & 50 & \grMrcDecLinguaSentFHidden & \grMrcDecRandWordFHidden \\
MultiRC, luna & 25 & \grMrcOaiLinguaSentTFHidden & \grMrcOaiRandWordTFHidden \\
 & 50 & \grMrcOaiLinguaSentFHidden & \grMrcOaiRandWordFHidden \\
FEVER, luna & 25 & \grFeverLinguaSentTFHidden & \grFeverRandWordTFHidden \\
 & 50 & \grFeverLinguaSentFHidden & \grFeverRandWordFHidden \\
\bottomrule
\end{tabular}
\end{table}

\paragraph{What the texts keep.} At 25\%, the two word policies leave at least one word in
\grMrcLinguaTFSentCov\% and \grMrcRandWordTFSentCov\% of MultiRC sentences (FEVER: \grFeverLinguaTFSentCov\% and
\grFeverRandWordTFSentCov\%). The two sentence policies keep \grMrcLinguaSentTFSentCov\% and
\grMrcRandSentTFSentCov\% of sentences (FEVER: \grFeverLinguaSentTFSentCov\% and \grFeverRandSentTFSentCov\%). The four keep
similar shares of the words of hidden questions' evidence sentences on MultiRC (word scores \grMrcLinguaTFEvWords\%,
sentence scores \grMrcLinguaSentTFEvWords\%, random words \grMrcRandWordTFEvWords\%, random sentences
\grMrcRandSentTFEvWords\%). On FEVER sentence scores keep the most (\grFeverLinguaSentTFEvWords\%, against
\grFeverLinguaTFEvWords\%, \grFeverRandWordTFEvWords\% and \grFeverRandSentTFEvWords\%) and are still less accurate than both word policies on hidden claims (\grFeverLinguaSentTFHidden\%, against
\grFeverLinguaTFHidden\% for word scores and \grFeverRandWordTFHidden\% for random words). Word scores and random words keep about the same share of evidence words, so what the
scores add is the choice of which words to keep.

\paragraph{The hybrid compactor.} \label{app:hybrid}

The plan, written before any request of this experiment, asked whether a compaction that keeps the known question's
evidence as whole sentences and a few scored words of every other sentence beats typed selection on hidden questions
while staying ahead of \LLMLtwo{} on the known one. At the token budget $T$ of the follow-ups
(Appendix~\ref{app:followup}), the hybrid first adds whole sentences in typed-selection order (each reader's own typed
scores) while they stay within $aT$, with $a = 0.5$ (primary) or $0.25$ (secondary); it then fills the remaining budget
with the other sentences' words in order of \LLMLtwo{} keep probability (the scores of the granularity test,
no question), keeping all pieces in document order. Typed selection, \LLMLtwo{}, random sentences and random words
(three masks each) are re-read in the same run, so no contrast mixes runs. Estimators are the follow-ups'.
Table~\ref{tab:hybrid} gives the 25\% accuracies and evidence kept. The plan predicted, in each of the three cells,
that the hybrid beats typed selection on hidden questions (H1), beats \LLMLtwo{} on the known question (H2) and is at
most 2 points below \LLMLtwo{} on hidden questions by the 90\% lower end (H3). Hybrid $-$ typed, hidden, is
\ci{\hyMrcOaiHybHalfTypedTFHiddenDelta}{\hyMrcOaiHybHalfTypedTFHiddenLo}{\hyMrcOaiHybHalfTypedTFHiddenHi} (MultiRC,
luna), \ci{\hyMrcDecHybHalfTypedTFHiddenDelta}{\hyMrcDecHybHalfTypedTFHiddenLo}{\hyMrcDecHybHalfTypedTFHiddenHi}
(dec) and \ci{\hyFeverHybHalfTypedTFHiddenDelta}{\hyFeverHybHalfTypedTFHiddenLo}{\hyFeverHybHalfTypedTFHiddenHi}
(FEVER): H1 is met in all three. Hybrid $-$ \LLMLtwo{}, known, is
\ci{\hyMrcOaiHybHalfLinguaTFKnownDelta}{\hyMrcOaiHybHalfLinguaTFKnownLo}{\hyMrcOaiHybHalfLinguaTFKnownHi},
\ci{\hyMrcDecHybHalfLinguaTFKnownDelta}{\hyMrcDecHybHalfLinguaTFKnownLo}{\hyMrcDecHybHalfLinguaTFKnownHi} and
\ci{\hyFeverHybHalfLinguaTFKnownDelta}{\hyFeverHybHalfLinguaTFKnownLo}{\hyFeverHybHalfLinguaTFKnownHi}: H2 is
\hyMrcOaiHTwoMet, \hyMrcDecHTwoMet{} and \hyFeverHTwoMet. Hybrid $-$ \LLMLtwo{}, hidden, is
\hyMrcOaiHybHalfLinguaTFHiddenDelta{} [\hyMrcOaiHybHalfLinguaTFHiddenLoNinety, \hyMrcOaiHybHalfLinguaTFHiddenHiNinety],
\hyMrcDecHybHalfLinguaTFHiddenDelta{} [\hyMrcDecHybHalfLinguaTFHiddenLoNinety, \hyMrcDecHybHalfLinguaTFHiddenHiNinety]
and \hyFeverHybHalfLinguaTFHiddenDelta{} [\hyFeverHybHalfLinguaTFHiddenLoNinety, \hyFeverHybHalfLinguaTFHiddenHiNinety]
(90\% intervals): H3 is \hyMrcOaiHThreeMet, \hyMrcDecHThreeMet{} and \hyFeverHThreeMet. The quarter-budget hybrid was planned without a prediction.

\begin{table}[h]
\centering\small
\setlength{\tabcolsep}{3.5pt}
\caption{Hybrid compaction at 25\% of the tokens. Top: known and hidden accuracy (\%); bottom: share of the known and
hidden questions' evidence words kept (\%). The hybrid keeps typed whole sentences up to half (or a quarter) of the
budget and fills the rest with \LLMLtwo{} words. Hidden accuracy here covers all hidden questions; Table~\ref{tab:words}
covers those with annotated evidence, so the two differ by up to 0.1 points.}
\label{tab:hybrid}
\begin{tabular}{lrrrrrr}
\toprule
& \multicolumn{2}{c}{MultiRC, \texttt{gpt-6-luna}} & \multicolumn{2}{c}{MultiRC, \texttt{open-decider}} & \multicolumn{2}{c}{FEVER, \texttt{gpt-6-luna}} \\
policy & known & hidden & known & hidden & known & hidden \\
\midrule
typed & \hyMrcOaiTypedTFKnown & \hyMrcOaiTypedTFHidden & \hyMrcDecTypedTFKnown & \hyMrcDecTypedTFHidden & \hyFeverTypedTFKnown & \hyFeverTypedTFHidden \\
\LLMLtwo{} & \hyMrcOaiLinguaTFKnown & \hyMrcOaiLinguaTFHidden & \hyMrcDecLinguaTFKnown & \hyMrcDecLinguaTFHidden & \hyFeverLinguaTFKnown & \hyFeverLinguaTFHidden \\
hybrid, half & \hyMrcOaiHybHalfTFKnown & \hyMrcOaiHybHalfTFHidden & \hyMrcDecHybHalfTFKnown & \hyMrcDecHybHalfTFHidden & \hyFeverHybHalfTFKnown & \hyFeverHybHalfTFHidden \\
hybrid, quarter & \hyMrcOaiHybQuarterTFKnown & \hyMrcOaiHybQuarterTFHidden & \hyMrcDecHybQuarterTFKnown & \hyMrcDecHybQuarterTFHidden & \hyFeverHybQuarterTFKnown & \hyFeverHybQuarterTFHidden \\
random sentences & \hyMrcOaiRandSentTFKnown & \hyMrcOaiRandSentTFHidden & \hyMrcDecRandSentTFKnown & \hyMrcDecRandSentTFHidden & \hyFeverRandSentTFKnown & \hyFeverRandSentTFHidden \\
random words & \hyMrcOaiRandWordTFKnown & \hyMrcOaiRandWordTFHidden & \hyMrcDecRandWordTFKnown & \hyMrcDecRandWordTFHidden & \hyFeverRandWordTFKnown & \hyFeverRandWordTFHidden \\
\midrule
\multicolumn{7}{l}{\textit{evidence words kept}} \\
typed & \hyMrcOaiTypedTFEvKnown & \hyMrcOaiTypedTFEvHidden & \hyMrcDecTypedTFEvKnown & \hyMrcDecTypedTFEvHidden & \hyFeverTypedTFEvKnown & \hyFeverTypedTFEvHidden \\
\LLMLtwo{} & \hyMrcOaiLinguaTFEvKnown & \hyMrcOaiLinguaTFEvHidden & \hyMrcDecLinguaTFEvKnown & \hyMrcDecLinguaTFEvHidden & \hyFeverLinguaTFEvKnown & \hyFeverLinguaTFEvHidden \\
hybrid, half & \hyMrcOaiHybHalfTFEvKnown & \hyMrcOaiHybHalfTFEvHidden & \hyMrcDecHybHalfTFEvKnown & \hyMrcDecHybHalfTFEvHidden & \hyFeverHybHalfTFEvKnown & \hyFeverHybHalfTFEvHidden \\
hybrid, quarter & \hyMrcOaiHybQuarterTFEvKnown & \hyMrcOaiHybQuarterTFEvHidden & \hyMrcDecHybQuarterTFEvKnown & \hyMrcDecHybQuarterTFEvHidden & \hyFeverHybQuarterTFEvKnown & \hyFeverHybQuarterTFEvHidden \\
\bottomrule
\end{tabular}
\end{table}

\section{Content-word evidence retention (exploratory)}
\label{app:whywords}

This analysis was not planned in advance and uses only the texts and reads of the hybrid experiment
(Appendix~\ref{app:hybrid}); it sends no new request. A word is a \emph{content word} if, lower-cased and stripped of
surrounding punctuation, it contains a letter or digit and is not in scikit-learn's English stop list. For each hidden
question and compacted text we measure four shares. \emph{All words} and \emph{content} are the shares of its annotated
evidence's words and content words kept. \emph{Breadth} is the share of its evidence sentences with at least one
content word kept, and, on MultiRC, \emph{answer} is the share of each candidate answer's content words that occur in
the kept text. The unit is one candidate answer (MultiRC) or claim (FEVER) in one compacted text, scored
right or wrong, with refusals dropped and weights that make every document and policy count once. A linear probability
model with document fixed effects and policy indicators pools the six policies of a cell; a feature enters as
indicators of gold label by feature bin (0 and 1 are their own bins, interior values are cut at their pooled deciles)
or, in the linear form, as one slope per gold label. The policy gap left after conditioning is the difference of policy
coefficients, with 2000 document-bootstrap resamples.

The FEVER evidence-support check of Section~\ref{sec:compressors} uses the binary reads. A claim's target is
``supported'' if it is true and its evidence is fully kept (all evidence sentences for sentence policies, all evidence
content words for word policies), ``not supported'' if its evidence is entirely deleted, and undefined otherwise;
accuracy weights the two targets equally. Unlike the three-way target of Appendix~\ref{app:support}, it leaves a false claim
with fully kept evidence undefined, so the FEVER typed $-$ random contrast covers fewer claims (\fsOrigStrictTypedRandSentN{}
against \fuCFeverTypedVsRandomTFSupportNClaims). The label split of Section~\ref{sec:compressors} covers
\fsHyLabTrueLinguaRandSentN{} true and \fsHyLabFalseLinguaRandSentN{} false hidden claims; the bin with at most a fifth of the
evidence content words kept holds \fsHyYesRLeRelaxLinguaN{} true claims after \LLMLtwo{} and \fsHyYesRLeRelaxRandSentN{} after
random sentence deletion; the target is defined for \fsHyStrictRefLinguaN{} of the \fsHyLabAllLinguaN{} hidden claims answered under
\LLMLtwo{} (one claim refused there is dropped), for \fsOrigStrictTypedRandSentN{} claims in the typed $-$ random contrast and for \fsHyStrictHybHalfTypedN{} in the
hybrid $-$ typed contrast. The relaxed variant counts evidence as kept above 80\% and as deleted below
20\% of its content words; under it \LLMLtwo{} $-$ random sentence deletion is
\ci{\fsHyRelaxedLinguaRandSentBalDelta}{\fsHyRelaxedLinguaRandSentBalLo}{\fsHyRelaxedLinguaRandSentBalHi} points on the
\fsHyRelaxedLinguaRandSentN{} claims where both texts define it.

Table~\ref{tab:whywords} gives the gaps at 25\%. Content-word retention lowers every gap and leaves no detected
\LLMLtwo{} $-$ random-sentence gap on FEVER. Breadth alone over-corrects: conditioning on it turns several gaps
negative, because random word deletion also keeps some word of most evidence sentences yet scores lower. On FEVER the
all-word measure, binned, also absorbs most of the \LLMLtwo{} gap, because the relation between FEVER accuracy and
retention is strongly non-linear; in the linear form content-word retention leaves \wwFeverLinguaRandSentTFHiddenEvContentLinGap{}
points there, so on FEVER the two measures are not separated.

\begin{table}[t]
\centering\scriptsize
\setlength{\tabcolsep}{2.6pt}
\caption{Hidden-question gaps (points) at 25\% before and after conditioning on a retention measure (exploratory;
document-bootstrap 95\% intervals). Columns after the raw gap: binned all-word (all), content-word, content-word plus answer
and breadth retention, and content-word retention in the linear form (linear). Raw: the model without retention terms, on hidden
questions with annotated evidence; the paired contrasts of Appendices~\ref{app:hybrid} and~\ref{app:followup} differ
slightly. luna: \texttt{gpt-6-luna}; dec:
\texttt{open-decider}.}
\label{tab:whywords}
\begin{tabular}{llllllll}
\toprule
cell & gap & raw & all & content & +answer & breadth & linear \\
\midrule
MultiRC, luna & \LLMLtwo{} $-$ random sent. & \ci{\wwMrcOaiLinguaRandSentTFHiddenRawGap}{\wwMrcOaiLinguaRandSentTFHiddenRawLo}{\wwMrcOaiLinguaRandSentTFHiddenRawHi} & \wwMrcOaiLinguaRandSentTFHiddenEvWordsGap & \ci{\wwMrcOaiLinguaRandSentTFHiddenEvContentGap}{\wwMrcOaiLinguaRandSentTFHiddenEvContentLo}{\wwMrcOaiLinguaRandSentTFHiddenEvContentHi} & \wwMrcOaiLinguaRandSentTFHiddenContentAnsGap & \wwMrcOaiLinguaRandSentTFHiddenEvBreadthGap & \wwMrcOaiLinguaRandSentTFHiddenEvContentLinGap \\
MultiRC, luna & hybrid $-$ typed & \ci{\wwMrcOaiHybHalfTypedTFHiddenRawGap}{\wwMrcOaiHybHalfTypedTFHiddenRawLo}{\wwMrcOaiHybHalfTypedTFHiddenRawHi} & \wwMrcOaiHybHalfTypedTFHiddenEvWordsGap & \ci{\wwMrcOaiHybHalfTypedTFHiddenEvContentGap}{\wwMrcOaiHybHalfTypedTFHiddenEvContentLo}{\wwMrcOaiHybHalfTypedTFHiddenEvContentHi} & \wwMrcOaiHybHalfTypedTFHiddenContentAnsGap & \wwMrcOaiHybHalfTypedTFHiddenEvBreadthGap & \wwMrcOaiHybHalfTypedTFHiddenEvContentLinGap \\
MultiRC, dec & \LLMLtwo{} $-$ random sent. & \ci{\wwMrcDecLinguaRandSentTFHiddenRawGap}{\wwMrcDecLinguaRandSentTFHiddenRawLo}{\wwMrcDecLinguaRandSentTFHiddenRawHi} & \wwMrcDecLinguaRandSentTFHiddenEvWordsGap & \ci{\wwMrcDecLinguaRandSentTFHiddenEvContentGap}{\wwMrcDecLinguaRandSentTFHiddenEvContentLo}{\wwMrcDecLinguaRandSentTFHiddenEvContentHi} & \wwMrcDecLinguaRandSentTFHiddenContentAnsGap & \wwMrcDecLinguaRandSentTFHiddenEvBreadthGap & \wwMrcDecLinguaRandSentTFHiddenEvContentLinGap \\
MultiRC, dec & hybrid $-$ typed & \ci{\wwMrcDecHybHalfTypedTFHiddenRawGap}{\wwMrcDecHybHalfTypedTFHiddenRawLo}{\wwMrcDecHybHalfTypedTFHiddenRawHi} & \wwMrcDecHybHalfTypedTFHiddenEvWordsGap & \ci{\wwMrcDecHybHalfTypedTFHiddenEvContentGap}{\wwMrcDecHybHalfTypedTFHiddenEvContentLo}{\wwMrcDecHybHalfTypedTFHiddenEvContentHi} & \wwMrcDecHybHalfTypedTFHiddenContentAnsGap & \wwMrcDecHybHalfTypedTFHiddenEvBreadthGap & \wwMrcDecHybHalfTypedTFHiddenEvContentLinGap \\
FEVER, luna & \LLMLtwo{} $-$ random sent. & \ci{\wwFeverLinguaRandSentTFHiddenRawGap}{\wwFeverLinguaRandSentTFHiddenRawLo}{\wwFeverLinguaRandSentTFHiddenRawHi} & \wwFeverLinguaRandSentTFHiddenEvWordsGap & \ci{\wwFeverLinguaRandSentTFHiddenEvContentGap}{\wwFeverLinguaRandSentTFHiddenEvContentLo}{\wwFeverLinguaRandSentTFHiddenEvContentHi} & --- & \wwFeverLinguaRandSentTFHiddenEvBreadthGap & \wwFeverLinguaRandSentTFHiddenEvContentLinGap \\
FEVER, luna & hybrid $-$ typed & \ci{\wwFeverHybHalfTypedTFHiddenRawGap}{\wwFeverHybHalfTypedTFHiddenRawLo}{\wwFeverHybHalfTypedTFHiddenRawHi} & \wwFeverHybHalfTypedTFHiddenEvWordsGap & \ci{\wwFeverHybHalfTypedTFHiddenEvContentGap}{\wwFeverHybHalfTypedTFHiddenEvContentLo}{\wwFeverHybHalfTypedTFHiddenEvContentHi} & --- & \wwFeverHybHalfTypedTFHiddenEvBreadthGap & \wwFeverHybHalfTypedTFHiddenEvContentLinGap \\
MultiRC, luna & \LLMLtwo{} $-$ random words & \ci{\wwMrcOaiLinguaRandWordTFHiddenRawGap}{\wwMrcOaiLinguaRandWordTFHiddenRawLo}{\wwMrcOaiLinguaRandWordTFHiddenRawHi} & \wwMrcOaiLinguaRandWordTFHiddenEvWordsGap & \ci{\wwMrcOaiLinguaRandWordTFHiddenEvContentGap}{\wwMrcOaiLinguaRandWordTFHiddenEvContentLo}{\wwMrcOaiLinguaRandWordTFHiddenEvContentHi} & \wwMrcOaiLinguaRandWordTFHiddenContentAnsGap & \wwMrcOaiLinguaRandWordTFHiddenEvBreadthGap & \wwMrcOaiLinguaRandWordTFHiddenEvContentLinGap \\
MultiRC, dec & \LLMLtwo{} $-$ random words & \ci{\wwMrcDecLinguaRandWordTFHiddenRawGap}{\wwMrcDecLinguaRandWordTFHiddenRawLo}{\wwMrcDecLinguaRandWordTFHiddenRawHi} & \wwMrcDecLinguaRandWordTFHiddenEvWordsGap & \ci{\wwMrcDecLinguaRandWordTFHiddenEvContentGap}{\wwMrcDecLinguaRandWordTFHiddenEvContentLo}{\wwMrcDecLinguaRandWordTFHiddenEvContentHi} & \wwMrcDecLinguaRandWordTFHiddenContentAnsGap & \wwMrcDecLinguaRandWordTFHiddenEvBreadthGap & \wwMrcDecLinguaRandWordTFHiddenEvContentLinGap \\
FEVER, luna & \LLMLtwo{} $-$ random words & \ci{\wwFeverLinguaRandWordTFHiddenRawGap}{\wwFeverLinguaRandWordTFHiddenRawLo}{\wwFeverLinguaRandWordTFHiddenRawHi} & \wwFeverLinguaRandWordTFHiddenEvWordsGap & \ci{\wwFeverLinguaRandWordTFHiddenEvContentGap}{\wwFeverLinguaRandWordTFHiddenEvContentLo}{\wwFeverLinguaRandWordTFHiddenEvContentHi} & --- & \wwFeverLinguaRandWordTFHiddenEvBreadthGap & \wwFeverLinguaRandWordTFHiddenEvContentLinGap \\
\bottomrule
\end{tabular}
\end{table}

\section{The three-way read and the support target}
\label{app:support}

This prespecified follow-up (C1) re-reads the hidden claims with three answers. C1 predicted that
typed selection beats random deletion in label recovery (labels weighted equally) on both corpora at 25\%; it does,
by \ci{\fuCFeverTypedVsRandomTFRecoveryDelta}{\fuCFeverTypedVsRandomTFRecoveryLo}{\fuCFeverTypedVsRandomTFRecoveryHi} on FEVER and
\ci{\fuCEiTypedVsRandomTFRecoveryDelta}{\fuCEiTypedVsRandomTFRecoveryLo}{\fuCEiTypedVsRandomTFRecoveryHi} on Evidence Inference. The texts are those of the paper's reads (full, typed selection and random
deletion at 25\% and 50\%, the paper's mask). One request per text asks, for each claim, a Choice among
``supported'', ``refuted'' and ``not enough info'' (``the passage does not contain enough information to decide
the claim''), with no other question. Gold labels: FEVER SUPPORTS and Evidence Inference significant differences
are ``supported'', the rest ``refuted''. Label recovery scores against the document's label (``not enough info''
counts wrong). The support target of a claim in a text is the document's label if all of the
claim's annotated evidence is kept, ``not enough info'' if none is, and undefined if part of it is kept
(\fuCFeverKnownTFHiddenNPartial{} of the hidden FEVER claims after typed selection at 25\%,
\fuCEiKnownTFHiddenNPartial{} on Evidence Inference). Table~\ref{tab:fu-c} gives each text's label-recovery levels over all claims its read
answered.

Typed selection and random deletion are compared on a common set (Table~\ref{tab:fu-c-paired}): for each corpus
and budget, the hidden claims that both reads answered and, for the support target, that have a target in both
texts. On that set, let $a_{t,\ell,d}$ be text $t$'s accuracy on the claims of document $d$ whose target in $t$ is
$\ell$, and $\bar a_{t,\ell}$ its mean over the documents that have such a claim. Text $t$ scores
$S_t = |L|^{-1} \sum_{\ell \in L} \bar a_{t,\ell}$, with $L$ the two labels for recovery and the three for support,
and the contrast is $S_{\text{typed}} - S_{\text{random}}$ with a 2000-resample document bootstrap. The table's two
levels thus subtract to the contrast up to rounding. A claim's support target can differ between the two texts, so the support
contrast compares the reader on the claim mix each text leaves. Typed selection recovers the label more often
than random deletion at both budgets on both corpora, as predicted at 25\%; the support target had no prediction
(Table~\ref{tab:fu-c-paired}). Comparing the
two texts only within documents where both have a claim with a given target (exploratory) gives
\fuCFeverTypedVsRandomTFSupportWithinDocumentsDelta{} [\fuCFeverTypedVsRandomTFSupportWithinDocumentsLo,
\fuCFeverTypedVsRandomTFSupportWithinDocumentsHi] and \fuCEiTypedVsRandomTFSupportWithinDocumentsDelta{}
[\fuCEiTypedVsRandomTFSupportWithinDocumentsLo, \fuCEiTypedVsRandomTFSupportWithinDocumentsHi] at 25\% (FEVER,
Evidence Inference): on Evidence Inference at 25\%, the small support gain depends on this choice. At 50\% on Evidence Inference, the support contrast exceeds the label contrast.

\begin{table}[t]
\centering\footnotesize
\setlength{\tabcolsep}{3pt}
\caption{Typed selection recovers the document's label more often than random deletion. Three-way read of the
hidden claims (\texttt{gpt-6-luna}; \%, document-macro, over all claims each read answered). Document label
recovered: per gold label and label-balanced. ``Not enough info'': share of claims whose annotated evidence was all
deleted that the reader answers so. Bottom: Label recovery and annotation-derived support accuracy on common claim sets.
Hidden claims of the three-way read (\texttt{gpt-6-luna}) on the common set of each row: claims answered in both
texts and, for support, with a target in both. Claims, documents: size of that set. Typed, random: label-balanced
document-macro accuracy $S_t$ (\%, defined in the text). Typed $-$ random: $S_{\text{typed}} -
S_{\text{random}}$, equal to the difference of the two levels up to rounding (points, document-bootstrap 95\%
interval).}
\label{tab:fu-c}
\input{generated/p4fu_c.tex}
\medskip
{\footnotesize
\setlength{\tabcolsep}{3pt}
\label{tab:fu-c-paired}
\input{generated/p4fu_c_paired.tex}
}
\end{table}

\section{The prespecified KEEP--DROP contrast}
\label{app:retention}

Before any request of this experiment, we specified the corpora, the two arms, the
reader and prompt, the size-based exclusion, the estimand, the bootstrap seeds, both predictions with their test and level,
and which outcomes are descriptive. Only item construction and token counting, which use sentence sets alone, came first.

\paragraph{Design.} The sample is every document already read under typed selection at 25\%: \prtFeverNDocs{}
FEVER documents and \prtEiNDocs{} Evidence Inference articles. Let $K$ be typed selection's kept set, $k=|K|$,
and $E$ the union of a hidden claim's annotated rationale groups. KEEP is $K$ with $E\setminus K$ added and as many sentences of
$K\setminus E$ removed, lowest typed-selection score first; DROP is $K$ with $K\cap E$ removed and as many sentences
drawn uniformly (seeded per claim) from outside $K\cup E$ added. Both have $k$ sentences in document order; KEEP
holds all of $E$ and DROP none of it, and when $K$ already holds all of $E$ (or none), KEEP (or DROP) is $K$
itself. Sentences swapped in or out may be evidence for other claims, so the two arms can differ in more than the
claim's evidence. Averaged over the analyzed hidden claims (a claim mean, unlike the retention shares of Table~\ref{tab:third}), $K$ already holds
\prtFeverEvInKnown\% of a hidden claim's evidence on FEVER and \prtEiEvInKnown\% on Evidence Inference.
\texttt{gpt-6-luna} answers one claim per request with the claim and sufficiency checks of the main reads,
worded as before; a claim counts as correct when its probability is on the side of its gold label. All
\prtNRequests{} requests were sent as one shuffled list, so arms and corpora are interleaved in time
(\prtInputTokens{} input tokens, \$\prtCostUSD).

\paragraph{Exclusions.} Hidden claims are those with a single gold label (Appendix~\ref{app:fever}): the
\prtLabFeverNConflicting{} FEVER claims that the source lists for the same document as both SUPPORTS and REFUTES are
not among them. This exclusion is not in the plan and changes no decision. Thus \prtLabNRequestsAnalyzed{} of the \prtNRequests{} requests enter the analysis. A claim with
$|E|>k$ has no KEEP with the same number of sentences. This leaves \prtFeverNEligible{}
of the \prtFeverNHidden{} FEVER hidden claims and all \prtEiNHidden{} on Evidence Inference. Every eligible
pair was answered, and none was refused; Table~\ref{tab:retention} gives the analyzed pairs ($n$).

\paragraph{Tests.} The estimand is the mean over documents of each document's mean correct(KEEP) $-$ correct(DROP), with 2000 document-bootstrap resamples. The plan makes two predictions per corpus, KEEP $-$ DROP $>0$ on all
hidden claims and on the disjoint ones, each tested one-sided at nominal level 0.05 by the share of bootstrap means at
or below zero; the overlapping claims are a secondary test. Both predictions are supported on both corpora (KD1 and KD2 on FEVER, KD3 and KD4 on Evidence Inference)
(Table~\ref{tab:retention}).

\begin{table}[t]
\centering\small
\setlength{\tabcolsep}{3pt}
\caption{KEEP beats DROP at 25\% (\texttt{gpt-6-luna}). Split by gold label, the gap is on true claims; false
claims are answered correctly at \prtLabFeverFalseAccKeep\% or more in every arm. KEEP $-$ DROP hidden-claim accuracy (points, document bootstrap
95\% interval; all and disjoint prespecified, overlapping secondary, label split post hoc) and document-macro
accuracy (\%) of KEEP, DROP and, descriptively, the main typed-selection, random and full-text reads of the
same claims, which asked several claims per request.}
\label{tab:retention}\label{tab:retention-label}
\begin{tabular}{llrrlrrrrr}
\toprule
corpus & claims & $n$ & docs & KEEP $-$ DROP & KEEP & DROP & typed & random & full \\
\midrule
FEVER & all & \prtFeverNPairsAnalyzed & \prtFeverKeepDropAllNDocs & \ci{\prtFeverKeepDropAllDelta}{\prtFeverKeepDropAllLo}{\prtFeverKeepDropAllHi} & \prtFeverAccKeepAll & \prtFeverAccDropAll & \prtFeverAccKnownAll & \prtFeverAccRandomAll & \prtFeverAccOriginalAll \\
 & disjoint & \prtFeverNDisjoint & \prtFeverKeepDropDisjointNDocs & \ci{\prtFeverKeepDropDisjointDelta}{\prtFeverKeepDropDisjointLo}{\prtFeverKeepDropDisjointHi} & \prtFeverAccKeepDisjoint & \prtFeverAccDropDisjoint & \prtFeverAccKnownDisjoint & \prtFeverAccRandomDisjoint & \prtFeverAccOriginalDisjoint \\
 & overlapping & \prtFeverNOverlap & \prtFeverKeepDropOverlapNDocs & \ci{\prtFeverKeepDropOverlapDelta}{\prtFeverKeepDropOverlapLo}{\prtFeverKeepDropOverlapHi} & \prtFeverAccKeepOverlap & \prtFeverAccDropOverlap & \prtFeverAccKnownOverlap & \prtFeverAccRandomOverlap & \prtFeverAccOriginalOverlap \\
 & true & \prtLabFeverTrueN & \prtLabFeverTrueNDocs & --- & \prtLabFeverTrueAccKeep & \prtLabFeverTrueAccDrop & \prtLabFeverTrueAccKnown & \prtLabFeverTrueAccRandom & \prtLabFeverTrueAccOriginal \\
 & false & \prtLabFeverFalseN & \prtLabFeverFalseNDocs & --- & \prtLabFeverFalseAccKeep & \prtLabFeverFalseAccDrop & \prtLabFeverFalseAccKnown & \prtLabFeverFalseAccRandom & \prtLabFeverFalseAccOriginal \\
\midrule
Ev.\ Inf. & all & \prtEiNPairsAnalyzed & \prtEiKeepDropAllNDocs & \ci{\prtEiKeepDropAllDelta}{\prtEiKeepDropAllLo}{\prtEiKeepDropAllHi} & \prtEiAccKeepAll & \prtEiAccDropAll & \prtEiAccKnownAll & \prtEiAccRandomAll & \prtEiAccOriginalAll \\
 & disjoint & \prtEiNDisjoint & \prtEiKeepDropDisjointNDocs & \ci{\prtEiKeepDropDisjointDelta}{\prtEiKeepDropDisjointLo}{\prtEiKeepDropDisjointHi} & \prtEiAccKeepDisjoint & \prtEiAccDropDisjoint & \prtEiAccKnownDisjoint & \prtEiAccRandomDisjoint & \prtEiAccOriginalDisjoint \\
 & overlapping & \prtEiNOverlap & \prtEiKeepDropOverlapNDocs & \ci{\prtEiKeepDropOverlapDelta}{\prtEiKeepDropOverlapLo}{\prtEiKeepDropOverlapHi} & \prtEiAccKeepOverlap & \prtEiAccDropOverlap & \prtEiAccKnownOverlap & \prtEiAccRandomOverlap & \prtEiAccOriginalOverlap \\
 & true & \prtLabEiTrueN & \prtLabEiTrueNDocs & --- & \prtLabEiTrueAccKeep & \prtLabEiTrueAccDrop & \prtLabEiTrueAccKnown & \prtLabEiTrueAccRandom & \prtLabEiTrueAccOriginal \\
 & false & \prtLabEiFalseN & \prtLabEiFalseNDocs & --- & \prtLabEiFalseAccKeep & \prtLabEiFalseAccDrop & \prtLabEiFalseAccKnown & \prtLabEiFalseAccRandom & \prtLabEiFalseAccOriginal \\
\bottomrule
\end{tabular}
\end{table}

\section{Token counts and input-token cost}
\label{app:cost}

\paragraph{Tokens kept at sentence and turn budgets.} Under the \texttt{o200k\_base} encoding, each policy keeps
within 5 points of the budget's share of the tokens, except TF-IDF selection on Evidence Inference
(\csEiTfidfTFTokShare\% at 25\%, \csEiTfidfFTokShare\% at 50\%) and question-agnostic selection on LoCoMo
(\csLocomoAgnosticTFTokShare\% and \csLocomoAgnosticFTokShare\%); typed selection keeps \csEiTypedTFTokShare\% and
\csEiTypedFTokShare\% on Evidence Inference. KEEP texts are longer than DROP texts: \csFeverKDKeep{} against
\csFeverKDDrop{} tokens on FEVER (\csFeverKDDiff{} [\csFeverKDDiffLo, \csFeverKDDiffHi]), \csEiKDKeep{} against
\csEiKDDrop{} on Evidence Inference (\csEiKDDiff{} [\csEiKDDiffLo, \csEiKDDiffHi]) and \csLocomoKDKeep{} against
\csLocomoKDDrop{} on LoCoMo (\csLocomoKDDiff{} [\csLocomoKDDiffLo, \csLocomoKDDiffHi]; document or conversation
bootstrap).

\paragraph{Input-token cost.} Table~\ref{tab:cost} counts \texttt{gpt-6-luna} input tokens per document. Scoring
and the batched reads, which ask all of a document's questions in one request, are measured. Per-question costs
are modeled for one-question requests by a token model fitted to each corpus's recorded requests (on MultiRC,
scoring and batched reads of about 60 questions each only, so its one-question costs are an extrapolation; mean absolute error \csFeverCalOneQuestionReadMAE\% on FEVER, \csEiCalOneQuestionReadMAE\% on Evidence Inference and
\csLocomoCalOneQuestionReadMAE\% on LoCoMo). With batched reads, scoring plus the 25\% read costs
\csMrcMeasRatio, \csFeverMeasRatio, \csEiMeasRatio{} and \csLocomoMeasRatio{} times the full read (MultiRC, FEVER,
Evidence Inference, LoCoMo). With one request per later question, compacting once costs \csMrcCompactRatioKbar,
\csFeverCompactRatioKbar, \csEiCompactRatioKbar{} and \csLocomoCompactRatioKbar{} times always reading the full text
at each corpus's mean number of later questions $\bar k$. Adding the sufficiency check and a full re-read for a
quarter of the questions raises this to \csMrcGateRatioKbar, \csFeverGateRatioKbar, \csEiGateRatioKbar{} and
\csLocomoGateRatioKbar{} with the check in the read request.

\begin{table}[t]
\centering\scriptsize
\setlength{\tabcolsep}{2.5pt}
\caption{Input tokens for typed selection at 25\% (\texttt{gpt-6-luna}). Top: per document, measured scoring and
batched reads, and modeled one-question costs (read, sufficiency check as a separate call or joint with the read,
and a full re-read charged at a quarter of the full read). Bottom: thousand input tokens per document to answer
$k$ later questions, one request each, the ratio to always reading the full text at $k = \bar k$, the corpus's mean
number of later questions, and the
break-even $k$, the number of later questions at which the policy and the full text cost the same (never: at no
$k$). Prompt caching is not modeled.}
\label{tab:cost}
\input{generated/p4_cost_table.tex}
\end{table}

\section{The LoCoMo agent-memory test}
\label{app:locomo}

\begin{table}[t]
\centering\small
\setlength{\tabcolsep}{3pt}
\caption{LoCoMo hidden-question accuracy at the last checkpoint (conversation-macro \%, chance 25) by memory, budget
and reader (luna: \texttt{gpt-6-luna}; qwen: \texttt{qwen3.8-flash} on the same memories), and the share of hidden
questions' evidence turns kept at 25\% (\%, question mean). Full and empty memory do not depend on the budget. Tests
H1--H4: under Test results below.}
\label{tab:locomo}
\begin{tabular}{lrrrr}
\toprule
& \multicolumn{3}{c}{accuracy (\%)} & evidence kept, \\
\cmidrule(lr){2-4}
memory & luna 25\% & luna 50\% & qwen 25\% & 25\% (\%) \\
\midrule
typed & \lcmAccTypedTFKD & \lcmAccTypedFKD & \lcmAccQwenTypedTFKD & \lcmTypedTFRetained \\
agnostic & \lcmAccAgnosticTFKD & \lcmAccAgnosticFKD & --- & \lcmAgnosticTFRetained \\
TF-IDF & \lcmAccTfidfTFKD & \lcmAccTfidfFKD & \lcmAccQwenTfidfTFKD & \lcmTfidfTFRetained \\
random & \lcmAccRandomTFKD & \lcmAccRandomFKD & \lcmAccQwenRandomTFKD & \lcmRandomTFRetained \\
recency & \lcmAccRecencyTFKD & \lcmAccRecencyFKD & \lcmAccQwenRecencyTFKD & \lcmRecencyTFRetained \\
lead & \lcmAccLeadTFKD & \lcmAccLeadFKD & --- & \lcmLeadTFRetained \\
full & \lcmAccFullKD & \lcmAccFullKD & \lcmAccQwenFullKD & --- \\
empty & \lcmAccNoneKD & \lcmAccNoneKD & \lcmAccQwenNoneKD & --- \\
\bottomrule
\end{tabular}
\end{table}

\paragraph{KEEP beats DROP.} At the last checkpoint, \texttt{gpt-6-luna} answers a hidden question more accurately from
the typed 25\% memory with its evidence (KEEP) than from one with the same number of turns without it (DROP). KEEP
$-$ DROP is \lcmHTwoADelta{} points, positive in all \lcmHTwoANConvPos{} conversations, on all questions and on the
\lcmHTwoBNQ{} evidence-disjoint ones, so H2a and H2b are supported. Read by \texttt{qwen3.8-flash}, the same texts
also meet H2's decision rule (\kdQwenAllDelta{} [\kdQwenAllLo, \kdQwenAllHi] on all questions). The four options
include no ``not supported'' answer, so missing evidence cannot make the reader right by default.

\paragraph{Typed versus random deletion is unresolved at 25\%.} With empty memory at \lcmAccNoneKD\%, absolute
differences are small. H1, typed $-$ random hidden accuracy at 25\%, is \lcmHOneVerdict{} for both readers, and so is
the typed $-$ random difference in evidence kept. H3, which asks the first to follow the second, therefore cannot be
assessed. H4 is supported: low sufficiency ranks hidden questions with dropped evidence first (AUROC
\lcmHFourDelta{} [\lcmHFourLo, \lcmHFourHi]). Question-agnostic selection keeps more of hidden questions' evidence turns than
typed selection (\lcmAgnosticTFRetained\% against \lcmTypedTFRetained\% at 25\%), but also longer turns
(Appendix~\ref{app:cost}). With harder answer options, in a prespecified follow-up, KEEP still beats DROP and typed
versus random stays unresolved (Table~\ref{tab:fu-d}).

\paragraph{Plan, data and cost.} Before any request of
this experiment, we specified the data, checkpoints, known-question rule, policies, read format, readers, the four
hypotheses with their tests, the bootstrap seed and the outcomes reported without a decision rule. Only item
construction and token estimation (no model answer) came first. The data are
\texttt{locomo10.json} from the authors' repository \citep{maharana2024locomo} at commit \texttt{3eb6f2c}, released under CC BY-NC 4.0; our use is non-commercial research, and
we use the released photo captions as text. \texttt{gpt-6-luna}
cost \$\lcmLunaUSD{} (\lcmLunaTokens{} input tokens) and \texttt{qwen3.8-flash} \$\lcmQwenUSD{} (\lcmQwenTokens{} prompt tokens).

\paragraph{Design.} Eligible questions are LoCoMo's categories 1--4 (multi-hop, temporal, open-domain,
single-hop) with a gold answer and at least one valid evidence turn, 1,536 in all; adversarial questions have
no gold answer. Memory is rendered as session blocks, each headed by its session's date, then one line per kept
turn with its speaker and any photo caption. Headers do not count against the budget, since temporal answers
are relative to session dates. With $S$ sessions, checkpoint $k$ closes session $\lceil kS/4\rceil$. The
known question at $k$ is, among unused eligible questions whose last evidence turn lies in segment $k$, the
one with the smallest CRC32 checksum of a fixed key. At checkpoint $k$ with $N_k$ turns seen, a policy keeps
$\mathrm{round}(bN_k)$ of the previous memory plus the new turns, and a dropped turn never returns. Typed and
agnostic use \texttt{gpt-6-luna} scores with the known- and agnostic-policy wording of the earlier corpora,
``sentence'' replaced by ``turn''; TF-IDF, random and lead are as before, recency keeps the latest turns. Hidden
questions at $k$ are every eligible question whose evidence lies before checkpoint $k$, except the
conversation's four known questions: 351, 671, 1,061 and 1,496 at $k=1,\dots,4$. The three distractors of a
question are answers to other questions of the same conversation, same category first, with near-duplicates
excluded and the option order fixed by a hash. \texttt{gpt-6-luna} reads up to 100 questions per request, each
with the Choice and the sufficiency check (``does the conversation memory contain the information needed to
answer this question?''). \texttt{qwen3.8-flash} answers one question per call from first-token
log-probabilities over the option letters (thinking off), at the last checkpoint, for typed, TF-IDF, random and
recency at 25\%, full and empty memory, on \texttt{gpt-6-luna}'s selections. A refusal counts as wrong;
dropping instead every question refused under either memory condition of a contrast changes no planned decision.

\paragraph{Tests.} Every estimate is the mean over conversations of the per-conversation mean paired
difference, with 2000 conversation-bootstrap resamples. H1 (two-sided) is positive or negative if the 95\%
interval excludes zero and unresolved otherwise; H2a and H2b (one-sided, nominal level 0.05) are supported if the 5th
percentile is above zero. KEEP is the typed 25\% memory at the last checkpoint with the question's missing
evidence added and as many non-evidence turns removed, lowest typed score first. DROP removes its kept evidence
and adds as many turns drawn at random from outside the memory and the evidence. Each of the 2,992 KEEP and DROP
reads asks one question, in one shuffled list. H3 is consistent or inconsistent only if H1 and the
typed $-$ random retention difference are both resolved. H4 is supported if the 2.5th percentile of a
conversation bootstrap of the pooled AUROC exceeds 0.5. With ten clusters the percentile bootstrap may
under-cover. Pooled over questions, KEEP scores \lcmKDAllKeep\% and DROP \lcmKDAllDrop\%.

\paragraph{Test results.} H1, typed $-$ random at 25\%, is \ci{\lcmHOneDelta}{\lcmHOneLo}{\lcmHOneHi} with
\texttt{gpt-6-luna} and \ci{\lcmHOneQwenDelta}{\lcmHOneQwenLo}{\lcmHOneQwenHi} with \texttt{qwen3.8-flash}, both
\lcmHOneVerdict. KEEP $-$ DROP is \ci{\lcmHTwoADelta}{\lcmHTwoALo}{\lcmHTwoAHi} on all hidden questions (H2a,
\lcmHTwoAVerdict) and \ci{\lcmHTwoBDelta}{\lcmHTwoBLo}{\lcmHTwoBHi} on evidence-disjoint ones (H2b, \lcmHTwoBVerdict).
The conversation-macro typed $-$ random difference in evidence kept is \ci{\lcmHThreeDelta}{\lcmHThreeLo}{\lcmHThreeHi}, so H3 is
\lcmHThreeVerdict. The sufficiency AUROC against dropped evidence is \lcmHFourDelta{}
[\lcmHFourLo, \lcmHFourHi] (H4, \lcmHFourVerdict).

\paragraph{Second reader.} In the plan \texttt{qwen3.8-flash} reads six memory conditions to replicate H1;
its KEEP--DROP reads are an exploratory extension under H2's decision rule. \texttt{qwen3.8-flash} read the same \kdQwenAllNQ{} KEEP and DROP pairs, one question
per call, with no refusals. KEEP $-$ DROP is \kdQwenAllDelta{} [\kdQwenAllLo, \kdQwenAllHi] points on all
questions (\kdQwenAllKeep\% vs.\ \kdQwenAllDrop\%; 5th percentile \kdQwenAllLoNinety), \kdQwenDisjointDelta{}
[\kdQwenDisjointLo, \kdQwenDisjointHi] on the \kdQwenDisjointNQ{} disjoint ones (5th percentile
\kdQwenDisjointLoNinety) and \kdQwenOverlapDelta{} [\kdQwenOverlapLo, \kdQwenOverlapHi] on the
\kdQwenOverlapNQ{} overlapping ones, positive in \kdQwenAllNConvPos{}, \kdQwenDisjointNConvPos{} and
\kdQwenOverlapNConvPos{} of ten conversations. Both 5th percentiles are above zero, so the H2a and H2b rules are also met on this reader. The accuracies in parentheses are
question-pooled and the contrasts conversation-macro, so they need not subtract exactly.

\paragraph{Empty memory already answers most questions.} The two readers answer \lcmAccNoneKD\% and
\lcmAccQwenNoneKD\% of hidden questions from empty memory (full memory \lcmAccFullKD\% and \lcmAccQwenFullKD\%). The
distractors are answers to other questions of the same conversation, and they can often be ruled out; the
prespecified follow-up with harder options (below) lowers this floor.

\begin{table}[t]
\centering\small
\caption{With harder options, \texttt{gpt-6-luna} refuses more with empty memory, KEEP stays far above DROP, and
typed versus random stays unresolved. LoCoMo hidden questions at the last checkpoint, conversation-macro
(\%). Refused: share of questions refused. Accuracy: over all questions, refusal counts as wrong (chance 25). Answered acc.: over answered questions. First row: empty memory with the
original options, on the same questions. KEEP and DROP are the texts of Section~\ref{sec:locomo}.
\texttt{qwen3.8-flash} did not read the full memory. }
\label{tab:fu-d}
\input{generated/p4fu_d.tex}
\end{table}

\paragraph{Harder options (follow-up D).} With harder options, empty memory drops to \fuDLunaNoneAcc\% (\texttt{gpt-6-luna}) and \fuDQwenNoneAcc\% (\texttt{qwen3.8-flash}; Table~\ref{tab:fu-d}). Only the three wrong options of each question change; questions,
memories and KEEP/DROP texts are those of Section~\ref{sec:locomo}. For temporal questions and questions asking how
many, how long or when, the wrong options are the gold answer with its date shifted by one to four weeks (same
month first) or its number shifted (\fuDShareShifted\% of wrong options). Otherwise they are answers to questions
of the same form (same head noun or opening words), ranked by TF-IDF similarity of the questions. They come from other
conversations if they name no speaker (\fuDShareOther\%), or from the same conversation if their question is not
a near duplicate (\fuDShareSame\%). Yes/no answers keep the original options (\fuDShareOriginal\%). The plan
set a target of at most 60\% question-pooled empty-memory accuracy for \texttt{gpt-6-luna}. The construction reaches a question-pooled \fuDPooledHardAcc\%
(original options \fuDPooledOrigAcc\%), missing the target by 0.2 points. Table~\ref{tab:fu-d} gives, per memory and
reader, the refusal rate, the accuracy over all questions (refusals wrong) and the accuracy over answered questions.
Full-memory accuracy falls to \fuDLunaFullAcc\% (original options: \lcmAccFullKD\%), since a shifted date can be hard
to rule out even with the whole conversation. The plan applies the rules of H1 (D1) and H2a and H2b (D2a, D2b). Typed $-$ random at 25\% is
\fuDLunaTypedTFVsRandomTFAllDelta{} [\fuDLunaTypedTFVsRandomTFAllLo, \fuDLunaTypedTFVsRandomTFAllHi]
(\texttt{gpt-6-luna}) and \fuDQwenTypedTFVsRandomTFAllDelta{} [\fuDQwenTypedTFVsRandomTFAllLo,
\fuDQwenTypedTFVsRandomTFAllHi] (\texttt{qwen3.8-flash}), unresolved. KEEP $-$ DROP is
\fuDLunaKEEPVsDROPAllDelta{} [\fuDLunaKEEPVsDROPAllLo, \fuDLunaKEEPVsDROPAllHi] and \fuDQwenKEEPVsDROPAllDelta{}
[\fuDQwenKEEPVsDROPAllLo, \fuDQwenKEEPVsDROPAllHi], and on evidence-disjoint questions
\ci{\fuDLunaKEEPVsDROPDisjointDelta}{\fuDLunaKEEPVsDROPDisjointLo}{\fuDLunaKEEPVsDROPDisjointHi} and \fuDQwenKEEPVsDROPDisjointDelta{}, supported for both readers. Dropping refused answers (exploratory) leaves every decision unchanged.

\section{The $\tau$-bench test}
\label{app:tau}

\paragraph{Selectors and results.} The sample has \taNStates{} states from \taNTasks{} tasks with
\taNHidden{} hidden questions, excluding the trajectories of a pilot. Budgets keep 25\% of the tokens, and intervals
bootstrap tasks. Besides the selectors above, an \emph{oracle} keeps the units most hidden questions need. Two
selectors anticipate later calls. \emph{Anticipatory} selection keeps units needed by any of ten likely later
questions that \texttt{qwen3.8-flash} writes from the state, and \emph{expected-coverage} selection (ECov) weighs
those questions against the known one (below). Of \taNPred{} prespecified predictions, \taNMet{} are
met (Table~\ref{tab:audit}).

\begin{table}[t]
\centering\small
\setlength{\tabcolsep}{4pt}
\caption{On $\tau$-bench retail trajectories at 25\% of the tokens, typed selection beats random deletion on hidden
questions, \LLMLtwo{} does not, and anticipating later calls costs the known question. Accuracy (\%, state means)
with \texttt{gpt-6-luna} (luna) and \texttt{qwen3.8-flash} (qwen); ECov: expected-coverage selection; id kept: share of hidden questions whose gold
identifier survives; $G$: share of the random--oracle gap closed on hidden questions (luna). $^\dagger$On a \taNECovStates-state subsample, where typed selection scores \taECovSubTypedH{}
(hidden) and \taECovSubTypedK{} (known). Further arms are on all states; their contrasts with ECov (M4, M5 in Appendix~\ref{app:tau}) use the ECov subsample.}
\label{tab:tau}\label{tab:tau-arms}
\begin{tabular}{lrrrrr}
\toprule
text & hidden, luna & known, luna & hidden, qwen & id kept & $G$ \\
\midrule
full state & \taAccFullH & \taAccFullK & \taQwenAccFullH & --- & --- \\
oracle & \taAccOracleH & \taAccOracleK & \taQwenAccOracleH & \taGoldOracleH & 1.00 \\
anticipatory & \taAccAntH & \taAccAntK & \taQwenAccAntH & \taGoldAntH & \taGAnt \\
ECov$^\dagger$ & \taAccECovTH & \taAccECovTK & \taQwenAccECovTH & \taGoldECovTH & \taGECovT \\
typed & \taAccTypedH & \taAccTypedK & \taQwenAccTypedH & \taGoldTypedH & \taGTyped \\
recency & \taAccRecencyH & \taAccRecencyK & \taQwenAccRecencyH & \taGoldRecencyH & \taGRecency \\
TF-IDF & \taAccTfidfH & \taAccTfidfK & --- & \taGoldTfidfH & \taGTfidf \\
\LLMLtwo{} & \taAccLinguaH & \taAccLinguaK & \taQwenAccLinguaH & \taGoldLinguaH & \taGLingua \\
random & \taAccRandomH & \taAccRandomK & \taQwenAccRandomH & \taGoldRandomH & 0.00 \\
empty & \taAccEmptyH & \taAccEmptyK & \taQwenAccEmptyH & --- & --- \\
\midrule
\multicolumn{6}{@{}l}{\textit{further arms ($G$ with 95\% interval; typed: \taGTyped{} [\taGTypedLo, \taGTypedHi])}}\\
typed, newest-first & \taAccTypedRecH & \taAccTypedRecK & --- & --- & \taGTypedRec{} [\taGTypedRecLo, \taGTypedRecHi] \\
ECov-embed & \taAccECovEH & \taAccECovEK & --- & --- & \taGECovE{} [\taGECovELo, \taGECovEHi] \\
hybrid & \taAccHybridH & \taAccHybridK & --- & --- & \taGHybrid{} [\taGHybridLo, \taGHybridHi] \\
gist & \taAccGistH & \taAccGistK & --- & --- & \taGGist{} [\taGGistLo, \taGGistHi] \\
\bottomrule
\end{tabular}
\end{table}

\paragraph{Typed selection beats random deletion on hidden questions.} Typed $-$ random is
\ci{\taTypedVsRandomHDelta}{\taTypedVsRandomHLo}{\taTypedVsRandomHHi} points with \texttt{gpt-6-luna} and
\ci{\taQwenTypedVsRandomHDelta}{\taQwenTypedVsRandomHLo}{\taQwenTypedVsRandomHHi} with \texttt{qwen3.8-flash}
(Table~\ref{tab:tau}), unlike on MultiRC. Typed selection also keeps the hidden question's
identifier more often (\taGoldTypedH\% against \taGoldRandomH\%), and hidden questions whose identifier survives score
\taGoldKeptAcc\%, against \taGoldLostAcc\% when it is lost. This association is consistent with the relation between gain and evidence kept of Section~\ref{sec:mechanism}. \LLMLtwo{} keeps the identifier for \taGoldLinguaH\% and
shows no detected gain (\ci{\taLinguaVsRandomHDelta}{\taLinguaVsRandomHLo}{\taLinguaVsRandomHHi}). Recency, which keeps
the newest units, gains \taRecencyVsRandomHDelta{} points on hidden questions but scores only \taAccRecencyK\% on the
known one, because the identifiers sit in earlier units.

\paragraph{Anticipating later calls costs the known question.} ECov $-$ typed is
\ci{\taECovTVsTypedHDelta}{\taECovTVsTypedHLo}{\taECovTVsTypedHHi} on hidden questions but \taECovTVsTypedKDelta{}
(90\% interval [\taECovTVsTypedKLoNinety, \taECovTVsTypedKHiNinety]) on the known one; anticipatory selection gives
\taAntVsTypedHDelta{} and \taAntVsTypedKDelta. No selector we tested beats typed selection on both. Fetching dropped
units on demand also stays well below the full state at about the same passage tokens (recoverable arm below).

\paragraph{Sample and questions.} The trajectories are the successful (reward 1) retail trajectories of the two
agents whose histories $\tau$-bench releases (MIT license); airline trajectories (\taNAirStates{} states) are descriptive only: \texttt{gpt-6-luna} scores
\taAirAccFullH\% (full), \taAirAccTypedH\% (typed), \taAirAccRandomH\% (random) and \taAirAccEmptyH\% (empty) on its
hidden questions. The plan, with its predictions, was written before any request. The
\taNPilot{} pilot trajectories were inspected to choose the newest-first tie-break of anticipatory selection and are
excluded. The checkpoint is the middle tool output. Questions whose gold identifier is ambiguous, not identifier-like
or a bare list entry are dropped, and in quoted text every substring with the gold's character pattern is masked. In six retail hidden questions a later user
turn gives the identifier without its leading ``\#'', which this mask misses; dropping them moves typed $-$ random
and \LLMLtwo{} $-$ random by at most 0.2 points. A
known question exists when the next call's identifier argument occurs in the state with at least one same-shaped
distractor (\taNKnown{} known questions, one per identifier argument of the next call). A refusal counts as wrong; the full-state refusal rate is
\taRefusalRate\% (95\% upper limit \taRefusalUpper\%), so the prespecified validity check \taRefusalPass. Other arms refuse more
often (retail, \texttt{gpt-6-luna}): random \taRefusalRandom\%, \LLMLtwo{} \taRefusalLingua\%, typed \taRefusalTyped\%,
anticipatory \taRefusalAnt\%, ECov \taRefusalECovT\% and empty \taRefusalEmpty\%. On answered questions only, typed $-$
random is \ci{\taAnsTypedVsRandomHDelta}{\taAnsTypedVsRandomHLo}{\taAnsTypedVsRandomHHi}, \LLMLtwo{} $-$ random
\ci{\taAnsLinguaVsRandomHDelta}{\taAnsLinguaVsRandomHLo}{\taAnsLinguaVsRandomHHi}, ECov $-$ typed
\ci{\taAnsECovTVsTypedHDelta}{\taAnsECovTVsTypedHLo}{\taAnsECovTVsTypedHHi}, anticipatory $-$ typed
\ci{\taAnsAntVsTypedHDelta}{\taAnsAntVsTypedHLo}{\taAnsAntVsTypedHHi} and recency $-$ random
\ci{\taAnsRecencyVsRandomHDelta}{\taAnsRecencyVsRandomHLo}{\taAnsRecencyVsRandomHHi} points on hidden questions.

\paragraph{Selectors.} Budgets are $T = \mathrm{round}(0.25 \times$ the state's o200k tokens$)$, filled unit by unit in
rank order. Typed selection ranks units by \texttt{gpt-6-luna}'s probability that the unit is needed for the known
question; \emph{typed, newest-first} breaks its ties toward newer units. ECov adds units greedily by their gain in
$F(S) = \sum_q w_q\,[1 - \prod_{u \in S}(1 - p(u, q))]$, with weight 0.5 on the known question ($p$ = the typed
probability) and 0.05 on each of the ten generated questions ($p$ from a \texttt{gpt-6-luna} Boolean per unit and
question). \emph{ECov-embed} replaces the latter by a logistic function of the all-MiniLM-L6-v2 cosine between unit and
question. \emph{Hybrid} keeps typed whole units up to half the budget and fills the rest with \LLMLtwo{} words.
\emph{Gist} writes every unit as a numbered line holding its top \LLMLtwo{} words within 60\% of the budget, then
restores typed whole units. Table~\ref{tab:tau-arms} gives these arms; their predictions are
under Prespecified predictions below. Under recency the newest unit takes a median \taRecencyLastShare\% of the budget, and the rest of its budget goes to the next-newest units.

\paragraph{Prespecified predictions.} Hidden questions unless stated (M1--M4 on the ECov subsample): ECov $-$ typed is
\ci{\taECovTVsTypedHDelta}{\taECovTVsTypedHLo}{\taECovTVsTypedHHi} (M1, at least $+5$ with the interval above 0:
\taMOneMet); on the known question it is \taECovTVsTypedKDelta{} [\taECovTVsTypedKLoNinety, \taECovTVsTypedKHiNinety]
(90\%; M2, lower end above $-3$: \taMTwoMet); ECov closes $G = \taGECovT$ of the random--oracle gap (M3, $G \geq 0.70$: \taMThreeMet); ECov-embed $-$ ECov is \taECovEVsECovTHDelta{} [\taECovEVsECovTHLoNinety,
\taECovEVsECovTHHiNinety] (90\%; M4, within $\pm 3$: \taMFourMet); ECov $-$ gist is
\ci{\taECovTVsGistHDelta}{\taECovTVsGistHLo}{\taECovTVsGistHHi} (M5, above 0: \taMFiveMet); the recoverable gist
arm is below (M6, at least 95\% of full-state accuracy at no more than 60\% of its tokens: \taMSixMet). Typed $-$
random is \ci{\taTypedVsRandomHDelta}{\taTypedVsRandomHLo}{\taTypedVsRandomHHi} (T1: \taTOneMet), recency $-$
random \ci{\taRecencyVsRandomHDelta}{\taRecencyVsRandomHLo}{\taRecencyVsRandomHHi} (T2: \taTTwoMet) and
anticipatory $-$ random \ci{\taAntVsRandomHDelta}{\taAntVsRandomHLo}{\taAntVsRandomHHi} (T3: \taTThreeMet).
Questions whose identifier is kept score \taGoldKeptVsLost{} points above those whose identifier is lost (T4, at
least 40: \taTFourMet). \LLMLtwo{} keeps fewer identifiers than typed selection (\taGoldLinguaH\% against
\taGoldTypedH\%) and \LLMLtwo{} $-$ typed is \taLinguaVsTypedHDelta{} points (T6: \taTSixMet). Hybrid $-$ typed is
\taHybridVsTypedHDelta{} [\taHybridVsTypedHLoNinety, \taHybridVsTypedHHiNinety] (90\%; T7, lower end above $-2$:
\taTSevenMet). With \texttt{qwen3.8-flash} typed $-$ random has the same sign as with \texttt{gpt-6-luna} (T9:
\taTNineMet).

\paragraph{Recoverable arm.} Dropped units leave either a stub naming the run of dropped units and their roles, or the
gist line above, so each unit keeps its position. A hidden question whose sufficiency is below a threshold $\theta$
is re-read alone with dropped units added in order of TF-IDF match between the question and what remains visible of
them, up to a quarter of the full state's tokens. With gists at $\theta = 0.5$ it reaches \taRecGistPtrAccB\% against
\taFullItemAcc\% for the full state, at \taRecGistPtrTokShareB\% of the full state's passage tokens (o200k, without the
question and option text) per question
(Table~\ref{tab:tau-rec}). Low sufficiency ranks questions whose identifier was dropped first (AUROC
\taRecTypedStubAUROC{} with stubs, \taRecGistPtrAUROC{} with gists), so the main failure is choosing the units to fetch. The first pass reads all questions of a state plus the sufficiency
check (Section~\ref{sec:gate}) for each hidden question; it scores \taRecTypedStubFirst\% on hidden questions with stubs
and \taRecGistPtrFirst\% with gists. Tokens per question count the first read divided by the state's hidden questions,
plus the re-read. Table~\ref{tab:tau-rec} gives question-mean accuracy; the full state scores \taFullItemAcc\% on the
same questions.

\begin{table}[t]
\centering\small
\caption{Recoverable arm (retail, 25\%, \texttt{gpt-6-luna}): hidden accuracy (\%), share of hidden questions re-read
(\%) and passage tokens per question as \% of the full state's, by threshold $\theta$.}
\label{tab:tau-rec}
\begin{tabular}{lrrrrrr}
\toprule
& \multicolumn{3}{c}{stubs} & \multicolumn{3}{c}{gists} \\
$\theta$ & acc. & re-read & tokens & acc. & re-read & tokens \\
\midrule
0.3 & \taRecTypedStubAccA & \taRecTypedStubFetchedA & \taRecTypedStubTokShareA & \taRecGistPtrAccA & \taRecGistPtrFetchedA & \taRecGistPtrTokShareA \\
0.5 & \taRecTypedStubAccB & \taRecTypedStubFetchedB & \taRecTypedStubTokShareB & \taRecGistPtrAccB & \taRecGistPtrFetchedB & \taRecGistPtrTokShareB \\
0.7 & \taRecTypedStubAccC & \taRecTypedStubFetchedC & \taRecTypedStubTokShareC & \taRecGistPtrAccC & \taRecGistPtrFetchedC & \taRecGistPtrTokShareC \\
\bottomrule
\end{tabular}
\end{table}

\section{Abstractive summaries}
\label{app:summary}

\paragraph{Design.} Paragraphs, known questions, reader (\texttt{gpt-6-luna}) and the 25\% token budget $T$ are those
of the hybrid test (Appendix~\ref{app:hybrid}), and random sentence deletion, typed selection and \LLMLtwo{} are that
test's reads. Each summarizer writes two summaries per paragraph. The generic prompt asks for a summary ``for a
reader who will later be asked questions about it that are not known yet''; the focused prompt asks for one ``so that
the summary answers the question'', followed by the known question. \texttt{qwen3.8-flash} (temperature 0) is asked
for at most $0.75T$ words; \texttt{gpt-6.1-sol} (temperature 1, reasoning effort low) is
asked for about $T/1.25$ words, close to that length. A summary longer than $T$ tokens is cut to its first $T$
tokens. The provider's input filter rejected \texttt{qwen3.8-flash} requests for \smSumFocNoText{} paragraphs
(focused) and \smSumGenNoText{} (generic); these drop out of that arm's contrasts. Estimators are the follow-ups'
(Appendix~\ref{app:followup}).

\paragraph{Plan and results.} A plan written before any request fixed the prompts, the contrasts of
Table~\ref{tab:summary} and two predictions for \texttt{qwen3.8-flash}: the generic summary beats random sentence
deletion on hidden questions (S1), and the focused summary beats it on the known question (S2). A second plan,
written after that run and before any \texttt{gpt-6.1-sol} summary, added \texttt{gpt-6.1-sol} with S1 and S2, the
prediction that its generic summary stays below \LLMLtwo{} on hidden questions (S3), and the prediction that both its
summaries use a larger share of $T$ than \texttt{qwen3.8-flash}'s (S4). All six are met (Table~\ref{tab:summary}).
On hidden questions \texttt{gpt-6.1-sol} $-$ \texttt{qwen3.8-flash} is
\ci{\smGptGenSumGenHiddenDelta}{\smGptGenSumGenHiddenLo}{\smGptGenSumGenHiddenHi} points for the generic summaries
and \ci{\smGptFocSumFocHiddenDelta}{\smGptFocSumFocHiddenLo}{\smGptFocSumFocHiddenHi} for the focused ones; since
\texttt{gpt-6.1-sol} also fills more of the budget, the writer and the length are not separated. Of its generic
summaries, \smGptGenCutShare\% were cut at $T$ tokens.

\begin{table}[ht]
\centering\footnotesize
\setlength{\tabcolsep}{2.5pt}
\caption{Abstractive summaries on MultiRC at 25\% of the tokens, read by \texttt{gpt-6-luna}. Left: known and hidden
accuracy (\%), share of $T$ used and share of summaries cut at $T$ (\%). Right: summary $-$ comparator on hidden
questions, and $-$ random deletion on the known question (points, 95\% interval).}
\label{tab:summary}
\begin{tabular}{lrrrrlll}
\toprule
compaction & known & hidden & $T$ used & cut & $-$ random, hidden & $-$ \LLMLtwo{}, hidden & $-$ random, known \\
\midrule
random sentences & \smRandSentKnown & \smRandSentHidden & & & & & \\
typed & \smTypedKnown & \smTypedHidden & & & & & \\
\LLMLtwo{} & \smLinguaKnown & \smLinguaHidden & & & & & \\
\multicolumn{8}{l}{\textit{\texttt{qwen3.8-flash}}} \\
generic & \smSumGenKnown & \smSumGenHidden & \smSumGenTokShare & \smSumGenCutShare & \ci{\smSumGenRandSentHiddenDelta}{\smSumGenRandSentHiddenLo}{\smSumGenRandSentHiddenHi} & \ci{\smSumGenLinguaHiddenDelta}{\smSumGenLinguaHiddenLo}{\smSumGenLinguaHiddenHi} & \ci{\smSumGenRandSentKnownDelta}{\smSumGenRandSentKnownLo}{\smSumGenRandSentKnownHi} \\
focused & \smSumFocKnown & \smSumFocHidden & \smSumFocTokShare & \smSumFocCutShare & \ci{\smSumFocRandSentHiddenDelta}{\smSumFocRandSentHiddenLo}{\smSumFocRandSentHiddenHi} & \ci{\smSumFocLinguaHiddenDelta}{\smSumFocLinguaHiddenLo}{\smSumFocLinguaHiddenHi} & \ci{\smSumFocRandSentKnownDelta}{\smSumFocRandSentKnownLo}{\smSumFocRandSentKnownHi} \\
\multicolumn{8}{l}{\textit{\texttt{gpt-6.1-sol}}} \\
generic & \smGptGenKnown & \smGptGenHidden & \smGptGenTokShare & \smGptGenCutShare & \ci{\smGptGenRandSentHiddenDelta}{\smGptGenRandSentHiddenLo}{\smGptGenRandSentHiddenHi} & \ci{\smGptGenLinguaHiddenDelta}{\smGptGenLinguaHiddenLo}{\smGptGenLinguaHiddenHi} & \ci{\smGptGenRandSentKnownDelta}{\smGptGenRandSentKnownLo}{\smGptGenRandSentKnownHi} \\
focused & \smGptFocKnown & \smGptFocHidden & \smGptFocTokShare & \smGptFocCutShare & \ci{\smGptFocRandSentHiddenDelta}{\smGptFocRandSentHiddenLo}{\smGptFocRandSentHiddenHi} & \ci{\smGptFocLinguaHiddenDelta}{\smGptFocLinguaHiddenLo}{\smGptFocLinguaHiddenHi} & \ci{\smGptFocRandSentKnownDelta}{\smGptFocRandSentKnownLo}{\smGptFocRandSentKnownHi} \\
\bottomrule
\end{tabular}
\end{table}

\end{document}

%% file: generated/numbers3.tex
\newcommand{\CpNParas}{300}
\newcommand{\CpNQuestions}{3,355}
\newcommand{\CpNAnswers}{17,895}
\newcommand{\CpNSentLo}{6}
\newcommand{\CpNSentHi}{20}

\newcommand{\ErNQuestions}{3,300}

\newcommand{\CpDecNParas}{300}
\newcommand{\CpDecOrigKnown}{74}

\newcommand{\CpDecKnownTFKnown}{70}

\newcommand{\CpDecRandomTFKnown}{61}

\newcommand{\CpDecSFNParasLo}{166}
\newcommand{\CpDecSFNParasHi}{171}
\newcommand{\CpDecKnownVsrandomTFHiddenDelta}{\ensuremath{-}0.6}
\newcommand{\CpDecKnownVsrandomTFHiddenLo}{\ensuremath{-}1.9}
\newcommand{\CpDecKnownVsrandomTFHiddenHi}{+0.7}

\newcommand{\CpDecKnownVsrandomFHiddenDelta}{\ensuremath{-}0.6}
\newcommand{\CpDecKnownVsrandomFHiddenLo}{\ensuremath{-}1.7}
\newcommand{\CpDecKnownVsrandomFHiddenHi}{+0.6}

\newcommand{\ErDecKnownTFHiddenSurvive}{9}

\newcommand{\ErDecKnownTFRandHiddenSurvive}{4}

\newcommand{\ErDecKnownTFFlipBroken}{36}
\newcommand{\ErDecKnownTFFlipSurvived}{14}
\newcommand{\ErDecKnownTFRiskDiff}{+22}
\newcommand{\ErDecKnownTFRiskLo}{+19}
\newcommand{\ErDecKnownTFRiskHi}{+25}

\newcommand{\ErDecRandomTFFlipBroken}{35}
\newcommand{\ErDecRandomTFFlipSurvived}{11}

\newcommand{\ErDecRandExpKnownRecall}{25}

\newcommand{\GtDecKnownTFAurocBroken}{0.81}
\newcommand{\GtDecKnownTFAurocFlip}{0.69}
\newcommand{\GtDecKnownTFAurocFlipLo}{0.67}
\newcommand{\GtDecKnownTFAurocFlipHi}{0.71}

\newcommand{\GtDecKnownTFBrokenRate}{91}
\newcommand{\GtDecKnownTFAccCompact}{62}
\newcommand{\GtDecKnownTFAccOrig}{76}

\newcommand{\GtDecKnownTFGainTwentyFiveLo}{+1.2}
\newcommand{\GtDecKnownTFGainTwentyFiveHi}{+2.2}

\newcommand{\GtDecKnownTFGatedTwentyFive}{67}
\newcommand{\GtDecKnownTFRandomTwentyFive}{66}
\newcommand{\GtDecKnownTFGainTwentyFive}{+1.7}
\newcommand{\GtDecKnownTFGatedFifty}{71}

\newcommand{\GtDecKnownTFGainFifty}{+2.1}

\newcommand{\CpOaiOrigKnown}{89}

\newcommand{\CpOaiKnownTFKnown}{77}

\newcommand{\CpOaiRandomTFKnown}{67}

\newcommand{\CpOaiKnownVsrandomTFHiddenDelta}{\ensuremath{-}0.8}
\newcommand{\CpOaiKnownVsrandomTFHiddenLo}{\ensuremath{-}1.9}
\newcommand{\CpOaiKnownVsrandomTFHiddenHi}{+0.3}

\newcommand{\CpOaiKnownVsrandomFHiddenDelta}{\ensuremath{-}0.4}
\newcommand{\CpOaiKnownVsrandomFHiddenLo}{\ensuremath{-}1.7}
\newcommand{\CpOaiKnownVsrandomFHiddenHi}{+0.9}

\newcommand{\ErOaiKnownTFKnownRecall}{50}

\newcommand{\ErOaiKnownTFHiddenSurvive}{5}

\newcommand{\ErOaiKnownTFFlipBroken}{29}
\newcommand{\ErOaiKnownTFFlipSurvived}{5}
\newcommand{\ErOaiKnownTFRiskDiff}{+24}
\newcommand{\ErOaiKnownTFRiskLo}{+21}
\newcommand{\ErOaiKnownTFRiskHi}{+26}

\newcommand{\ErOaiRandomTFFlipBroken}{28}
\newcommand{\ErOaiRandomTFFlipSurvived}{5}

\newcommand{\GtOaiKnownTFAurocBroken}{0.83}
\newcommand{\GtOaiKnownTFAurocFlip}{0.71}
\newcommand{\GtOaiKnownTFAurocFlipLo}{0.69}
\newcommand{\GtOaiKnownTFAurocFlipHi}{0.73}

\newcommand{\GtOaiKnownTFBrokenRate}{96}
\newcommand{\GtOaiKnownTFAccCompact}{67}
\newcommand{\GtOaiKnownTFAccOrig}{89}

\newcommand{\GtOaiKnownTFGainTwentyFiveLo}{+1.1}
\newcommand{\GtOaiKnownTFGainTwentyFiveHi}{+1.8}

\newcommand{\GtOaiKnownTFGatedTwentyFive}{74}
\newcommand{\GtOaiKnownTFRandomTwentyFive}{73}
\newcommand{\GtOaiKnownTFGainTwentyFive}{+1.5}
\newcommand{\GtOaiKnownTFGatedFifty}{81}

\newcommand{\GtOaiKnownTFGainFifty}{+2.9}

\newcommand{\CpOaiNRefused}{710}
\newcommand{\CpOaiNRefusedReq}{290}
\newcommand{\CpOaiNChunkedReq}{38}
\newcommand{\CpOaiChunkSize}{125}

\newcommand{\GtOaiRepNItems}{100}
\newcommand{\GtOaiRepNAnswers}{6,603}

%% file: generated/restore.tex
\newcommand{\RsParas}{297}
\newcommand{\RsMeanAdded}{7.4}

\newcommand{\RsRecoveredDiff}{81}
\newcommand{\RsDiffLo}{78}
\newcommand{\RsDiffHi}{84}

\newcommand{\RsPlaceboMeanAdded}{2.6}

\newcommand{\RsDoseNewParas}{166}

\newcommand{\RsConcEligible}{297}
\newcommand{\RsConcSkipped}{61}

\newcommand{\RsConcAnswers}{11,296}
\newcommand{\RsConcFlips}{2,795}

\newcommand{\RsConcExcluded}{166}
\newcommand{\RsConcRecovered}{46\%}
\newcommand{\RsConcRecoveredPlacebo}{20\%}
\newcommand{\RsConcDiff}{27}
\newcommand{\RsConcDiffLo}{23}
\newcommand{\RsConcDiffHi}{32}

\newcommand{\RsConcDiffDec}{26.8}
\newcommand{\RsConcDiffLoDec}{22.6}
\newcommand{\RsConcDiffHiDec}{31.5}
\newcommand{\RsConcAllBackFlips}{801}
\newcommand{\RsConcAllBackRecovered}{86\%}
\newcommand{\RsConcAllBackRecoveredPlacebo}{21\%}
\newcommand{\RsConcAllBackDiff}{65}
\newcommand{\RsConcAllBackDiffLo}{59}
\newcommand{\RsConcAllBackDiffHi}{71}
\newcommand{\RsConcPartialFlips}{1,994}
\newcommand{\RsConcPartialRecovered}{31\%}
\newcommand{\RsConcPartialRecoveredPlacebo}{19\%}
\newcommand{\RsConcAccCompact}{68\%}
\newcommand{\RsConcAccRestore}{78\%}
\newcommand{\RsConcAccPlacebo}{71\%}
\newcommand{\RsConcAccFull}{89\%}

%% file: generated/p4_extra.tex
\newcommand{\pqMargin}{2}

%% file: generated/compress_fever.tex
\newcommand{\cfNDocs}{200}

\newcommand{\cfNHidden}{5,433}

\newcommand{\cfNCoreReadRequests}{1,809}
\newcommand{\cfNScoreRequests}{400}
\newcommand{\cfNCoreClaimAnswers}{50,733}
\newcommand{\cfNCoreSuffAnswers}{50,733}
\newcommand{\cfNCoreRefused}{0}
\newcommand{\cfNScoreAnswers}{5,880}
\newcommand{\cfNScoreRefused}{0}
\newcommand{\cfCoreInputTokens}{20,142,083}
\newcommand{\cfMeanSent}{14.7}
\newcommand{\cfMeanClaims}{28.2}
\newcommand{\cfSupportsShare}{71}
\newcommand{\cfOrigKnown}{92}

\newcommand{\cfKnownTFKnown}{92}

\newcommand{\cfRandomTFKnown}{56}

\newcommand{\cfKnownRandomTFHiddenDelta}{+17.2}
\newcommand{\cfKnownRandomTFHiddenLo}{+13.4}
\newcommand{\cfKnownRandomTFHiddenHi}{+21.2}
\newcommand{\cfAgnosticRandomTFHiddenDelta}{+2.1}
\newcommand{\cfAgnosticRandomTFHiddenLo}{\ensuremath{-}1.7}
\newcommand{\cfAgnosticRandomTFHiddenHi}{+5.8}

\newcommand{\cfKnownRandomTFHiddenDisjointDelta}{+0.2}
\newcommand{\cfKnownRandomTFHiddenDisjointLo}{\ensuremath{-}4.8}
\newcommand{\cfKnownRandomTFHiddenDisjointHi}{+5.2}

\newcommand{\cfKnownRandomFHiddenDelta}{+10.0}
\newcommand{\cfKnownRandomFHiddenLo}{+6.4}
\newcommand{\cfKnownRandomFHiddenHi}{+13.7}
\newcommand{\cfAgnosticRandomFHiddenDelta}{+3.5}
\newcommand{\cfAgnosticRandomFHiddenLo}{\ensuremath{-}0.2}
\newcommand{\cfAgnosticRandomFHiddenHi}{+7.1}

\newcommand{\cfKnownRandomFHiddenDisjointDelta}{\ensuremath{-}1.1}
\newcommand{\cfKnownRandomFHiddenDisjointLo}{\ensuremath{-}5.6}
\newcommand{\cfKnownRandomFHiddenDisjointHi}{+3.2}

\newcommand{\cfKnownTFFlipBroken}{45}
\newcommand{\cfKnownTFFlipSurvived}{1}

\newcommand{\cfKnownTFAurocBroken}{0.77}
\newcommand{\cfKnownTFAurocFlip}{0.87}
\newcommand{\cfKnownTFAurocFlipLo}{0.85}
\newcommand{\cfKnownTFAurocFlipHi}{0.88}
\newcommand{\cfKnownTFGainTwentyFive}{+8.9}
\newcommand{\cfKnownTFGainTwentyFiveLo}{+8.3}
\newcommand{\cfKnownTFGainTwentyFiveHi}{+9.4}
\newcommand{\cfKnownTFAccCompact}{71}
\newcommand{\cfKnownTFGatedTwentyFive}{86}
\newcommand{\cfKnownTFRandomTwentyFive}{77}

%% file: generated/p4_overlap.tex
\newcommand{\poFeverOverlapShare}{51}
\newcommand{\poFeverMinSent}{8}
\newcommand{\poFeverMaxSent}{40}
\newcommand{\poFeverMinClaims}{3}
\newcommand{\poFeverMaxQuestions}{200}
\newcommand{\poFeverClaimsPerRequest}{100}

\newcommand{\poOaiTFOverlapRecall}{36}

\newcommand{\poOaiTFDisjointRecall}{18}
\newcommand{\poOaiTFDisjointRandRecall}{25}
\newcommand{\poDecTFOverlapRecall}{54}
\newcommand{\poDecTFOverlapRandRecall}{25}
\newcommand{\poDecTFDisjointRecall}{14}
\newcommand{\poDecTFDisjointRandRecall}{25}
\newcommand{\poFeverTFOverlapRecall}{76}
\newcommand{\poFeverTFOverlapRandRecall}{24}
\newcommand{\poFeverTFDisjointRecall}{22}
\newcommand{\poFeverTFDisjointRandRecall}{24}

\newcommand{\poOverlapShare}{35}

\newcommand{\poOaiTFOverlapDelta}{+4.3}
\newcommand{\poOaiTFOverlapLo}{+2.2}
\newcommand{\poOaiTFOverlapHi}{+6.4}

\newcommand{\poOaiTFDisjointDelta}{\ensuremath{-}2.9}
\newcommand{\poOaiTFDisjointLo}{\ensuremath{-}4.3}
\newcommand{\poOaiTFDisjointHi}{\ensuremath{-}1.4}

\newcommand{\poOaiFOverlapDelta}{+2.6}
\newcommand{\poOaiFOverlapLo}{+0.5}
\newcommand{\poOaiFOverlapHi}{+4.7}

\newcommand{\poOaiFDisjointDelta}{\ensuremath{-}1.6}
\newcommand{\poOaiFDisjointLo}{\ensuremath{-}3.3}
\newcommand{\poOaiFDisjointHi}{+0.1}

\newcommand{\poDecTFOverlapDelta}{+5.3}
\newcommand{\poDecTFOverlapLo}{+2.9}
\newcommand{\poDecTFOverlapHi}{+7.7}

\newcommand{\poDecTFDisjointDelta}{\ensuremath{-}3.0}
\newcommand{\poDecTFDisjointLo}{\ensuremath{-}4.5}
\newcommand{\poDecTFDisjointHi}{\ensuremath{-}1.4}

\newcommand{\poDecFOverlapDelta}{+4.2}
\newcommand{\poDecFOverlapLo}{+1.9}
\newcommand{\poDecFOverlapHi}{+6.3}

\newcommand{\poDecFDisjointDelta}{\ensuremath{-}3.4}
\newcommand{\poDecFDisjointLo}{\ensuremath{-}5.0}
\newcommand{\poDecFDisjointHi}{\ensuremath{-}1.8}

\newcommand{\poFeverTFOverlapDelta}{+25.8}
\newcommand{\poFeverTFOverlapLo}{+21.3}
\newcommand{\poFeverTFOverlapHi}{+30.3}

\newcommand{\poFeverFOverlapDelta}{+15.3}
\newcommand{\poFeverFOverlapLo}{+11.0}
\newcommand{\poFeverFOverlapHi}{+19.6}

%% file: generated/compress_third.tex
\newcommand{\ctNDocs}{792}
\newcommand{\ctNClaims}{3,707}
\newcommand{\ctNHidden}{2,915}
\newcommand{\ctNOverlap}{759}
\newcommand{\ctNDisjoint}{2,156}
\newcommand{\ctOverlapShare}{26}
\newcommand{\ctNRequests}{4,752}
\newcommand{\ctInputTokens}{34,695,247}
\newcommand{\ctCostUSD}{3.47}
\newcommand{\ctNAnswers}{37,070}
\newcommand{\ctNRefused}{1}
\newcommand{\ctTrueShare}{55}
\newcommand{\ctMeanSent}{112}
\newcommand{\ctMeanClaims}{4.7}
\newcommand{\ctTFOverlapDelta}{+21.4}
\newcommand{\ctTFOverlapLo}{+17.3}
\newcommand{\ctTFOverlapHi}{+25.5}

\newcommand{\ctHOneTFVerdict}{supported}
\newcommand{\ctTFDisjointDelta}{+9.8}
\newcommand{\ctTFDisjointLo}{+7.4}
\newcommand{\ctTFDisjointHi}{+12.1}

\newcommand{\ctTFDisjointPPos}{<0.0005}

\newcommand{\ctHTwoTFVerdict}{contradicted}
\newcommand{\ctTFAllDelta}{+12.5}
\newcommand{\ctTFAllLo}{+10.4}
\newcommand{\ctTFAllHi}{+14.5}

\newcommand{\ctFOverlapDelta}{+7.0}
\newcommand{\ctFOverlapLo}{+4.2}
\newcommand{\ctFOverlapHi}{+9.9}

\newcommand{\ctFDisjointDelta}{+3.9}
\newcommand{\ctFDisjointLo}{+2.0}
\newcommand{\ctFDisjointHi}{+5.9}

\newcommand{\ctFAllDelta}{+3.9}
\newcommand{\ctFAllLo}{+2.4}
\newcommand{\ctFAllHi}{+5.6}

\newcommand{\ctOrigHiddenOverlap}{91}
\newcommand{\ctOrigHiddenDisjoint}{89}

\newcommand{\ctOrigKnown}{90}
\newcommand{\ctKnownTFHiddenOverlap}{92}
\newcommand{\ctKnownTFHiddenDisjoint}{80}

\newcommand{\ctKnownTFKnown}{91}
\newcommand{\ctKnownTFRecallOverlap}{89}
\newcommand{\ctKnownTFRecallDisjoint}{46}
\newcommand{\ctRandomTFHiddenOverlap}{71}
\newcommand{\ctRandomTFHiddenDisjoint}{70}

\newcommand{\ctRandomTFKnown}{70}
\newcommand{\ctRandomTFRecallOverlap}{23}
\newcommand{\ctRandomTFRecallDisjoint}{25}

\newcommand{\ctKnownFRecallOverlap}{94}
\newcommand{\ctKnownFRecallDisjoint}{67}

\newcommand{\ctRandomFRecallOverlap}{50}
\newcommand{\ctRandomFRecallDisjoint}{47}

%% file: generated/p4_retention.tex
\newcommand{\prtNRequests}{16,102}
\newcommand{\prtInputTokens}{15,135,563}
\newcommand{\prtCostUSD}{1.51}
\newcommand{\prtFeverNHidden}{5,433}
\newcommand{\prtFeverNEligible}{5,132}
\newcommand{\prtFeverNPairsAnalyzed}{5,132}
\newcommand{\prtFeverNDocs}{200}

\newcommand{\prtFeverNDisjoint}{2,551}
\newcommand{\prtFeverNOverlap}{2,581}
\newcommand{\prtFeverEvInKnown}{49}
\newcommand{\prtFeverAccKeepAll}{95}
\newcommand{\prtFeverAccKeepDisjoint}{95}
\newcommand{\prtFeverAccKeepOverlap}{95}
\newcommand{\prtFeverAccDropAll}{45}
\newcommand{\prtFeverAccDropDisjoint}{53}
\newcommand{\prtFeverAccDropOverlap}{45}
\newcommand{\prtFeverAccKnownAll}{76}
\newcommand{\prtFeverAccKnownDisjoint}{62}
\newcommand{\prtFeverAccKnownOverlap}{88}
\newcommand{\prtFeverAccRandomAll}{59}
\newcommand{\prtFeverAccRandomDisjoint}{62}
\newcommand{\prtFeverAccRandomOverlap}{61}
\newcommand{\prtFeverAccOriginalAll}{95}
\newcommand{\prtFeverAccOriginalDisjoint}{93}
\newcommand{\prtFeverAccOriginalOverlap}{95}
\newcommand{\prtEiNHidden}{2,915}

\newcommand{\prtEiNPairsAnalyzed}{2,915}
\newcommand{\prtEiNDocs}{792}

\newcommand{\prtEiNDisjoint}{2,156}
\newcommand{\prtEiNOverlap}{759}
\newcommand{\prtEiEvInKnown}{57}
\newcommand{\prtEiAccKeepAll}{90}
\newcommand{\prtEiAccKeepDisjoint}{90}
\newcommand{\prtEiAccKeepOverlap}{93}
\newcommand{\prtEiAccDropAll}{72}
\newcommand{\prtEiAccDropDisjoint}{71}
\newcommand{\prtEiAccDropOverlap}{79}
\newcommand{\prtEiAccKnownAll}{82}
\newcommand{\prtEiAccKnownDisjoint}{80}
\newcommand{\prtEiAccKnownOverlap}{92}
\newcommand{\prtEiAccRandomAll}{70}
\newcommand{\prtEiAccRandomDisjoint}{70}
\newcommand{\prtEiAccRandomOverlap}{71}
\newcommand{\prtEiAccOriginalAll}{90}
\newcommand{\prtEiAccOriginalDisjoint}{89}
\newcommand{\prtEiAccOriginalOverlap}{91}
\newcommand{\prtFeverKeepDropAllDelta}{+50.0}
\newcommand{\prtFeverKeepDropAllLo}{+47.4}
\newcommand{\prtFeverKeepDropAllHi}{+52.4}
\newcommand{\prtFeverKeepDropAllNDocs}{200}

\newcommand{\prtFeverKeepDropDisjointDelta}{+41.1}
\newcommand{\prtFeverKeepDropDisjointLo}{+36.2}
\newcommand{\prtFeverKeepDropDisjointHi}{+45.8}
\newcommand{\prtFeverKeepDropDisjointNDocs}{158}

\newcommand{\prtFeverKeepDropOverlapDelta}{+50.1}
\newcommand{\prtFeverKeepDropOverlapLo}{+46.7}
\newcommand{\prtFeverKeepDropOverlapHi}{+53.5}
\newcommand{\prtFeverKeepDropOverlapNDocs}{196}

\newcommand{\prtEiKeepDropAllDelta}{+17.8}
\newcommand{\prtEiKeepDropAllLo}{+15.9}
\newcommand{\prtEiKeepDropAllHi}{+19.7}
\newcommand{\prtEiKeepDropAllNDocs}{792}

\newcommand{\prtEiKeepDropDisjointDelta}{+18.9}
\newcommand{\prtEiKeepDropDisjointLo}{+16.8}
\newcommand{\prtEiKeepDropDisjointHi}{+21.2}
\newcommand{\prtEiKeepDropDisjointNDocs}{729}

\newcommand{\prtEiKeepDropOverlapDelta}{+13.4}
\newcommand{\prtEiKeepDropOverlapLo}{+10.1}
\newcommand{\prtEiKeepDropOverlapHi}{+16.5}
\newcommand{\prtEiKeepDropOverlapNDocs}{406}

%% file: generated/p4_retention_labels.tex
\newcommand{\prtLabFeverNConflicting}{4}
\newcommand{\prtLabFeverTrueN}{3,658}
\newcommand{\prtLabFeverTrueNDocs}{199}
\newcommand{\prtLabFeverTrueAccKeep}{94}
\newcommand{\prtLabFeverTrueAccDrop}{21}
\newcommand{\prtLabFeverTrueAccKnown}{68}
\newcommand{\prtLabFeverTrueAccRandom}{41}
\newcommand{\prtLabFeverTrueAccOriginal}{94}
\newcommand{\prtLabFeverFalseN}{1,474}
\newcommand{\prtLabFeverFalseNDocs}{199}
\newcommand{\prtLabFeverFalseAccKeep}{97}
\newcommand{\prtLabFeverFalseAccDrop}{98}
\newcommand{\prtLabFeverFalseAccKnown}{98}
\newcommand{\prtLabFeverFalseAccRandom}{99}
\newcommand{\prtLabFeverFalseAccOriginal}{97}
\newcommand{\prtLabEiTrueN}{1,535}
\newcommand{\prtLabEiTrueNDocs}{636}
\newcommand{\prtLabEiTrueAccKeep}{82}
\newcommand{\prtLabEiTrueAccDrop}{50}
\newcommand{\prtLabEiTrueAccKnown}{67}
\newcommand{\prtLabEiTrueAccRandom}{44}
\newcommand{\prtLabEiTrueAccOriginal}{81}
\newcommand{\prtLabEiFalseN}{1,380}
\newcommand{\prtLabEiFalseNDocs}{590}
\newcommand{\prtLabEiFalseAccKeep}{98}
\newcommand{\prtLabEiFalseAccDrop}{98}
\newcommand{\prtLabEiFalseAccKnown}{98}
\newcommand{\prtLabEiFalseAccRandom}{98}
\newcommand{\prtLabEiFalseAccOriginal}{98}
\newcommand{\prtLabNRequestsAnalyzed}{16,094}

%% file: generated/locomo.tex
\newcommand{\lcmAccAgnosticTFKD}{90.4}

\newcommand{\lcmAccAgnosticFKD}{94.8}

\newcommand{\lcmAccFullKD}{96.8}

\newcommand{\lcmAccLeadTFKD}{82.9}

\newcommand{\lcmAccLeadFKD}{87.2}

\newcommand{\lcmAccNoneKD}{81.4}

\newcommand{\lcmAccRandomTFKD}{88.3}

\newcommand{\lcmAccRandomFKD}{89.7}

\newcommand{\lcmAccRecencyTFKD}{83.1}

\newcommand{\lcmAccRecencyFKD}{87.6}

\newcommand{\lcmAccTfidfTFKD}{88.7}

\newcommand{\lcmAccTfidfFKD}{91.4}

\newcommand{\lcmAccTypedTFKD}{87.9}

\newcommand{\lcmAccTypedFKD}{91.9}

\newcommand{\lcmAccQwenFullKD}{96.6}

\newcommand{\lcmAccQwenNoneKD}{85.1}

\newcommand{\lcmAccQwenRandomTFKD}{83.2}

\newcommand{\lcmAccQwenRecencyTFKD}{82.4}

\newcommand{\lcmAccQwenTfidfTFKD}{85.7}

\newcommand{\lcmAccQwenTypedTFKD}{83.8}
\newcommand{\lcmHOneVerdict}{unresolved}
\newcommand{\lcmHOneDelta}{\ensuremath{-}0.4}
\newcommand{\lcmHOneLo}{\ensuremath{-}2.1}
\newcommand{\lcmHOneHi}{+1.0}

\newcommand{\lcmHTwoAVerdict}{supported}
\newcommand{\lcmHTwoADelta}{+13.3}
\newcommand{\lcmHTwoALo}{+10.9}
\newcommand{\lcmHTwoAHi}{+16.4}

\newcommand{\lcmHTwoANConvPos}{10}

\newcommand{\lcmHTwoBVerdict}{supported}
\newcommand{\lcmHTwoBDelta}{+13.3}
\newcommand{\lcmHTwoBLo}{+10.7}
\newcommand{\lcmHTwoBHi}{+16.8}

\newcommand{\lcmHTwoBNQ}{1,408}

\newcommand{\lcmHThreeVerdict}{not assessable}
\newcommand{\lcmHThreeDelta}{+2.1}
\newcommand{\lcmHThreeLo}{\ensuremath{-}2.5}
\newcommand{\lcmHThreeHi}{+6.5}

\newcommand{\lcmHFourVerdict}{supported}
\newcommand{\lcmHFourDelta}{0.81}
\newcommand{\lcmHFourLo}{0.79}
\newcommand{\lcmHFourHi}{0.84}

\newcommand{\lcmHOneQwenDelta}{+0.6}
\newcommand{\lcmHOneQwenLo}{\ensuremath{-}1.8}
\newcommand{\lcmHOneQwenHi}{+2.8}

\newcommand{\lcmTypedTFRetained}{27.2}

\newcommand{\lcmAgnosticTFRetained}{48.6}

\newcommand{\lcmTfidfTFRetained}{25.7}

\newcommand{\lcmRandomTFRetained}{25.4}

\newcommand{\lcmLeadTFRetained}{27.8}

\newcommand{\lcmRecencyTFRetained}{25.3}

\newcommand{\lcmKDAllKeep}{97.7}
\newcommand{\lcmKDAllDrop}{85.1}

\newcommand{\lcmLunaTokens}{50,816,715}
\newcommand{\lcmLunaUSD}{5.08}
\newcommand{\lcmQwenTokens}{80,631,483}
\newcommand{\lcmQwenUSD}{12.10}

%% file: generated/p4_audit.tex
\newcommand{\auRdOaiAcc}{88.2}

\newcommand{\auRdDecAcc}{75.3}

\newcommand{\auRdNo}{55.6}

%% file: generated/p4new.tex
\newcommand{\rmMrcOaiKnownTFKnownDelta}{+9.7}
\newcommand{\rmMrcOaiKnownTFKnownLo}{+7.2}
\newcommand{\rmMrcOaiKnownTFKnownHi}{+11.9}

\newcommand{\rmMrcOaiKnownTFHiddenDelta}{\ensuremath{-}0.2}
\newcommand{\rmMrcOaiKnownTFHiddenLo}{\ensuremath{-}1.1}
\newcommand{\rmMrcOaiKnownTFHiddenHi}{+0.7}

\newcommand{\rmMrcOaiKnownTFOverlapDelta}{+4.4}
\newcommand{\rmMrcOaiKnownTFOverlapLo}{+2.6}
\newcommand{\rmMrcOaiKnownTFOverlapHi}{+6.1}

\newcommand{\rmMrcOaiKnownTFDisjointDelta}{\ensuremath{-}2.2}
\newcommand{\rmMrcOaiKnownTFDisjointLo}{\ensuremath{-}3.4}
\newcommand{\rmMrcOaiKnownTFDisjointHi}{\ensuremath{-}1.0}

\newcommand{\rmMrcOaiAgnosticTFHiddenDelta}{\ensuremath{-}0.3}
\newcommand{\rmMrcOaiAgnosticTFHiddenLo}{\ensuremath{-}1.1}
\newcommand{\rmMrcOaiAgnosticTFHiddenHi}{+0.7}

\newcommand{\rmMrcOaiKnownFHiddenDelta}{\ensuremath{-}0.2}
\newcommand{\rmMrcOaiKnownFHiddenLo}{\ensuremath{-}1.3}
\newcommand{\rmMrcOaiKnownFHiddenHi}{+0.8}

\newcommand{\rmMrcOaiKnownFOverlapDelta}{+2.8}
\newcommand{\rmMrcOaiKnownFOverlapLo}{+1.3}
\newcommand{\rmMrcOaiKnownFOverlapHi}{+4.4}

\newcommand{\rmMrcOaiKnownFDisjointDelta}{\ensuremath{-}1.5}
\newcommand{\rmMrcOaiKnownFDisjointLo}{\ensuremath{-}2.8}
\newcommand{\rmMrcOaiKnownFDisjointHi}{\ensuremath{-}0.3}

\newcommand{\rmMrcOaiAgnosticFHiddenDelta}{\ensuremath{-}0.1}
\newcommand{\rmMrcOaiAgnosticFHiddenLo}{\ensuremath{-}1.1}
\newcommand{\rmMrcOaiAgnosticFHiddenHi}{+0.8}

\newcommand{\rmMrcDecKnownTFKnownDelta}{+8.6}
\newcommand{\rmMrcDecKnownTFKnownLo}{+6.0}
\newcommand{\rmMrcDecKnownTFKnownHi}{+11.4}

\newcommand{\rmMrcDecKnownTFHiddenDelta}{\ensuremath{-}0.1}
\newcommand{\rmMrcDecKnownTFHiddenLo}{\ensuremath{-}1.1}
\newcommand{\rmMrcDecKnownTFHiddenHi}{+0.8}

\newcommand{\rmMrcDecKnownTFOverlapDelta}{+6.0}
\newcommand{\rmMrcDecKnownTFOverlapLo}{+4.3}
\newcommand{\rmMrcDecKnownTFOverlapHi}{+7.7}

\newcommand{\rmMrcDecKnownTFDisjointDelta}{\ensuremath{-}2.9}
\newcommand{\rmMrcDecKnownTFDisjointLo}{\ensuremath{-}4.0}
\newcommand{\rmMrcDecKnownTFDisjointHi}{\ensuremath{-}1.9}

\newcommand{\rmMrcDecAgnosticTFHiddenDelta}{+0.8}
\newcommand{\rmMrcDecAgnosticTFHiddenLo}{\ensuremath{-}0.1}
\newcommand{\rmMrcDecAgnosticTFHiddenHi}{+1.7}

\newcommand{\rmMrcDecKnownFHiddenDelta}{\ensuremath{-}0.6}
\newcommand{\rmMrcDecKnownFHiddenLo}{\ensuremath{-}1.6}
\newcommand{\rmMrcDecKnownFHiddenHi}{+0.3}

\newcommand{\rmMrcDecKnownFOverlapDelta}{+3.9}
\newcommand{\rmMrcDecKnownFOverlapLo}{+2.4}
\newcommand{\rmMrcDecKnownFOverlapHi}{+5.4}

\newcommand{\rmMrcDecKnownFDisjointDelta}{\ensuremath{-}3.2}
\newcommand{\rmMrcDecKnownFDisjointLo}{\ensuremath{-}4.4}
\newcommand{\rmMrcDecKnownFDisjointHi}{\ensuremath{-}1.9}

\newcommand{\rmMrcDecAgnosticFHiddenDelta}{+0.4}
\newcommand{\rmMrcDecAgnosticFHiddenLo}{\ensuremath{-}0.5}
\newcommand{\rmMrcDecAgnosticFHiddenHi}{+1.3}

\newcommand{\rmFeverKnownTFHiddenDelta}{+17.1}
\newcommand{\rmFeverKnownTFHiddenLo}{+13.7}
\newcommand{\rmFeverKnownTFHiddenHi}{+20.2}

\newcommand{\rmFeverKnownTFOverlapDelta}{+26.4}
\newcommand{\rmFeverKnownTFOverlapLo}{+22.8}
\newcommand{\rmFeverKnownTFOverlapHi}{+29.6}

\newcommand{\rmFeverKnownTFDisjointDelta}{\ensuremath{-}0.6}
\newcommand{\rmFeverKnownTFDisjointLo}{\ensuremath{-}4.4}
\newcommand{\rmFeverKnownTFDisjointHi}{+2.9}

\newcommand{\rmFeverAgnosticTFHiddenDelta}{+2.0}
\newcommand{\rmFeverAgnosticTFHiddenLo}{\ensuremath{-}0.9}
\newcommand{\rmFeverAgnosticTFHiddenHi}{+5.0}

\newcommand{\rmFeverKnownFHiddenDelta}{+9.1}
\newcommand{\rmFeverKnownFHiddenLo}{+6.6}
\newcommand{\rmFeverKnownFHiddenHi}{+11.5}

\newcommand{\rmFeverKnownFOverlapDelta}{+14.9}
\newcommand{\rmFeverKnownFOverlapLo}{+12.3}
\newcommand{\rmFeverKnownFOverlapHi}{+17.7}

\newcommand{\rmFeverKnownFDisjointDelta}{\ensuremath{-}3.3}
\newcommand{\rmFeverKnownFDisjointLo}{\ensuremath{-}6.6}
\newcommand{\rmFeverKnownFDisjointHi}{\ensuremath{-}0.1}

\newcommand{\rmFeverAgnosticFHiddenDelta}{+2.5}
\newcommand{\rmFeverAgnosticFHiddenLo}{\ensuremath{-}0.3}
\newcommand{\rmFeverAgnosticFHiddenHi}{+5.1}

\newcommand{\rmEiKnownTFHiddenDelta}{+12.4}
\newcommand{\rmEiKnownTFHiddenLo}{+11.0}
\newcommand{\rmEiKnownTFHiddenHi}{+13.9}

\newcommand{\rmEiKnownTFOverlapDelta}{+21.9}
\newcommand{\rmEiKnownTFOverlapLo}{+19.2}
\newcommand{\rmEiKnownTFOverlapHi}{+24.6}

\newcommand{\rmEiKnownTFDisjointDelta}{+9.6}
\newcommand{\rmEiKnownTFDisjointLo}{+8.0}
\newcommand{\rmEiKnownTFDisjointHi}{+11.4}

\newcommand{\rmEiKnownFHiddenDelta}{+4.4}
\newcommand{\rmEiKnownFHiddenLo}{+3.2}
\newcommand{\rmEiKnownFHiddenHi}{+5.6}

\newcommand{\rmEiKnownFOverlapDelta}{+8.0}
\newcommand{\rmEiKnownFOverlapLo}{+5.9}
\newcommand{\rmEiKnownFOverlapHi}{+9.9}

\newcommand{\rmEiKnownFDisjointDelta}{+3.6}
\newcommand{\rmEiKnownFDisjointLo}{+2.2}
\newcommand{\rmEiKnownFDisjointHi}{+5.0}

\newcommand{\rmMrcOaiMasks}{6}
\newcommand{\rmMrcDecMasks}{11}

\newcommand{\rmEiMasks}{4--7}

\newcommand{\rmMrcNTests}{12}
\newcommand{\rmMrcMaxBound}{1.6}

\newcommand{\rmMrcOaiKnownSpreadTFMin}{\ensuremath{-}0.8}
\newcommand{\rmMrcOaiKnownSpreadTFMax}{+0.3}

\newcommand{\rmDriftOaiSameSide}{100.0}
\newcommand{\rmDriftDecSameSide}{99.2}
\newcommand{\llMrcOaiLinguaRandomTFKnownDelta}{+12.8}
\newcommand{\llMrcOaiLinguaRandomTFKnownLo}{+10.3}
\newcommand{\llMrcOaiLinguaRandomTFKnownHi}{+15.1}

\newcommand{\llMrcOaiKnownLinguaTFKnownDelta}{\ensuremath{-}3.0}
\newcommand{\llMrcOaiKnownLinguaTFKnownLo}{\ensuremath{-}5.9}
\newcommand{\llMrcOaiKnownLinguaTFKnownHi}{\ensuremath{-}0.1}
\newcommand{\llMrcOaiLinguaRandomTFHiddenDelta}{+12.9}
\newcommand{\llMrcOaiLinguaRandomTFHiddenLo}{+11.8}
\newcommand{\llMrcOaiLinguaRandomTFHiddenHi}{+14.1}
\newcommand{\llMrcOaiAccTFHidden}{80.3}
\newcommand{\llMrcOaiKnownLinguaTFHiddenDelta}{\ensuremath{-}13.1}
\newcommand{\llMrcOaiKnownLinguaTFHiddenLo}{\ensuremath{-}14.5}
\newcommand{\llMrcOaiKnownLinguaTFHiddenHi}{\ensuremath{-}11.8}
\newcommand{\llMrcOaiLinguaRandomFKnownDelta}{+9.9}
\newcommand{\llMrcOaiLinguaRandomFKnownLo}{+8.0}
\newcommand{\llMrcOaiLinguaRandomFKnownHi}{+12.0}

\newcommand{\llMrcOaiKnownLinguaFKnownDelta}{\ensuremath{-}2.3}
\newcommand{\llMrcOaiKnownLinguaFKnownLo}{\ensuremath{-}4.3}
\newcommand{\llMrcOaiKnownLinguaFKnownHi}{\ensuremath{-}0.2}
\newcommand{\llMrcOaiLinguaRandomFHiddenDelta}{+10.2}
\newcommand{\llMrcOaiLinguaRandomFHiddenLo}{+9.3}
\newcommand{\llMrcOaiLinguaRandomFHiddenHi}{+11.1}

\newcommand{\llMrcOaiKnownLinguaFHiddenDelta}{\ensuremath{-}10.4}
\newcommand{\llMrcOaiKnownLinguaFHiddenLo}{\ensuremath{-}11.7}
\newcommand{\llMrcOaiKnownLinguaFHiddenHi}{\ensuremath{-}9.2}
\newcommand{\llMrcDecLinguaRandomTFKnownDelta}{+6.9}
\newcommand{\llMrcDecLinguaRandomTFKnownLo}{+4.6}
\newcommand{\llMrcDecLinguaRandomTFKnownHi}{+9.3}

\newcommand{\llMrcDecKnownLinguaTFKnownDelta}{+1.7}
\newcommand{\llMrcDecKnownLinguaTFKnownLo}{\ensuremath{-}1.3}
\newcommand{\llMrcDecKnownLinguaTFKnownHi}{+4.8}
\newcommand{\llMrcDecLinguaRandomTFHiddenDelta}{+6.5}
\newcommand{\llMrcDecLinguaRandomTFHiddenLo}{+5.4}
\newcommand{\llMrcDecLinguaRandomTFHiddenHi}{+7.6}

\newcommand{\llMrcDecKnownLinguaTFHiddenDelta}{\ensuremath{-}6.6}
\newcommand{\llMrcDecKnownLinguaTFHiddenLo}{\ensuremath{-}8.0}
\newcommand{\llMrcDecKnownLinguaTFHiddenHi}{\ensuremath{-}5.2}
\newcommand{\llMrcDecLinguaRandomFKnownDelta}{+3.4}
\newcommand{\llMrcDecLinguaRandomFKnownLo}{+1.3}
\newcommand{\llMrcDecLinguaRandomFKnownHi}{+5.5}

\newcommand{\llMrcDecKnownLinguaFKnownDelta}{+2.8}
\newcommand{\llMrcDecKnownLinguaFKnownLo}{+0.3}
\newcommand{\llMrcDecKnownLinguaFKnownHi}{+5.2}
\newcommand{\llMrcDecLinguaRandomFHiddenDelta}{+5.3}
\newcommand{\llMrcDecLinguaRandomFHiddenLo}{+4.4}
\newcommand{\llMrcDecLinguaRandomFHiddenHi}{+6.3}

\newcommand{\llMrcDecKnownLinguaFHiddenDelta}{\ensuremath{-}6.0}
\newcommand{\llMrcDecKnownLinguaFHiddenLo}{\ensuremath{-}7.3}
\newcommand{\llMrcDecKnownLinguaFHiddenHi}{\ensuremath{-}4.6}

\newcommand{\llFeverLinguaRandomTFKnownDelta}{+22.0}
\newcommand{\llFeverLinguaRandomTFKnownLo}{+16.2}
\newcommand{\llFeverLinguaRandomTFKnownHi}{+27.9}

\newcommand{\llFeverKnownLinguaTFKnownDelta}{+12.5}
\newcommand{\llFeverKnownLinguaTFKnownLo}{+7.0}
\newcommand{\llFeverKnownLinguaTFKnownHi}{+18.5}
\newcommand{\llFeverLinguaRandomTFHiddenDelta}{+20.8}
\newcommand{\llFeverLinguaRandomTFHiddenLo}{+18.8}
\newcommand{\llFeverLinguaRandomTFHiddenHi}{+23.0}
\newcommand{\llFeverAccTFHidden}{81.3}
\newcommand{\llFeverKnownLinguaTFHiddenDelta}{\ensuremath{-}3.7}
\newcommand{\llFeverKnownLinguaTFHiddenLo}{\ensuremath{-}6.7}
\newcommand{\llFeverKnownLinguaTFHiddenHi}{\ensuremath{-}0.6}
\newcommand{\llFeverLinguaRandomFKnownDelta}{+17.8}
\newcommand{\llFeverLinguaRandomFKnownLo}{+13.3}
\newcommand{\llFeverLinguaRandomFKnownHi}{+22.7}

\newcommand{\llFeverKnownLinguaFKnownDelta}{+5.5}
\newcommand{\llFeverKnownLinguaFKnownLo}{+2.0}
\newcommand{\llFeverKnownLinguaFKnownHi}{+9.0}
\newcommand{\llFeverLinguaRandomFHiddenDelta}{+15.9}
\newcommand{\llFeverLinguaRandomFHiddenLo}{+14.3}
\newcommand{\llFeverLinguaRandomFHiddenHi}{+17.6}

\newcommand{\llFeverKnownLinguaFHiddenDelta}{\ensuremath{-}6.9}
\newcommand{\llFeverKnownLinguaFHiddenLo}{\ensuremath{-}9.2}
\newcommand{\llFeverKnownLinguaFHiddenHi}{\ensuremath{-}4.7}
\newcommand{\llEiLinguaRandomTFKnownDelta}{+2.4}
\newcommand{\llEiLinguaRandomTFKnownLo}{\ensuremath{-}0.4}
\newcommand{\llEiLinguaRandomTFKnownHi}{+5.1}

\newcommand{\llEiKnownLinguaTFKnownDelta}{+18.8}
\newcommand{\llEiKnownLinguaTFKnownLo}{+15.8}
\newcommand{\llEiKnownLinguaTFKnownHi}{+22.1}
\newcommand{\llEiLinguaRandomTFHiddenDelta}{+0.2}
\newcommand{\llEiLinguaRandomTFHiddenLo}{\ensuremath{-}1.5}
\newcommand{\llEiLinguaRandomTFHiddenHi}{+1.9}
\newcommand{\llEiAccTFHidden}{69.8}
\newcommand{\llEiKnownLinguaTFHiddenDelta}{+12.2}
\newcommand{\llEiKnownLinguaTFHiddenLo}{+10.1}
\newcommand{\llEiKnownLinguaTFHiddenHi}{+14.4}
\newcommand{\llEiLinguaRandomFKnownDelta}{+5.3}
\newcommand{\llEiLinguaRandomFKnownLo}{+3.3}
\newcommand{\llEiLinguaRandomFKnownHi}{+7.3}

\newcommand{\llEiKnownLinguaFKnownDelta}{+4.3}
\newcommand{\llEiKnownLinguaFKnownLo}{+2.4}
\newcommand{\llEiKnownLinguaFKnownHi}{+6.3}
\newcommand{\llEiLinguaRandomFHiddenDelta}{+3.4}
\newcommand{\llEiLinguaRandomFHiddenLo}{+2.0}
\newcommand{\llEiLinguaRandomFHiddenHi}{+4.8}

\newcommand{\llEiKnownLinguaFHiddenDelta}{+1.0}
\newcommand{\llEiKnownLinguaFHiddenLo}{\ensuremath{-}0.5}
\newcommand{\llEiKnownLinguaFHiddenHi}{+2.6}
\newcommand{\llEiTFLinguaSentCov}{93.9}
\newcommand{\llEiTFLinguaKeptTokens}{26.1}
\newcommand{\llEiTFLinguaKeptChars}{26.4}

\newcommand{\llEiTFRandomKeptChars}{25.2}
\newcommand{\llEiTFLinguaKeptWords}{20.8}

\newcommand{\llEiTFLinguaKnownEv}{20.3}
\newcommand{\llEiTFKnownKnownEv}{95.6}

\newcommand{\llEiTFLinguaHiddenEv}{21.4}
\newcommand{\llEiTFKnownHiddenEv}{57.9}
\newcommand{\llEiTFRandomHiddenEv}{24.8}

\newcommand{\llFeverTFLinguaSentCov}{96.4}
\newcommand{\llFeverTFLinguaKeptTokens}{24.7}
\newcommand{\llFeverTFLinguaKeptChars}{25.9}

\newcommand{\llFeverTFRandomKeptChars}{24.5}
\newcommand{\llFeverTFLinguaKeptWords}{20.2}

\newcommand{\llFeverTFLinguaKnownEv}{23.7}
\newcommand{\llFeverTFKnownKnownEv}{80.3}

\newcommand{\llFeverTFLinguaHiddenEv}{23.2}
\newcommand{\llFeverTFKnownHiddenEv}{57.9}
\newcommand{\llFeverTFRandomHiddenEv}{23.8}

\newcommand{\llMrcTFLinguaSentCov}{92.2}
\newcommand{\llMrcTFLinguaKeptTokens}{23.5}
\newcommand{\llMrcTFLinguaKeptChars}{27.9}

\newcommand{\llMrcTFRandomKeptChars}{24.8}
\newcommand{\llMrcTFLinguaKeptWords}{20.1}

\newcommand{\llMrcTFLinguaKnownEv}{21.6}
\newcommand{\llMrcTFKnownKnownEv}{49.4}

\newcommand{\llMrcTFLinguaHiddenEv}{21.8}
\newcommand{\llMrcTFKnownHiddenEv}{23.5}
\newcommand{\llMrcTFRandomHiddenEv}{23.8}

\newcommand{\kdQwenAllDelta}{+10.7}
\newcommand{\kdQwenAllLo}{+8.3}
\newcommand{\kdQwenAllHi}{+13.3}
\newcommand{\kdQwenAllLoNinety}{+8.6}
\newcommand{\kdQwenAllKeep}{90.8}
\newcommand{\kdQwenAllDrop}{80.5}
\newcommand{\kdQwenAllNQ}{1,496}
\newcommand{\kdQwenAllNConvPos}{10}
\newcommand{\kdQwenDisjointDelta}{+10.6}
\newcommand{\kdQwenDisjointLo}{+8.0}
\newcommand{\kdQwenDisjointHi}{+13.3}
\newcommand{\kdQwenDisjointLoNinety}{+8.5}

\newcommand{\kdQwenDisjointNQ}{1,408}
\newcommand{\kdQwenDisjointNConvPos}{10}
\newcommand{\kdQwenOverlapDelta}{+9.1}
\newcommand{\kdQwenOverlapLo}{+4.1}
\newcommand{\kdQwenOverlapHi}{+15.2}

\newcommand{\kdQwenOverlapNQ}{88}
\newcommand{\kdQwenOverlapNConvPos}{6}

%% file: generated/p4fu.tex
\newcommand{\fuCEiKnownTFHiddenNPartial}{510}

\newcommand{\fuCEiTypedVsRandomTFRecoveryDelta}{+23.1}
\newcommand{\fuCEiTypedVsRandomTFRecoveryHi}{+25.9}
\newcommand{\fuCEiTypedVsRandomTFRecoveryLo}{+20.2}

\newcommand{\fuCEiTypedVsRandomTFSupportWithinDocumentsDelta}{\ensuremath{-}0.3}
\newcommand{\fuCEiTypedVsRandomTFSupportWithinDocumentsHi}{+2.7}
\newcommand{\fuCEiTypedVsRandomTFSupportWithinDocumentsLo}{\ensuremath{-}3.1}

\newcommand{\fuCFeverKnownTFHiddenNPartial}{732}

\newcommand{\fuCFeverTypedVsRandomTFRecoveryDelta}{+25.2}
\newcommand{\fuCFeverTypedVsRandomTFRecoveryHi}{+30.8}
\newcommand{\fuCFeverTypedVsRandomTFRecoveryLo}{+19.9}

\newcommand{\fuCFeverTypedVsRandomTFSupportDelta}{\ensuremath{-}1.0}
\newcommand{\fuCFeverTypedVsRandomTFSupportHi}{+2.2}
\newcommand{\fuCFeverTypedVsRandomTFSupportLo}{\ensuremath{-}3.9}
\newcommand{\fuCFeverTypedVsRandomTFSupportNClaims}{4,528}

\newcommand{\fuCFeverTypedVsRandomTFSupportWithinDocumentsDelta}{\ensuremath{-}1.6}
\newcommand{\fuCFeverTypedVsRandomTFSupportWithinDocumentsHi}{+1.6}
\newcommand{\fuCFeverTypedVsRandomTFSupportWithinDocumentsLo}{\ensuremath{-}4.5}

\newcommand{\fuDLunaFullAcc}{90.9}

\newcommand{\fuDLunaKEEPVsDROPAllDelta}{+18.5}
\newcommand{\fuDLunaKEEPVsDROPAllHi}{+21.3}
\newcommand{\fuDLunaKEEPVsDROPAllLo}{+15.6}

\newcommand{\fuDLunaKEEPVsDROPDisjointDelta}{+18.7}
\newcommand{\fuDLunaKEEPVsDROPDisjointHi}{+21.5}
\newcommand{\fuDLunaKEEPVsDROPDisjointLo}{+15.8}

\newcommand{\fuDLunaNoneAcc}{60.2}

\newcommand{\fuDLunaTypedTFVsRandomTFAllDelta}{+1.3}
\newcommand{\fuDLunaTypedTFVsRandomTFAllHi}{+4.3}
\newcommand{\fuDLunaTypedTFVsRandomTFAllLo}{\ensuremath{-}1.7}

\newcommand{\fuDPooledHardAcc}{60.2}
\newcommand{\fuDPooledOrigAcc}{82.2}

\newcommand{\fuDQwenKEEPVsDROPAllDelta}{+14.7}
\newcommand{\fuDQwenKEEPVsDROPAllHi}{+18.8}
\newcommand{\fuDQwenKEEPVsDROPAllLo}{+10.7}

\newcommand{\fuDQwenKEEPVsDROPDisjointDelta}{+15.1}

\newcommand{\fuDQwenNoneAcc}{71.5}

\newcommand{\fuDQwenTypedTFVsRandomTFAllDelta}{+0.2}
\newcommand{\fuDQwenTypedTFVsRandomTFAllHi}{+3.2}
\newcommand{\fuDQwenTypedTFVsRandomTFAllLo}{\ensuremath{-}2.9}

\newcommand{\fuDShareOriginal}{2.3}
\newcommand{\fuDShareOther}{59.5}
\newcommand{\fuDShareSame}{17.0}
\newcommand{\fuDShareShifted}{21.1}

\newcommand{\fuFeverLinguaVsRandomTFHiddenDelta}{+21.0}

\newcommand{\fuFeverTypedVsLinguaTFKnownDelta}{+13.0}
\newcommand{\fuFeverTypedVsLinguaTFKnownHi}{+18.5}

\newcommand{\fuFeverTypedVsLinguaTFKnownLo}{+7.5}

\newcommand{\fuFeverTypedVsRandomFHiddenDelta}{+10.4}
\newcommand{\fuFeverTypedVsRandomFHiddenHi}{+13.0}

\newcommand{\fuFeverTypedVsRandomFHiddenLo}{+7.9}

\newcommand{\fuFeverTypedVsRandomFKnownDelta}{+20.8}

\newcommand{\fuFeverTypedVsRandomTFHiddenDelta}{+17.8}
\newcommand{\fuFeverTypedVsRandomTFHiddenHi}{+20.8}

\newcommand{\fuFeverTypedVsRandomTFHiddenLo}{+15.0}

\newcommand{\fuFeverTypedVsRandomTFKnownDelta}{+35.7}

\newcommand{\fuMrcDecLongVsRandomFKnownDelta}{+1.5}
\newcommand{\fuMrcDecLongVsRandomFKnownHi}{+4.0}

\newcommand{\fuMrcDecLongVsRandomFKnownLo}{\ensuremath{-}1.0}

\newcommand{\fuMrcDecTypedVsLinguaFKnownDelta}{+3.5}
\newcommand{\fuMrcDecTypedVsLinguaFKnownHi}{+6.0}

\newcommand{\fuMrcDecTypedVsLinguaFKnownLo}{+1.0}

\newcommand{\fuMrcDecTypedVsRandomFKnownDelta}{+6.9}

\newcommand{\fuMrcDecTypedVsRandomTFHiddenHiNinety}{\ensuremath{-}0.3}

\newcommand{\fuMrcDecTypedVsRandomTFHiddenLoNinety}{\ensuremath{-}2.0}
\newcommand{\fuMrcDecTypedVsRandomTFKnownDelta}{+7.9}

\newcommand{\fuMrcLongNEquiv}{1}

\newcommand{\fuMrcMmrNEquiv}{4}
\newcommand{\fuMrcMmrNTests}{4}

\newcommand{\fuMrcOaiLinguaVsRandomTFHiddenDelta}{+12.1}

\newcommand{\fuMrcOaiLongVsRandomTFHiddenDelta}{\ensuremath{-}1.6}
\newcommand{\fuMrcOaiLongVsRandomTFHiddenHi}{\ensuremath{-}0.6}

\newcommand{\fuMrcOaiLongVsRandomTFHiddenLo}{\ensuremath{-}2.6}

\newcommand{\fuMrcOaiTypedVsLinguaFKnownDelta}{\ensuremath{-}1.9}

\newcommand{\fuMrcOaiTypedVsLinguaFKnownLo}{\ensuremath{-}3.8}

\newcommand{\fuMrcOaiTypedVsLinguaTFKnownDelta}{\ensuremath{-}2.3}
\newcommand{\fuMrcOaiTypedVsLinguaTFKnownHi}{+0.6}

\newcommand{\fuMrcOaiTypedVsLinguaTFKnownLo}{\ensuremath{-}4.9}

\newcommand{\fuMrcOaiTypedVsRandomFKnownDelta}{+7.0}

\newcommand{\fuMrcOaiTypedVsRandomTFHiddenDelta}{\ensuremath{-}0.4}
\newcommand{\fuMrcOaiTypedVsRandomTFHiddenHi}{+0.6}
\newcommand{\fuMrcOaiTypedVsRandomTFHiddenHiNinety}{+0.4}
\newcommand{\fuMrcOaiTypedVsRandomTFHiddenLo}{\ensuremath{-}1.5}
\newcommand{\fuMrcOaiTypedVsRandomTFHiddenLoNinety}{\ensuremath{-}1.3}
\newcommand{\fuMrcOaiTypedVsRandomTFKnownDelta}{+11.4}
\newcommand{\fuMrcOaiTypedVsRandomTFKnownHi}{+13.8}

\newcommand{\fuMrcOaiTypedVsRandomTFKnownLo}{+9.1}

\newcommand{\fuMrcTypedNEquiv}{4}
\newcommand{\fuMrcTypedNTests}{4}

\newcommand{\fuTokMaxDev}{1}

%% file: generated/p4_refusal.tex
\newcommand{\rfMaxRate}{1.2}
\newcommand{\rfMaxRateTokens}{4.8}
\newcommand{\rfTypedTFKnownWrong}{77}

\newcommand{\rfRandomTFKnownWrong}{66}

\newcommand{\rfTokensLongVsRandomFHiddenDropDelta}{\ensuremath{-}0.9}
\newcommand{\rfTokensLongVsRandomFHiddenDropLo}{\ensuremath{-}1.9}
\newcommand{\rfTokensLongVsRandomFHiddenDropHi}{+0.2}

\newcommand{\rfSingleTypedVsRandomTFHiddenDelta}{\ensuremath{-}1.0}

\newcommand{\rfMasksTypedVsRandomTFHiddenDelta}{\ensuremath{-}0.4}

\newcommand{\rfTokensLongVsRandomFHiddenDelta}{\ensuremath{-}1.4}
\newcommand{\rfTokensLongVsRandomFHiddenLo}{\ensuremath{-}2.4}
\newcommand{\rfTokensLongVsRandomFHiddenHi}{\ensuremath{-}0.3}

\newcommand{\rfNContrasts}{32}

\newcommand{\rfMaxShift}{2.1}
\newcommand{\rfOtherNContrasts}{48}

\newcommand{\rfOtherMaxShift}{0.03}

%% file: generated/p4_granularity.tex
\newcommand{\grFeverGranRandomFDelta}{+6.1}
\newcommand{\grFeverGranRandomFHi}{+8.0}
\newcommand{\grFeverGranRandomFLo}{+4.3}

\newcommand{\grFeverGranRandomTFDelta}{+6.3}
\newcommand{\grFeverGranRandomTFHi}{+8.2}
\newcommand{\grFeverGranRandomTFLo}{+4.4}

\newcommand{\grFeverGranScorerFDelta}{+14.6}
\newcommand{\grFeverGranScorerFHi}{+17.4}
\newcommand{\grFeverGranScorerFLo}{+11.9}

\newcommand{\grFeverGranScorerTFDelta}{+19.1}
\newcommand{\grFeverGranScorerTFHi}{+22.1}
\newcommand{\grFeverGranScorerTFLo}{+15.9}

\newcommand{\grFeverLinguaSentFHidden}{76.8}

\newcommand{\grFeverLinguaSentTFEvWords}{35.7}
\newcommand{\grFeverLinguaSentTFHidden}{63.1}

\newcommand{\grFeverLinguaSentTFSentCov}{29.6}
\newcommand{\grFeverLinguaTFEvWords}{26.4}
\newcommand{\grFeverLinguaTFHidden}{82.2}

\newcommand{\grFeverLinguaTFSentCov}{95.2}

\newcommand{\grFeverRandSentTFEvWords}{25.7}

\newcommand{\grFeverRandSentTFSentCov}{29.5}

\newcommand{\grFeverRandWordFHidden}{81.3}

\newcommand{\grFeverRandWordTFEvWords}{25.2}
\newcommand{\grFeverRandWordTFHidden}{67.5}

\newcommand{\grFeverRandWordTFSentCov}{98.3}
\newcommand{\grFeverScorerSentFDelta}{+1.7}
\newcommand{\grFeverScorerSentFHi}{+4.5}
\newcommand{\grFeverScorerSentFLo}{\ensuremath{-}1.1}

\newcommand{\grFeverScorerSentTFDelta}{+1.9}
\newcommand{\grFeverScorerSentTFHi}{+5.4}
\newcommand{\grFeverScorerSentTFLo}{\ensuremath{-}1.3}

\newcommand{\grFeverScorerWordFDelta}{+10.2}
\newcommand{\grFeverScorerWordFHi}{+11.5}
\newcommand{\grFeverScorerWordFLo}{+8.9}

\newcommand{\grFeverScorerWordTFDelta}{+14.7}
\newcommand{\grFeverScorerWordTFHi}{+16.8}
\newcommand{\grFeverScorerWordTFLo}{+12.7}

\newcommand{\grMrcDecGranRandomFDelta}{+0.5}
\newcommand{\grMrcDecGranRandomFHi}{+1.3}
\newcommand{\grMrcDecGranRandomFLo}{\ensuremath{-}0.3}

\newcommand{\grMrcDecGranRandomTFDelta}{+0.4}
\newcommand{\grMrcDecGranRandomTFHi}{+1.3}
\newcommand{\grMrcDecGranRandomTFLo}{\ensuremath{-}0.4}

\newcommand{\grMrcDecGranScorerFDelta}{+4.3}
\newcommand{\grMrcDecGranScorerFHi}{+5.5}
\newcommand{\grMrcDecGranScorerFLo}{+3.0}

\newcommand{\grMrcDecGranScorerTFDelta}{+4.6}
\newcommand{\grMrcDecGranScorerTFHi}{+5.9}
\newcommand{\grMrcDecGranScorerTFLo}{+3.2}

\newcommand{\grMrcDecLinguaSentFHidden}{67.4}

\newcommand{\grMrcDecLinguaSentTFHidden}{63.4}

\newcommand{\grMrcDecRandWordFHidden}{67.4}

\newcommand{\grMrcDecRandWordTFHidden}{62.6}

\newcommand{\grMrcDecScorerSentFDelta}{+0.5}
\newcommand{\grMrcDecScorerSentFHi}{+1.5}
\newcommand{\grMrcDecScorerSentFLo}{\ensuremath{-}0.4}

\newcommand{\grMrcDecScorerSentTFDelta}{+1.3}
\newcommand{\grMrcDecScorerSentTFHi}{+2.3}
\newcommand{\grMrcDecScorerSentTFLo}{+0.2}

\newcommand{\grMrcDecScorerWordFDelta}{+4.4}
\newcommand{\grMrcDecScorerWordFHi}{+5.5}
\newcommand{\grMrcDecScorerWordFLo}{+3.3}

\newcommand{\grMrcDecScorerWordTFDelta}{+5.4}
\newcommand{\grMrcDecScorerWordTFHi}{+6.6}
\newcommand{\grMrcDecScorerWordTFLo}{+4.2}

\newcommand{\grMrcLinguaSentTFEvWords}{28.4}
\newcommand{\grMrcLinguaSentTFSentCov}{32.6}
\newcommand{\grMrcLinguaTFEvWords}{26.3}
\newcommand{\grMrcLinguaTFSentCov}{92.9}
\newcommand{\grMrcOaiGranRandomFDelta}{+2.4}
\newcommand{\grMrcOaiGranRandomFHi}{+3.3}
\newcommand{\grMrcOaiGranRandomFLo}{+1.6}

\newcommand{\grMrcOaiGranRandomTFDelta}{+1.2}
\newcommand{\grMrcOaiGranRandomTFHi}{+2.1}
\newcommand{\grMrcOaiGranRandomTFLo}{+0.3}

\newcommand{\grMrcOaiGranScorerFDelta}{+9.7}
\newcommand{\grMrcOaiGranScorerFHi}{+10.9}
\newcommand{\grMrcOaiGranScorerFLo}{+8.6}

\newcommand{\grMrcOaiGranScorerTFDelta}{+11.7}
\newcommand{\grMrcOaiGranScorerTFHi}{+13.0}
\newcommand{\grMrcOaiGranScorerTFLo}{+10.4}

\newcommand{\grMrcOaiLinguaSentFHidden}{76.1}

\newcommand{\grMrcOaiLinguaSentTFHidden}{68.8}

\newcommand{\grMrcOaiRandWordFHidden}{78.5}

\newcommand{\grMrcOaiRandWordTFHidden}{70.9}

\newcommand{\grMrcOaiRandWordTFRefused}{6.4}
\newcommand{\grMrcOaiScorerSentFDelta}{+0.1}
\newcommand{\grMrcOaiScorerSentFHi}{+1.0}
\newcommand{\grMrcOaiScorerSentFLo}{\ensuremath{-}0.9}

\newcommand{\grMrcOaiScorerSentTFDelta}{+0.4}
\newcommand{\grMrcOaiScorerSentTFHi}{+1.5}
\newcommand{\grMrcOaiScorerSentTFLo}{\ensuremath{-}0.7}

\newcommand{\grMrcOaiScorerWordFDelta}{+7.3}
\newcommand{\grMrcOaiScorerWordFHi}{+8.3}
\newcommand{\grMrcOaiScorerWordFLo}{+6.5}

\newcommand{\grMrcOaiScorerWordTFDelta}{+10.0}
\newcommand{\grMrcOaiScorerWordTFHi}{+11.2}
\newcommand{\grMrcOaiScorerWordTFLo}{+8.9}

\newcommand{\grMrcRandSentTFEvWords}{26.1}
\newcommand{\grMrcRandSentTFSentCov}{29.3}

\newcommand{\grMrcRandWordTFEvWords}{25.0}
\newcommand{\grMrcRandWordTFSentCov}{94.5}

\newcommand{\grTokMaxDev}{1}

%% file: generated/p4_known.tex
\newcommand{\kqEiNDocs}{792}

\newcommand{\kqEiTFCommonDelta}{+12.6}

\newcommand{\kqEiTFDisjointDelta}{+8.1}
\newcommand{\kqEiTFDisjointHi}{+10.5}

\newcommand{\kqEiTFDisjointLo}{+5.6}

\newcommand{\kqEiTFHiddenDelta}{+11.6}
\newcommand{\kqEiTFHiddenHi}{+13.8}

\newcommand{\kqEiTFHiddenLo}{+9.5}

\newcommand{\kqEiTFInteractionDelta}{+12.7}
\newcommand{\kqEiTFInteractionHi}{+18.1}

\newcommand{\kqEiTFInteractionLo}{+7.6}

\newcommand{\kqEiTFKnownDelta}{+22.6}
\newcommand{\kqEiTFKnownHi}{+25.8}

\newcommand{\kqEiTFKnownLo}{+19.6}

\newcommand{\kqEiTFOverlapDelta}{+20.6}
\newcommand{\kqEiTFOverlapHi}{+24.5}

\newcommand{\kqEiTFOverlapLo}{+16.9}

\newcommand{\kqFeverNDocs}{200}

\newcommand{\kqFeverTFCommonDelta}{+15.5}

\newcommand{\kqFeverTFDisjointDelta}{\ensuremath{-}6.6}
\newcommand{\kqFeverTFDisjointHi}{\ensuremath{-}1.6}

\newcommand{\kqFeverTFDisjointLo}{\ensuremath{-}12.0}

\newcommand{\kqFeverTFHiddenDelta}{+16.0}
\newcommand{\kqFeverTFHiddenHi}{+20.3}

\newcommand{\kqFeverTFHiddenLo}{+11.9}

\newcommand{\kqFeverTFInteractionDelta}{+35.7}
\newcommand{\kqFeverTFInteractionHi}{+42.7}

\newcommand{\kqFeverTFInteractionLo}{+29.0}

\newcommand{\kqFeverTFKnownDelta}{+39.5}
\newcommand{\kqFeverTFKnownHi}{+46.5}

\newcommand{\kqFeverTFKnownLo}{+33.0}

\newcommand{\kqFeverTFOverlapDelta}{+29.6}
\newcommand{\kqFeverTFOverlapHi}{+34.0}

\newcommand{\kqFeverTFOverlapLo}{+24.7}

\newcommand{\kqIntercept}{\ensuremath{-}1.49}

\newcommand{\kqLineEiDisjointDKept}{+24.0}

\newcommand{\kqLineEiDisjointPred}{+9.1}
\newcommand{\kqLineEiOverlapDKept}{+61.8}

\newcommand{\kqLineEiOverlapPred}{+23.0}
\newcommand{\kqLineFeverDisjointDKept}{\ensuremath{-}5.7}

\newcommand{\kqLineFeverDisjointPred}{\ensuremath{-}1.8}
\newcommand{\kqLineFeverOverlapDKept}{+53.9}
\newcommand{\kqLineFeverOverlapObs}{+29.6}
\newcommand{\kqLineFeverOverlapPred}{+20.1}
\newcommand{\kqLineMrcDecDisjointDKept}{\ensuremath{-}10.8}

\newcommand{\kqLineMrcDecDisjointPred}{\ensuremath{-}3.6}
\newcommand{\kqLineMrcDecOverlapDKept}{+30.3}
\newcommand{\kqLineMrcDecOverlapObs}{+4.5}
\newcommand{\kqLineMrcDecOverlapPred}{+11.4}
\newcommand{\kqLineMrcOaiDisjointDKept}{\ensuremath{-}5.4}

\newcommand{\kqLineMrcOaiDisjointPred}{\ensuremath{-}1.6}
\newcommand{\kqLineMrcOaiOverlapDKept}{+8.0}

\newcommand{\kqLineMrcOaiOverlapPred}{+3.3}
\newcommand{\kqMAE}{3.4}
\newcommand{\kqMissMax}{9.5}
\newcommand{\kqMissSecond}{6.9}
\newcommand{\kqMissThird}{4.8}

\newcommand{\kqMrcDecTFCommonDelta}{\ensuremath{-}0.6}

\newcommand{\kqMrcDecTFDisjointDelta}{\ensuremath{-}3.3}
\newcommand{\kqMrcDecTFDisjointHi}{\ensuremath{-}1.8}

\newcommand{\kqMrcDecTFDisjointLo}{\ensuremath{-}4.9}

\newcommand{\kqMrcDecTFHiddenDelta}{\ensuremath{-}0.7}
\newcommand{\kqMrcDecTFHiddenHi}{+0.6}
\newcommand{\kqMrcDecTFHiddenHiNinety}{+0.4}

\newcommand{\kqMrcDecTFHiddenLo}{\ensuremath{-}2.0}
\newcommand{\kqMrcDecTFHiddenLoNinety}{\ensuremath{-}1.7}

\newcommand{\kqMrcDecTFInteractionDelta}{+7.6}
\newcommand{\kqMrcDecTFInteractionHi}{+10.4}

\newcommand{\kqMrcDecTFInteractionLo}{+4.9}

\newcommand{\kqMrcDecTFKnownDelta}{+6.2}
\newcommand{\kqMrcDecTFKnownHi}{+9.7}

\newcommand{\kqMrcDecTFKnownLo}{+3.0}

\newcommand{\kqMrcDecTFOverlapDelta}{+4.5}
\newcommand{\kqMrcDecTFOverlapHi}{+6.9}

\newcommand{\kqMrcDecTFOverlapLo}{+2.3}

\newcommand{\kqMrcNDocs}{300}

\newcommand{\kqMrcOaiFHiddenDelta}{+0.3}
\newcommand{\kqMrcOaiFHiddenHi}{+1.4}

\newcommand{\kqMrcOaiFHiddenLo}{\ensuremath{-}0.9}

\newcommand{\kqMrcOaiFKnownDelta}{+8.7}
\newcommand{\kqMrcOaiFKnownHi}{+11.4}

\newcommand{\kqMrcOaiFKnownLo}{+5.8}

\newcommand{\kqMrcOaiTFCommonDelta}{\ensuremath{-}0.6}

\newcommand{\kqMrcOaiTFDisjointDelta}{\ensuremath{-}1.2}
\newcommand{\kqMrcOaiTFDisjointHi}{+0.4}

\newcommand{\kqMrcOaiTFDisjointLo}{\ensuremath{-}2.7}

\newcommand{\kqMrcOaiTFHiddenDelta}{\ensuremath{-}0.4}
\newcommand{\kqMrcOaiTFHiddenHi}{+0.6}
\newcommand{\kqMrcOaiTFHiddenHiNinety}{+0.5}

\newcommand{\kqMrcOaiTFHiddenLo}{\ensuremath{-}1.5}
\newcommand{\kqMrcOaiTFHiddenLoNinety}{\ensuremath{-}1.4}

\newcommand{\kqMrcOaiTFInteractionDelta}{+3.0}
\newcommand{\kqMrcOaiTFInteractionHi}{+5.8}

\newcommand{\kqMrcOaiTFInteractionLo}{+0.3}

\newcommand{\kqMrcOaiTFKnownDelta}{+8.0}
\newcommand{\kqMrcOaiTFKnownHi}{+11.1}

\newcommand{\kqMrcOaiTFKnownLo}{+5.0}

\newcommand{\kqMrcOaiTFOverlapDelta}{+1.3}
\newcommand{\kqMrcOaiTFOverlapHi}{+3.5}

\newcommand{\kqMrcOaiTFOverlapLo}{\ensuremath{-}0.7}

\newcommand{\kqSlope}{0.416}
\newcommand{\kqSlopeHi}{0.462}
\newcommand{\kqSlopeLo}{0.369}

%% file: generated/p4_cost.tex
\newcommand{\csMrcKbar}{10.2}

\newcommand{\csEiKbar}{3.7}

\newcommand{\csMrcCompactRatioKbar}{1.05}
\newcommand{\csMrcCompactBreakEven}{14.0}

\newcommand{\csMrcGateRatioKbar}{1.42}

\newcommand{\csFeverCompactRatioKbar}{0.62}

\newcommand{\csFeverGateRatioKbar}{1.12}

\newcommand{\csEiCompactRatioKbar}{1.55}
\newcommand{\csEiCompactBreakEven}{6.6}

\newcommand{\csEiGateRatioKbar}{1.83}

\newcommand{\csLocomoCompactRatioKbar}{0.33}
\newcommand{\csLocomoCompactBreakEven}{11.0}

\newcommand{\csLocomoGateRatioKbar}{0.59}

\newcommand{\csMrcMeasRatio}{1.15}
\newcommand{\csFeverMeasRatio}{1.23}
\newcommand{\csEiMeasRatio}{3.94}
\newcommand{\csLocomoMeasRatio}{2.53}

\newcommand{\csFeverCalOneQuestionReadMAE}{2.2}

\newcommand{\csEiCalOneQuestionReadMAE}{8.2}

\newcommand{\csLocomoCalOneQuestionReadMAE}{1.6}

\newcommand{\csMrcLinguaTFTokShare}{23.3}

\newcommand{\csFeverLinguaTFTokShare}{24.6}

\newcommand{\csEiTypedTFTokShare}{28.9}

\newcommand{\csEiTypedFTokShare}{53.5}

\newcommand{\csEiLinguaTFTokShare}{25.8}

\newcommand{\csEiTfidfTFTokShare}{30.4}

\newcommand{\csEiTfidfFTokShare}{59.0}

\newcommand{\csLocomoAgnosticTFTokShare}{32.3}

\newcommand{\csLocomoAgnosticFTokShare}{60.6}

\newcommand{\csFeverKDKeep}{117}
\newcommand{\csFeverKDDrop}{106}
\newcommand{\csFeverKDDiff}{+10.8}
\newcommand{\csFeverKDDiffLo}{+8.3}
\newcommand{\csFeverKDDiffHi}{+13.1}

\newcommand{\csEiKDKeep}{1,328}
\newcommand{\csEiKDDrop}{1,294}
\newcommand{\csEiKDDiff}{+33.2}
\newcommand{\csEiKDDiffLo}{+27.3}
\newcommand{\csEiKDDiffHi}{+39.4}

\newcommand{\csLocomoKDKeep}{6,252}
\newcommand{\csLocomoKDDrop}{6,225}
\newcommand{\csLocomoKDDiff}{+26.6}
\newcommand{\csLocomoKDDiffLo}{+18.7}
\newcommand{\csLocomoKDDiffHi}{+34.2}

%% file: generated/p4_reader.tex
\newcommand{\rdEiQwenFullAcc}{91.0}

\newcommand{\rdEiTypedRandFHiddenDelta}{+4.3}
\newcommand{\rdEiTypedRandFHiddenHi}{+5.9}
\newcommand{\rdEiTypedRandFHiddenLo}{+2.7}

\newcommand{\rdEiTypedRandTFDisjointDelta}{+8.5}
\newcommand{\rdEiTypedRandTFDisjointHi}{+10.8}
\newcommand{\rdEiTypedRandTFDisjointLo}{+6.1}

\newcommand{\rdEiTypedRandTFHiddenDelta}{+11.0}
\newcommand{\rdEiTypedRandTFHiddenHi}{+13.1}
\newcommand{\rdEiTypedRandTFHiddenLo}{+8.9}

\newcommand{\rdEiTypedRandTFKnownDelta}{+17.4}
\newcommand{\rdEiTypedRandTFKnownHi}{+20.3}
\newcommand{\rdEiTypedRandTFKnownLo}{+14.5}

\newcommand{\rdEiTypedRandTFOverlapDelta}{+18.8}
\newcommand{\rdEiTypedRandTFOverlapHi}{+23.0}
\newcommand{\rdEiTypedRandTFOverlapLo}{+14.6}

\newcommand{\rdFeverAgnRandTFHiddenDelta}{+2.7}
\newcommand{\rdFeverAgnRandTFHiddenHi}{+6.6}
\newcommand{\rdFeverAgnRandTFHiddenLo}{\ensuremath{-}1.1}

\newcommand{\rdFeverQwenFullAcc}{95.1}

\newcommand{\rdFeverQwenLinguaTFHidden}{81.3}

\newcommand{\rdFeverQwenTypedTFHidden}{77.7}

\newcommand{\rdFeverTypedRandFHiddenDelta}{+10.1}
\newcommand{\rdFeverTypedRandFHiddenHi}{+13.9}
\newcommand{\rdFeverTypedRandFHiddenLo}{+6.6}

\newcommand{\rdFeverTypedRandTFDisjointDelta}{+1.3}
\newcommand{\rdFeverTypedRandTFDisjointHi}{+6.1}
\newcommand{\rdFeverTypedRandTFDisjointLo}{\ensuremath{-}3.4}

\newcommand{\rdFeverTypedRandTFHiddenDelta}{+18.5}
\newcommand{\rdFeverTypedRandTFHiddenHi}{+22.5}
\newcommand{\rdFeverTypedRandTFHiddenLo}{+14.3}

\newcommand{\rdFeverTypedRandTFKnownDelta}{+36.0}
\newcommand{\rdFeverTypedRandTFKnownHi}{+43.5}
\newcommand{\rdFeverTypedRandTFKnownLo}{+28.5}

\newcommand{\rdFeverTypedRandTFOverlapDelta}{+27.3}
\newcommand{\rdFeverTypedRandTFOverlapHi}{+32.0}
\newcommand{\rdFeverTypedRandTFOverlapLo}{+22.7}

\newcommand{\rdMrcAgnRandTFHiddenDelta}{\ensuremath{-}0.3}
\newcommand{\rdMrcAgnRandTFHiddenHi}{+0.8}
\newcommand{\rdMrcAgnRandTFHiddenLo}{\ensuremath{-}1.4}

\newcommand{\rdMrcAllNo}{55.6}

\newcommand{\rdMrcQwenFullAcc}{89.2}
\newcommand{\rdMrcQwenFullHi}{90.0}

\newcommand{\rdMrcQwenFullLo}{88.3}

\newcommand{\rdMrcQwenLinguaTFHidden}{81.3}

\newcommand{\rdMrcQwenTypedTFHidden}{66.5}

\newcommand{\rdMrcTypedRandFHiddenDelta}{+0.5}
\newcommand{\rdMrcTypedRandFHiddenHi}{+1.9}
\newcommand{\rdMrcTypedRandFHiddenLo}{\ensuremath{-}1.0}

\newcommand{\rdMrcTypedRandTFDisjointDelta}{\ensuremath{-}1.7}
\newcommand{\rdMrcTypedRandTFDisjointHi}{\ensuremath{-}0.2}
\newcommand{\rdMrcTypedRandTFDisjointLo}{\ensuremath{-}3.2}

\newcommand{\rdMrcTypedRandTFHiddenDelta}{\ensuremath{-}0.6}
\newcommand{\rdMrcTypedRandTFHiddenHi}{+0.6}
\newcommand{\rdMrcTypedRandTFHiddenLo}{\ensuremath{-}1.8}

\newcommand{\rdMrcTypedRandTFKnownDelta}{+10.4}
\newcommand{\rdMrcTypedRandTFKnownHi}{+13.3}
\newcommand{\rdMrcTypedRandTFKnownLo}{+7.4}

\newcommand{\rdMrcTypedRandTFOverlapDelta}{+3.5}
\newcommand{\rdMrcTypedRandTFOverlapHi}{+5.7}
\newcommand{\rdMrcTypedRandTFOverlapLo}{+1.5}

\newcommand{\rdQwenCalls}{337,455}

%% file: generated/p4_tau.tex
\newcommand{\taAccAntH}{75.9}
\newcommand{\taAccAntK}{47.2}
\newcommand{\taAccECovEH}{66.4}
\newcommand{\taAccECovEK}{51.3}
\newcommand{\taAccECovTH}{74.8}
\newcommand{\taAccECovTK}{49.3}
\newcommand{\taAccEmptyH}{20.2}
\newcommand{\taAccEmptyK}{0.0}
\newcommand{\taAccFullH}{91.0}
\newcommand{\taAccFullK}{70.5}
\newcommand{\taAccGistH}{50.2}
\newcommand{\taAccGistK}{47.1}

\newcommand{\taAccHybridH}{56.8}
\newcommand{\taAccHybridK}{52.6}
\newcommand{\taAccLinguaH}{48.9}
\newcommand{\taAccLinguaK}{19.5}
\newcommand{\taAccOracleH}{93.2}
\newcommand{\taAccOracleK}{45.5}
\newcommand{\taAccRandomH}{44.9}
\newcommand{\taAccRandomK}{23.7}
\newcommand{\taAccRecencyH}{53.9}
\newcommand{\taAccRecencyK}{28.9}
\newcommand{\taAccTfidfH}{49.0}
\newcommand{\taAccTfidfK}{37.6}
\newcommand{\taAccTypedH}{67.1}
\newcommand{\taAccTypedK}{59.2}
\newcommand{\taAccTypedRecH}{67.1}
\newcommand{\taAccTypedRecK}{59.9}

\newcommand{\taAirAccEmptyH}{27.2}

\newcommand{\taAirAccFullH}{76.7}

\newcommand{\taAirAccRandomH}{60.0}

\newcommand{\taAirAccTypedH}{66.8}

\newcommand{\taAnsAntVsTypedHDelta}{+2.9}
\newcommand{\taAnsAntVsTypedHHi}{+8.0}
\newcommand{\taAnsAntVsTypedHLo}{\ensuremath{-}1.5}

\newcommand{\taAnsECovTVsTypedHDelta}{+5.8}
\newcommand{\taAnsECovTVsTypedHHi}{+13.6}
\newcommand{\taAnsECovTVsTypedHLo}{\ensuremath{-}1.4}

\newcommand{\taAnsLinguaVsRandomHDelta}{+4.0}
\newcommand{\taAnsLinguaVsRandomHHi}{+8.9}
\newcommand{\taAnsLinguaVsRandomHLo}{\ensuremath{-}1.2}

\newcommand{\taAnsRecencyVsRandomHDelta}{+4.4}
\newcommand{\taAnsRecencyVsRandomHHi}{+11.2}
\newcommand{\taAnsRecencyVsRandomHLo}{\ensuremath{-}1.8}

\newcommand{\taAnsTypedVsRandomHDelta}{+23.0}
\newcommand{\taAnsTypedVsRandomHHi}{+27.8}
\newcommand{\taAnsTypedVsRandomHLo}{+17.4}

\newcommand{\taAntVsRandomHDelta}{+31.0}
\newcommand{\taAntVsRandomHHi}{+36.6}

\newcommand{\taAntVsRandomHLo}{+24.7}

\newcommand{\taAntVsTypedHDelta}{+8.7}

\newcommand{\taAntVsTypedKDelta}{\ensuremath{-}12.0}

\newcommand{\taECovEVsECovTHDelta}{\ensuremath{-}10.9}

\newcommand{\taECovEVsECovTHHiNinety}{\ensuremath{-}4.3}

\newcommand{\taECovEVsECovTHLoNinety}{\ensuremath{-}18.2}

\newcommand{\taECovSubTypedH}{62.1}
\newcommand{\taECovSubTypedK}{59.9}
\newcommand{\taECovTVsGistHDelta}{+26.3}
\newcommand{\taECovTVsGistHHi}{+36.0}

\newcommand{\taECovTVsGistHLo}{+16.4}

\newcommand{\taECovTVsTypedHDelta}{+12.8}
\newcommand{\taECovTVsTypedHHi}{+20.8}

\newcommand{\taECovTVsTypedHLo}{+5.2}

\newcommand{\taECovTVsTypedKDelta}{\ensuremath{-}10.6}

\newcommand{\taECovTVsTypedKHiNinety}{\ensuremath{-}2.9}

\newcommand{\taECovTVsTypedKLoNinety}{\ensuremath{-}17.8}

\newcommand{\taFullItemAcc}{88.8}

\newcommand{\taGAnt}{0.64}

\newcommand{\taGECovE}{0.45}
\newcommand{\taGECovEHi}{0.56}
\newcommand{\taGECovELo}{0.33}
\newcommand{\taGECovT}{0.64}

\newcommand{\taGGist}{0.11}
\newcommand{\taGGistHi}{0.24}
\newcommand{\taGGistLo}{\ensuremath{-}0.03}
\newcommand{\taGHybrid}{0.25}
\newcommand{\taGHybridHi}{0.39}
\newcommand{\taGHybridLo}{0.09}
\newcommand{\taGLingua}{0.08}

\newcommand{\taGRecency}{0.19}

\newcommand{\taGTfidf}{0.09}
\newcommand{\taGTyped}{0.46}
\newcommand{\taGTypedHi}{0.59}
\newcommand{\taGTypedLo}{0.31}
\newcommand{\taGTypedRec}{0.46}
\newcommand{\taGTypedRecHi}{0.60}
\newcommand{\taGTypedRecLo}{0.31}

\newcommand{\taGoldAntH}{77.1}

\newcommand{\taGoldECovTH}{73.6}

\newcommand{\taGoldKeptAcc}{85.9}
\newcommand{\taGoldKeptVsLost}{+67.0}
\newcommand{\taGoldLinguaH}{52.3}

\newcommand{\taGoldLostAcc}{18.9}
\newcommand{\taGoldOracleH}{100.0}

\newcommand{\taGoldRandomH}{42.4}

\newcommand{\taGoldRecencyH}{42.9}

\newcommand{\taGoldTfidfH}{48.0}

\newcommand{\taGoldTypedH}{66.4}

\newcommand{\taHybridVsTypedHDelta}{\ensuremath{-}10.4}

\newcommand{\taHybridVsTypedHHiNinety}{\ensuremath{-}5.6}

\newcommand{\taHybridVsTypedHLoNinety}{\ensuremath{-}15.0}

\newcommand{\taLinguaVsRandomHDelta}{+4.0}
\newcommand{\taLinguaVsRandomHHi}{+9.3}

\newcommand{\taLinguaVsRandomHLo}{\ensuremath{-}1.1}

\newcommand{\taLinguaVsTypedHDelta}{\ensuremath{-}18.2}

\newcommand{\taMFiveMet}{met}
\newcommand{\taMFourMet}{not met}
\newcommand{\taMOneMet}{met}
\newcommand{\taMSixMet}{not met}
\newcommand{\taMThreeMet}{not met}
\newcommand{\taMTwoMet}{not met}
\newcommand{\taNAirStates}{41}

\newcommand{\taNECovStates}{120}
\newcommand{\taNHidden}{837}
\newcommand{\taNKnown}{407}
\newcommand{\taNMet}{8}
\newcommand{\taNPilot}{40}
\newcommand{\taNPred}{13}
\newcommand{\taNStates}{390}
\newcommand{\taNTasks}{81}

\newcommand{\taQwenAccAntH}{80.0}

\newcommand{\taQwenAccECovTH}{77.3}

\newcommand{\taQwenAccEmptyH}{27.7}

\newcommand{\taQwenAccFullH}{96.4}

\newcommand{\taQwenAccLinguaH}{52.4}

\newcommand{\taQwenAccOracleH}{97.3}

\newcommand{\taQwenAccRandomH}{52.3}

\newcommand{\taQwenAccRecencyH}{56.4}

\newcommand{\taQwenAccTypedH}{74.0}

\newcommand{\taQwenTypedVsRandomHDelta}{+21.7}
\newcommand{\taQwenTypedVsRandomHHi}{+27.6}

\newcommand{\taQwenTypedVsRandomHLo}{+15.7}

\newcommand{\taRecGistPtrAUROC}{0.72}
\newcommand{\taRecGistPtrAccA}{67.7}
\newcommand{\taRecGistPtrAccB}{69.1}
\newcommand{\taRecGistPtrAccC}{69.7}
\newcommand{\taRecGistPtrFetchedA}{64.6}
\newcommand{\taRecGistPtrFetchedB}{73.2}
\newcommand{\taRecGistPtrFetchedC}{79.8}
\newcommand{\taRecGistPtrFirst}{51.3}
\newcommand{\taRecGistPtrTokShareA}{92.9}
\newcommand{\taRecGistPtrTokShareB}{102.0}
\newcommand{\taRecGistPtrTokShareC}{108.4}
\newcommand{\taRecTypedStubAUROC}{0.73}
\newcommand{\taRecTypedStubAccA}{72.3}
\newcommand{\taRecTypedStubAccB}{72.6}
\newcommand{\taRecTypedStubAccC}{72.3}
\newcommand{\taRecTypedStubFetchedA}{65.1}
\newcommand{\taRecTypedStubFetchedB}{72.8}
\newcommand{\taRecTypedStubFetchedC}{79.5}
\newcommand{\taRecTypedStubFirst}{64.0}
\newcommand{\taRecTypedStubTokShareA}{96.6}
\newcommand{\taRecTypedStubTokShareB}{104.5}
\newcommand{\taRecTypedStubTokShareC}{111.7}
\newcommand{\taRecencyLastShare}{8.5}
\newcommand{\taRecencyVsRandomHDelta}{+9.1}
\newcommand{\taRecencyVsRandomHHi}{+15.4}

\newcommand{\taRecencyVsRandomHLo}{+3.4}

\newcommand{\taRefusalAnt}{6.4}

\newcommand{\taRefusalECovT}{6.9}
\newcommand{\taRefusalEmpty}{66.7}

\newcommand{\taRefusalLingua}{16.2}

\newcommand{\taRefusalPass}{passes}
\newcommand{\taRefusalRandom}{16.0}
\newcommand{\taRefusalRate}{1.9}

\newcommand{\taRefusalTyped}{10.9}

\newcommand{\taRefusalUpper}{2.9}
\newcommand{\taTFourMet}{met}
\newcommand{\taTNineMet}{met}
\newcommand{\taTOneMet}{met}
\newcommand{\taTSevenMet}{not met}
\newcommand{\taTSixMet}{met}
\newcommand{\taTThreeMet}{met}
\newcommand{\taTTwoMet}{met}

\newcommand{\taTypedVsRandomHDelta}{+22.3}
\newcommand{\taTypedVsRandomHHi}{+28.9}

\newcommand{\taTypedVsRandomHLo}{+15.1}

%% file: generated/p4_hybrid.tex
\newcommand{\hyFeverHThreeMet}{met}
\newcommand{\hyFeverHTwoMet}{met}

\newcommand{\hyFeverHybHalfLinguaTFHiddenDelta}{+2.6}

\newcommand{\hyFeverHybHalfLinguaTFHiddenHiNinety}{+4.5}

\newcommand{\hyFeverHybHalfLinguaTFHiddenLoNinety}{+0.7}

\newcommand{\hyFeverHybHalfLinguaTFKnownDelta}{+9.5}
\newcommand{\hyFeverHybHalfLinguaTFKnownHi}{+15.5}

\newcommand{\hyFeverHybHalfLinguaTFKnownLo}{+3.0}

\newcommand{\hyFeverHybHalfTFEvHidden}{55.3}
\newcommand{\hyFeverHybHalfTFEvKnown}{73.2}
\newcommand{\hyFeverHybHalfTFHidden}{84.7}
\newcommand{\hyFeverHybHalfTFKnown}{89.0}

\newcommand{\hyFeverHybHalfTypedTFHiddenDelta}{+5.7}
\newcommand{\hyFeverHybHalfTypedTFHiddenHi}{+7.8}

\newcommand{\hyFeverHybHalfTypedTFHiddenLo}{+3.6}

\newcommand{\hyFeverHybQuarterTFEvHidden}{44.0}
\newcommand{\hyFeverHybQuarterTFEvKnown}{55.7}
\newcommand{\hyFeverHybQuarterTFHidden}{82.1}
\newcommand{\hyFeverHybQuarterTFKnown}{81.5}

\newcommand{\hyFeverLinguaTFEvHidden}{26.4}
\newcommand{\hyFeverLinguaTFEvKnown}{26.4}
\newcommand{\hyFeverLinguaTFHidden}{82.2}
\newcommand{\hyFeverLinguaTFKnown}{79.5}

\newcommand{\hyFeverRandSentTFHidden}{61.2}
\newcommand{\hyFeverRandSentTFKnown}{56.8}

\newcommand{\hyFeverRandWordTFHidden}{67.5}
\newcommand{\hyFeverRandWordTFKnown}{62.2}

\newcommand{\hyFeverTypedTFEvHidden}{59.1}
\newcommand{\hyFeverTypedTFEvKnown}{81.1}
\newcommand{\hyFeverTypedTFHidden}{79.0}
\newcommand{\hyFeverTypedTFKnown}{92.5}

\newcommand{\hyMrcDecHThreeMet}{not met}
\newcommand{\hyMrcDecHTwoMet}{not met}

\newcommand{\hyMrcDecHybHalfLinguaTFHiddenDelta}{\ensuremath{-}2.4}

\newcommand{\hyMrcDecHybHalfLinguaTFHiddenHiNinety}{\ensuremath{-}1.5}

\newcommand{\hyMrcDecHybHalfLinguaTFHiddenLoNinety}{\ensuremath{-}3.2}

\newcommand{\hyMrcDecHybHalfLinguaTFKnownDelta}{+0.7}
\newcommand{\hyMrcDecHybHalfLinguaTFKnownHi}{+3.5}

\newcommand{\hyMrcDecHybHalfLinguaTFKnownLo}{\ensuremath{-}1.8}

\newcommand{\hyMrcDecHybHalfTFEvHidden}{26.1}
\newcommand{\hyMrcDecHybHalfTFEvKnown}{59.1}
\newcommand{\hyMrcDecHybHalfTFHidden}{65.7}
\newcommand{\hyMrcDecHybHalfTFKnown}{70.7}

\newcommand{\hyMrcDecHybHalfTypedTFHiddenDelta}{+4.6}
\newcommand{\hyMrcDecHybHalfTypedTFHiddenHi}{+5.8}

\newcommand{\hyMrcDecHybHalfTypedTFHiddenLo}{+3.5}

\newcommand{\hyMrcDecHybQuarterTFEvHidden}{26.1}
\newcommand{\hyMrcDecHybQuarterTFEvKnown}{43.7}
\newcommand{\hyMrcDecHybQuarterTFHidden}{66.8}
\newcommand{\hyMrcDecHybQuarterTFKnown}{70.8}

\newcommand{\hyMrcDecLinguaTFEvHidden}{26.3}
\newcommand{\hyMrcDecLinguaTFEvKnown}{26.1}
\newcommand{\hyMrcDecLinguaTFHidden}{68.0}
\newcommand{\hyMrcDecLinguaTFKnown}{70.0}

\newcommand{\hyMrcDecRandSentTFHidden}{62.2}
\newcommand{\hyMrcDecRandSentTFKnown}{61.1}

\newcommand{\hyMrcDecRandWordTFHidden}{62.6}
\newcommand{\hyMrcDecRandWordTFKnown}{63.3}

\newcommand{\hyMrcDecTypedTFEvHidden}{26.3}
\newcommand{\hyMrcDecTypedTFEvKnown}{73.9}
\newcommand{\hyMrcDecTypedTFHidden}{61.0}
\newcommand{\hyMrcDecTypedTFKnown}{69.0}

\newcommand{\hyMrcOaiHThreeMet}{not met}
\newcommand{\hyMrcOaiHTwoMet}{not met}

\newcommand{\hyMrcOaiHybHalfLinguaTFHiddenDelta}{\ensuremath{-}4.2}

\newcommand{\hyMrcOaiHybHalfLinguaTFHiddenHiNinety}{\ensuremath{-}3.3}

\newcommand{\hyMrcOaiHybHalfLinguaTFHiddenLoNinety}{\ensuremath{-}5.1}

\newcommand{\hyMrcOaiHybHalfLinguaTFKnownDelta}{+1.3}
\newcommand{\hyMrcOaiHybHalfLinguaTFKnownHi}{+3.4}

\newcommand{\hyMrcOaiHybHalfLinguaTFKnownLo}{\ensuremath{-}1.0}

\newcommand{\hyMrcOaiHybHalfTFEvHidden}{25.6}
\newcommand{\hyMrcOaiHybHalfTFEvKnown}{43.9}
\newcommand{\hyMrcOaiHybHalfTFHidden}{76.5}
\newcommand{\hyMrcOaiHybHalfTFKnown}{82.2}

\newcommand{\hyMrcOaiHybHalfTypedTFHiddenDelta}{+8.3}
\newcommand{\hyMrcOaiHybHalfTypedTFHiddenHi}{+9.5}

\newcommand{\hyMrcOaiHybHalfTypedTFHiddenLo}{+7.0}

\newcommand{\hyMrcOaiHybQuarterTFEvHidden}{26.2}
\newcommand{\hyMrcOaiHybQuarterTFEvKnown}{35.4}
\newcommand{\hyMrcOaiHybQuarterTFHidden}{79.6}
\newcommand{\hyMrcOaiHybQuarterTFKnown}{79.6}

\newcommand{\hyMrcOaiLinguaTFEvHidden}{26.3}
\newcommand{\hyMrcOaiLinguaTFEvKnown}{26.1}
\newcommand{\hyMrcOaiLinguaTFHidden}{80.7}
\newcommand{\hyMrcOaiLinguaTFKnown}{80.9}

\newcommand{\hyMrcOaiRandSentTFHidden}{68.3}
\newcommand{\hyMrcOaiRandSentTFKnown}{67.3}

\newcommand{\hyMrcOaiRandWordTFHidden}{70.9}
\newcommand{\hyMrcOaiRandWordTFKnown}{71.9}

\newcommand{\hyMrcOaiTypedTFEvHidden}{24.4}
\newcommand{\hyMrcOaiTypedTFEvKnown}{51.2}
\newcommand{\hyMrcOaiTypedTFHidden}{67.9}
\newcommand{\hyMrcOaiTypedTFKnown}{78.7}

%% file: generated/p4_whywords.tex
\newcommand{\wwFeverHybHalfTFHiddenAcc}{84.7}

\newcommand{\wwFeverHybHalfTFHiddenEvContent}{59.2}
\newcommand{\wwFeverHybHalfTFHiddenEvWords}{55.3}

\newcommand{\wwFeverHybHalfTypedTFHiddenEvBreadthGap}{\ensuremath{-}5.8}

\newcommand{\wwFeverHybHalfTypedTFHiddenEvContentGap}{\ensuremath{-}0.5}
\newcommand{\wwFeverHybHalfTypedTFHiddenEvContentHi}{+0.8}

\newcommand{\wwFeverHybHalfTypedTFHiddenEvContentLinGap}{+5.7}

\newcommand{\wwFeverHybHalfTypedTFHiddenEvContentLo}{\ensuremath{-}1.8}

\newcommand{\wwFeverHybHalfTypedTFHiddenEvWordsGap}{+0.4}

\newcommand{\wwFeverHybHalfTypedTFHiddenRawGap}{+5.7}
\newcommand{\wwFeverHybHalfTypedTFHiddenRawHi}{+7.9}
\newcommand{\wwFeverHybHalfTypedTFHiddenRawLo}{+3.6}

\newcommand{\wwFeverLinguaRandSentTFHiddenEvBreadthGap}{\ensuremath{-}4.1}

\newcommand{\wwFeverLinguaRandSentTFHiddenEvContentGap}{+1.3}
\newcommand{\wwFeverLinguaRandSentTFHiddenEvContentHi}{+3.5}

\newcommand{\wwFeverLinguaRandSentTFHiddenEvContentLinGap}{+14.9}

\newcommand{\wwFeverLinguaRandSentTFHiddenEvContentLo}{\ensuremath{-}1.1}

\newcommand{\wwFeverLinguaRandSentTFHiddenEvWordsGap}{+2.7}

\newcommand{\wwFeverLinguaRandSentTFHiddenRawGap}{+21.0}
\newcommand{\wwFeverLinguaRandSentTFHiddenRawHi}{+23.4}
\newcommand{\wwFeverLinguaRandSentTFHiddenRawLo}{+18.7}

\newcommand{\wwFeverLinguaRandWordTFHiddenEvBreadthGap}{+14.2}

\newcommand{\wwFeverLinguaRandWordTFHiddenEvContentGap}{+9.6}
\newcommand{\wwFeverLinguaRandWordTFHiddenEvContentHi}{+11.7}

\newcommand{\wwFeverLinguaRandWordTFHiddenEvContentLinGap}{+8.4}

\newcommand{\wwFeverLinguaRandWordTFHiddenEvContentLo}{+7.5}

\newcommand{\wwFeverLinguaRandWordTFHiddenEvWordsGap}{+15.7}

\newcommand{\wwFeverLinguaRandWordTFHiddenRawGap}{+14.7}
\newcommand{\wwFeverLinguaRandWordTFHiddenRawHi}{+16.9}
\newcommand{\wwFeverLinguaRandWordTFHiddenRawLo}{+12.6}

\newcommand{\wwFeverLinguaTFHiddenAcc}{82.2}

\newcommand{\wwFeverLinguaTFHiddenEvContent}{40.7}
\newcommand{\wwFeverLinguaTFHiddenEvWords}{26.4}

\newcommand{\wwFeverRandSentTFHiddenAcc}{61.2}

\newcommand{\wwFeverRandSentTFHiddenEvContent}{25.6}
\newcommand{\wwFeverRandSentTFHiddenEvWords}{25.7}

\newcommand{\wwFeverRandWordTFHiddenAcc}{67.5}

\newcommand{\wwFeverRandWordTFHiddenEvContent}{25.2}
\newcommand{\wwFeverRandWordTFHiddenEvWords}{25.2}

\newcommand{\wwFeverTypedTFHiddenAcc}{79.0}

\newcommand{\wwFeverTypedTFHiddenEvContent}{59.2}
\newcommand{\wwFeverTypedTFHiddenEvWords}{59.1}

\newcommand{\wwMrcDecHybHalfTFHiddenAcc}{65.7}

\newcommand{\wwMrcDecHybHalfTFHiddenEvContent}{37.6}
\newcommand{\wwMrcDecHybHalfTFHiddenEvWords}{26.1}

\newcommand{\wwMrcDecHybHalfTypedTFHiddenContentAnsGap}{+1.1}

\newcommand{\wwMrcDecHybHalfTypedTFHiddenEvBreadthGap}{\ensuremath{-}2.9}

\newcommand{\wwMrcDecHybHalfTypedTFHiddenEvContentGap}{+2.5}
\newcommand{\wwMrcDecHybHalfTypedTFHiddenEvContentHi}{+3.7}

\newcommand{\wwMrcDecHybHalfTypedTFHiddenEvContentLinGap}{+2.8}

\newcommand{\wwMrcDecHybHalfTypedTFHiddenEvContentLo}{+1.1}

\newcommand{\wwMrcDecHybHalfTypedTFHiddenEvWordsGap}{+4.6}

\newcommand{\wwMrcDecHybHalfTypedTFHiddenRawGap}{+4.7}
\newcommand{\wwMrcDecHybHalfTypedTFHiddenRawHi}{+5.9}
\newcommand{\wwMrcDecHybHalfTypedTFHiddenRawLo}{+3.5}

\newcommand{\wwMrcDecLinguaRandSentTFHiddenContentAnsGap}{+0.6}
\newcommand{\wwMrcDecLinguaRandSentTFHiddenContentAnsHi}{+1.8}

\newcommand{\wwMrcDecLinguaRandSentTFHiddenContentAnsLo}{\ensuremath{-}0.5}

\newcommand{\wwMrcDecLinguaRandSentTFHiddenEvBreadthGap}{\ensuremath{-}2.2}

\newcommand{\wwMrcDecLinguaRandSentTFHiddenEvContentGap}{+2.0}
\newcommand{\wwMrcDecLinguaRandSentTFHiddenEvContentHi}{+3.2}

\newcommand{\wwMrcDecLinguaRandSentTFHiddenEvContentLinGap}{+2.3}

\newcommand{\wwMrcDecLinguaRandSentTFHiddenEvContentLo}{+0.7}

\newcommand{\wwMrcDecLinguaRandSentTFHiddenEvWordsGap}{+5.5}

\newcommand{\wwMrcDecLinguaRandSentTFHiddenRawGap}{+6.0}
\newcommand{\wwMrcDecLinguaRandSentTFHiddenRawHi}{+7.3}
\newcommand{\wwMrcDecLinguaRandSentTFHiddenRawLo}{+4.7}

\newcommand{\wwMrcDecLinguaRandWordTFHiddenContentAnsGap}{+0.4}

\newcommand{\wwMrcDecLinguaRandWordTFHiddenEvBreadthGap}{+4.5}

\newcommand{\wwMrcDecLinguaRandWordTFHiddenEvContentGap}{+1.7}
\newcommand{\wwMrcDecLinguaRandWordTFHiddenEvContentHi}{+2.9}

\newcommand{\wwMrcDecLinguaRandWordTFHiddenEvContentLinGap}{+1.7}

\newcommand{\wwMrcDecLinguaRandWordTFHiddenEvContentLo}{+0.5}

\newcommand{\wwMrcDecLinguaRandWordTFHiddenEvWordsGap}{+5.3}

\newcommand{\wwMrcDecLinguaRandWordTFHiddenRawGap}{+5.6}
\newcommand{\wwMrcDecLinguaRandWordTFHiddenRawHi}{+6.7}
\newcommand{\wwMrcDecLinguaRandWordTFHiddenRawLo}{+4.3}

\newcommand{\wwMrcDecLinguaTFHiddenAcc}{68.1}

\newcommand{\wwMrcDecLinguaTFHiddenEvContent}{48.0}
\newcommand{\wwMrcDecLinguaTFHiddenEvWords}{26.3}

\newcommand{\wwMrcDecRandSentTFHiddenAcc}{62.1}

\newcommand{\wwMrcDecRandSentTFHiddenEvContent}{26.0}
\newcommand{\wwMrcDecRandSentTFHiddenEvWords}{26.2}

\newcommand{\wwMrcDecRandWordTFHiddenAcc}{62.6}

\newcommand{\wwMrcDecRandWordTFHiddenEvContent}{25.1}
\newcommand{\wwMrcDecRandWordTFHiddenEvWords}{25.0}

\newcommand{\wwMrcDecTypedTFHiddenAcc}{61.0}

\newcommand{\wwMrcDecTypedTFHiddenEvContent}{26.3}
\newcommand{\wwMrcDecTypedTFHiddenEvWords}{26.3}

\newcommand{\wwMrcOaiHybHalfTFHiddenAcc}{76.6}

\newcommand{\wwMrcOaiHybHalfTFHiddenEvContent}{37.3}
\newcommand{\wwMrcOaiHybHalfTFHiddenEvWords}{25.7}

\newcommand{\wwMrcOaiHybHalfTypedTFHiddenContentAnsGap}{+3.2}

\newcommand{\wwMrcOaiHybHalfTypedTFHiddenEvBreadthGap}{\ensuremath{-}0.7}

\newcommand{\wwMrcOaiHybHalfTypedTFHiddenEvContentGap}{+5.1}
\newcommand{\wwMrcOaiHybHalfTypedTFHiddenEvContentHi}{+6.4}

\newcommand{\wwMrcOaiHybHalfTypedTFHiddenEvContentLinGap}{+6.1}

\newcommand{\wwMrcOaiHybHalfTypedTFHiddenEvContentLo}{+3.8}

\newcommand{\wwMrcOaiHybHalfTypedTFHiddenEvWordsGap}{+7.2}

\newcommand{\wwMrcOaiHybHalfTypedTFHiddenRawGap}{+8.6}
\newcommand{\wwMrcOaiHybHalfTypedTFHiddenRawHi}{+9.9}
\newcommand{\wwMrcOaiHybHalfTypedTFHiddenRawLo}{+7.3}

\newcommand{\wwMrcOaiLinguaRandSentTFHiddenContentAnsGap}{+4.9}

\newcommand{\wwMrcOaiLinguaRandSentTFHiddenEvBreadthGap}{+2.6}

\newcommand{\wwMrcOaiLinguaRandSentTFHiddenEvContentGap}{+7.3}
\newcommand{\wwMrcOaiLinguaRandSentTFHiddenEvContentHi}{+8.4}

\newcommand{\wwMrcOaiLinguaRandSentTFHiddenEvContentLinGap}{+8.0}

\newcommand{\wwMrcOaiLinguaRandSentTFHiddenEvContentLo}{+6.1}

\newcommand{\wwMrcOaiLinguaRandSentTFHiddenEvWordsGap}{+10.1}

\newcommand{\wwMrcOaiLinguaRandSentTFHiddenRawGap}{+12.5}
\newcommand{\wwMrcOaiLinguaRandSentTFHiddenRawHi}{+13.6}
\newcommand{\wwMrcOaiLinguaRandSentTFHiddenRawLo}{+11.3}

\newcommand{\wwMrcOaiLinguaRandWordTFHiddenContentAnsGap}{+4.6}

\newcommand{\wwMrcOaiLinguaRandWordTFHiddenEvBreadthGap}{+9.3}

\newcommand{\wwMrcOaiLinguaRandWordTFHiddenEvContentGap}{+6.7}
\newcommand{\wwMrcOaiLinguaRandWordTFHiddenEvContentHi}{+8.0}

\newcommand{\wwMrcOaiLinguaRandWordTFHiddenEvContentLinGap}{+6.1}

\newcommand{\wwMrcOaiLinguaRandWordTFHiddenEvContentLo}{+5.5}

\newcommand{\wwMrcOaiLinguaRandWordTFHiddenEvWordsGap}{+10.5}

\newcommand{\wwMrcOaiLinguaRandWordTFHiddenRawGap}{+10.7}
\newcommand{\wwMrcOaiLinguaRandWordTFHiddenRawHi}{+11.9}
\newcommand{\wwMrcOaiLinguaRandWordTFHiddenRawLo}{+9.5}

\newcommand{\wwMrcOaiLinguaTFHiddenAcc}{80.8}

\newcommand{\wwMrcOaiLinguaTFHiddenEvContent}{48.0}
\newcommand{\wwMrcOaiLinguaTFHiddenEvWords}{26.3}

\newcommand{\wwMrcOaiRandSentTFHiddenAcc}{68.3}

\newcommand{\wwMrcOaiRandSentTFHiddenEvContent}{26.1}
\newcommand{\wwMrcOaiRandSentTFHiddenEvWords}{26.3}

\newcommand{\wwMrcOaiRandWordTFHiddenAcc}{70.8}

\newcommand{\wwMrcOaiRandWordTFHiddenEvContent}{25.2}
\newcommand{\wwMrcOaiRandWordTFHiddenEvWords}{25.0}

\newcommand{\wwMrcOaiTypedTFHiddenAcc}{67.9}

\newcommand{\wwMrcOaiTypedTFHiddenEvContent}{24.3}
\newcommand{\wwMrcOaiTypedTFHiddenEvWords}{24.4}

%% file: generated/p4_fever_support.tex
\newcommand{\fsHyLabAllLinguaN}{5,432}

\newcommand{\fsHyLabFalseLinguaRandSentDelta}{\ensuremath{-}2.2}
\newcommand{\fsHyLabFalseLinguaRandSentHi}{\ensuremath{-}1.0}
\newcommand{\fsHyLabFalseLinguaRandSentLo}{\ensuremath{-}3.7}
\newcommand{\fsHyLabFalseLinguaRandSentN}{1,582}

\newcommand{\fsHyLabTrueLinguaRandSentDelta}{+31.5}
\newcommand{\fsHyLabTrueLinguaRandSentHi}{+35.0}
\newcommand{\fsHyLabTrueLinguaRandSentLo}{+28.0}
\newcommand{\fsHyLabTrueLinguaRandSentN}{3,850}

\newcommand{\fsHyPartialShareLingua}{99}

\newcommand{\fsHyRelaxedLinguaRandSentBalDelta}{\ensuremath{-}19.8}
\newcommand{\fsHyRelaxedLinguaRandSentBalHi}{\ensuremath{-}12.0}
\newcommand{\fsHyRelaxedLinguaRandSentBalLo}{\ensuremath{-}29.7}

\newcommand{\fsHyRelaxedLinguaRandSentN}{675}

\newcommand{\fsHyStrictHybHalfTypedBalDelta}{\ensuremath{-}3.7}
\newcommand{\fsHyStrictHybHalfTypedBalHi}{+1.3}
\newcommand{\fsHyStrictHybHalfTypedBalLo}{\ensuremath{-}9.4}

\newcommand{\fsHyStrictHybHalfTypedN}{1,140}

\newcommand{\fsHyStrictRefLinguaN}{45}

\newcommand{\fsHyYesRLeRelaxLinguaAcc}{56.0}

\newcommand{\fsHyYesRLeRelaxLinguaN}{458}

\newcommand{\fsHyYesRLeRelaxRandSentAcc}{26.8}

\newcommand{\fsHyYesRLeRelaxRandSentN}{2,575}

\newcommand{\fsOrigStrictTypedRandSentBalDelta}{\ensuremath{-}0.2}
\newcommand{\fsOrigStrictTypedRandSentBalHi}{+3.0}
\newcommand{\fsOrigStrictTypedRandSentBalLo}{\ensuremath{-}3.4}

\newcommand{\fsOrigStrictTypedRandSentN}{3,685}

%% file: generated/p4_summary.tex
\newcommand{\smGptFocCutShare}{26.7}

\newcommand{\smGptFocHidden}{69.1}
\newcommand{\smGptFocKnown}{89.2}
\newcommand{\smGptFocLinguaHiddenDelta}{\ensuremath{-}11.4}
\newcommand{\smGptFocLinguaHiddenHi}{\ensuremath{-}9.8}
\newcommand{\smGptFocLinguaHiddenLo}{\ensuremath{-}12.8}

\newcommand{\smGptFocRandSentHiddenDelta}{+0.9}
\newcommand{\smGptFocRandSentHiddenHi}{+2.1}
\newcommand{\smGptFocRandSentHiddenLo}{\ensuremath{-}0.2}

\newcommand{\smGptFocRandSentKnownDelta}{+22.0}
\newcommand{\smGptFocRandSentKnownHi}{+24.1}
\newcommand{\smGptFocRandSentKnownLo}{+19.8}

\newcommand{\smGptFocSumFocHiddenDelta}{+3.9}
\newcommand{\smGptFocSumFocHiddenHi}{+4.9}
\newcommand{\smGptFocSumFocHiddenLo}{+3.0}

\newcommand{\smGptFocTokShare}{94.8}

\newcommand{\smGptGenCutShare}{74.3}

\newcommand{\smGptGenHidden}{79.3}
\newcommand{\smGptGenKnown}{78.9}
\newcommand{\smGptGenLinguaHiddenDelta}{\ensuremath{-}1.3}
\newcommand{\smGptGenLinguaHiddenHi}{\ensuremath{-}0.3}
\newcommand{\smGptGenLinguaHiddenLo}{\ensuremath{-}2.4}

\newcommand{\smGptGenRandSentHiddenDelta}{+10.9}
\newcommand{\smGptGenRandSentHiddenHi}{+12.1}
\newcommand{\smGptGenRandSentHiddenLo}{+9.8}

\newcommand{\smGptGenRandSentKnownDelta}{+11.5}
\newcommand{\smGptGenRandSentKnownHi}{+13.8}
\newcommand{\smGptGenRandSentKnownLo}{+9.2}

\newcommand{\smGptGenSumGenHiddenDelta}{+3.3}
\newcommand{\smGptGenSumGenHiddenHi}{+4.3}
\newcommand{\smGptGenSumGenHiddenLo}{+2.3}

\newcommand{\smGptGenTokShare}{99.2}

\newcommand{\smLinguaHidden}{80.7}
\newcommand{\smLinguaKnown}{80.9}

\newcommand{\smRandSentHidden}{68.3}
\newcommand{\smRandSentKnown}{67.3}

\newcommand{\smSumFocCutShare}{0.7}

\newcommand{\smSumFocHidden}{65.4}
\newcommand{\smSumFocKnown}{88.5}
\newcommand{\smSumFocLinguaHiddenDelta}{\ensuremath{-}15.1}
\newcommand{\smSumFocLinguaHiddenHi}{\ensuremath{-}13.6}
\newcommand{\smSumFocLinguaHiddenLo}{\ensuremath{-}16.7}

\newcommand{\smSumFocNoText}{3}
\newcommand{\smSumFocRandSentHiddenDelta}{\ensuremath{-}2.9}
\newcommand{\smSumFocRandSentHiddenHi}{\ensuremath{-}1.8}
\newcommand{\smSumFocRandSentHiddenLo}{\ensuremath{-}4.0}

\newcommand{\smSumFocRandSentKnownDelta}{+21.4}
\newcommand{\smSumFocRandSentKnownHi}{+23.5}
\newcommand{\smSumFocRandSentKnownLo}{+19.2}

\newcommand{\smSumFocTokShare}{69.5}

\newcommand{\smSumGenCutShare}{23.7}

\newcommand{\smSumGenHidden}{76.1}
\newcommand{\smSumGenKnown}{75.4}
\newcommand{\smSumGenLinguaHiddenDelta}{\ensuremath{-}4.5}
\newcommand{\smSumGenLinguaHiddenHi}{\ensuremath{-}3.2}
\newcommand{\smSumGenLinguaHiddenLo}{\ensuremath{-}5.7}

\newcommand{\smSumGenNoText}{4}
\newcommand{\smSumGenRandSentHiddenDelta}{+7.6}
\newcommand{\smSumGenRandSentHiddenHi}{+8.7}
\newcommand{\smSumGenRandSentHiddenLo}{+6.6}

\newcommand{\smSumGenRandSentKnownDelta}{+8.2}
\newcommand{\smSumGenRandSentKnownHi}{+10.3}
\newcommand{\smSumGenRandSentKnownLo}{+6.1}

\newcommand{\smSumGenTokShare}{88.4}

\newcommand{\smTypedHidden}{67.9}
\newcommand{\smTypedKnown}{78.7}

%% file: generated/compress_open_main.tex
\begin{tabular}{lllrr@{\hspace{1.5em}}l}
\toprule
 & & & \multicolumn{2}{c}{accuracy (\%)} & \\
\cmidrule(lr){4-5}
model & policy & budget & known & hidden & hidden $-$ random \\
\midrule
\texttt{open-decider} & typed & 25\% & 70 & 62 & \ci{\ensuremath{-}0.6}{\ensuremath{-}1.9}{+0.7} \\
\texttt{open-decider} & agnostic & 25\% & 63 & 63 & \ci{+0.3}{\ensuremath{-}0.9}{+1.4} \\
\texttt{open-decider} & random & 25\% & 61 & 62 & --- \\
\midrule
\texttt{open-decider} & typed & 50\% & 72 & 66 & \ci{\ensuremath{-}0.6}{\ensuremath{-}1.7}{+0.6} \\
\texttt{open-decider} & agnostic & 50\% & 67 & 67 & \ci{+0.4}{\ensuremath{-}0.7}{+1.6} \\
\texttt{open-decider} & random & 50\% & 66 & 67 & --- \\
\midrule
\texttt{gpt-6-luna} & typed & 25\% & 77 & 67 & \ci{\ensuremath{-}0.8}{\ensuremath{-}1.9}{+0.3} \\
\texttt{gpt-6-luna} & agnostic & 25\% & 67 & 67 & \ci{\ensuremath{-}0.8}{\ensuremath{-}1.9}{+0.3} \\
\texttt{gpt-6-luna} & random & 25\% & 67 & 68 & --- \\
\midrule
\texttt{gpt-6-luna} & typed & 50\% & 83 & 75 & \ci{\ensuremath{-}0.4}{\ensuremath{-}1.7}{+0.9} \\
\texttt{gpt-6-luna} & agnostic & 50\% & 76 & 75 & \ci{\ensuremath{-}0.3}{\ensuremath{-}1.6}{+1.0} \\
\texttt{gpt-6-luna} & random & 50\% & 75 & 76 & --- \\
\bottomrule
\end{tabular}

%% file: generated/p4_refusal_table.tex
\textit{(a) Refused answers (\%)}\\[2pt]
\begin{tabular}{lrrrr}
\toprule
 & \multicolumn{2}{c}{25\% budget} & \multicolumn{2}{c}{50\% budget} \\
\cmidrule(lr){2-3}\cmidrule(lr){4-5}
policy & known & hidden & known & hidden \\
\midrule
typed & 0.3 & 0.9 & 0.1 & 0.1 \\
agnostic & 1.0 & 1.2 & 0.0 & 0.0 \\
TF-IDF & 0.1 & 0.3 & 0.0 & 0.1 \\
lead & 0.5 & 0.3 & 0.2 & 0.1 \\
random & 0.4 & 0.5 & 0.1 & 0.1 \\
\midrule
random (all masks) & 0.5 & 0.5 & 0.1 & 0.1 \\
\midrule
typed, token-matched & 0.7 & 1.1 & 0.0 & 0.2 \\
LongLLMLingua, token-matched & 2.3 & 4.8 & 0.7 & 0.9 \\
MMR, token-matched & 0.6 & 0.9 & 0.1 & 0.2 \\
\LLMLtwo{}, token-matched & 1.7 & 1.0 & 0.1 & 0.0 \\
random, token-matched & 1.0 & 0.8 & 0.1 & 0.1 \\
\bottomrule
\end{tabular}

\medskip
\textit{(b) Contrasts (points, 95\% interval)}\\[2pt]
\begin{tabular}{llll}
\toprule
comparison & budget & refusals dropped & refusals wrong \\
\midrule
\multicolumn{4}{l}{\textit{token-matched}} \\
LongLLMLingua $-$ random, known & 25\% & \ci{+6.9}{+4.7}{+9.3} & \ci{+6.4}{+4.1}{+8.8} \\
LongLLMLingua $-$ random, hidden & 25\% & \ci{\ensuremath{-}1.6}{\ensuremath{-}2.6}{\ensuremath{-}0.6} & \ci{\ensuremath{-}3.7}{\ensuremath{-}4.8}{\ensuremath{-}2.6} \\
LongLLMLingua $-$ random, hidden & 50\% & \ci{\ensuremath{-}0.9}{\ensuremath{-}1.9}{+0.2} & \ci{\ensuremath{-}1.4}{\ensuremath{-}2.4}{\ensuremath{-}0.3} \\
typed $-$ \LLMLtwo{}, known & 25\% & \ci{\ensuremath{-}2.3}{\ensuremath{-}4.9}{+0.6} & \ci{\ensuremath{-}1.4}{\ensuremath{-}4.2}{+1.3} \\
\bottomrule
\end{tabular}

%% file: generated/compress_open_gate.tex
\begin{tabular}{lllrrrr}
\toprule
 & & & \multicolumn{2}{c}{sufficiency AUROC vs.} & \multicolumn{2}{c}{hidden acc.\ after 25\% re-read} \\
\cmidrule(lr){4-5}\cmidrule(lr){6-7}
model & policy & budget & evidence lost & answer flipped & lowest sufficiency & random \\
\midrule
\texttt{open-decider} & typed & 25\% & 0.81 & 0.69 & 67 & 66 \\
\texttt{open-decider} & agnostic & 25\% & 0.79 & 0.69 & 68 & 66 \\
\texttt{open-decider} & random & 25\% & 0.81 & 0.68 & 67 & 66 \\
\midrule
\texttt{open-decider} & typed & 50\% & 0.78 & 0.70 & 71 & 69 \\
\texttt{open-decider} & agnostic & 50\% & 0.71 & 0.66 & 71 & 70 \\
\texttt{open-decider} & random & 50\% & 0.70 & 0.67 & 71 & 70 \\
\midrule
\texttt{gpt-6-luna} & typed & 25\% & 0.83 & 0.71 & 74 & 73 \\
\texttt{gpt-6-luna} & agnostic & 25\% & 0.86 & 0.71 & 75 & 73 \\
\texttt{gpt-6-luna} & random & 25\% & 0.84 & 0.72 & 75 & 73 \\
\midrule
\texttt{gpt-6-luna} & typed & 50\% & 0.78 & 0.75 & 82 & 79 \\
\texttt{gpt-6-luna} & agnostic & 50\% & 0.76 & 0.75 & 83 & 80 \\
\texttt{gpt-6-luna} & random & 50\% & 0.74 & 0.74 & 82 & 79 \\
\bottomrule
\end{tabular}

%% file: generated/p4fu_ab_levels.tex
\begin{tabular}{lrrrrrrrrrrrr}
\toprule
 & \multicolumn{4}{c}{MultiRC, \texttt{gpt-6-luna}} & \multicolumn{4}{c}{MultiRC, \texttt{open-decider}} & \multicolumn{4}{c}{FEVER, \texttt{gpt-6-luna}} \\
\cmidrule(lr){2-5}\cmidrule(lr){6-9}\cmidrule(lr){10-13}
budget & \multicolumn{2}{c}{25\%} & \multicolumn{2}{c}{50\%} & \multicolumn{2}{c}{25\%} & \multicolumn{2}{c}{50\%} & \multicolumn{2}{c}{25\%} & \multicolumn{2}{c}{50\%} \\
\cmidrule(lr){2-3}\cmidrule(lr){4-5}\cmidrule(lr){6-7}\cmidrule(lr){8-9}\cmidrule(lr){10-11}\cmidrule(lr){12-13}
policy & known & hidden & known & hidden & known & hidden & known & hidden & known & hidden & known & hidden \\
\midrule
typed & 78.7 & 67.9 & 83.2 & 76.1 & 69.0 & 61.0 & 73.6 & 66.1 & 92.5 & 79.0 & 93.0 & 85.6 \\
agnostic & 68.5 & 67.5 & 76.0 & 75.5 & 63.3 & 62.2 & 66.4 & 66.4 & 62.3 & 61.6 & 76.0 & 76.7 \\
TF-IDF & 76.9 & 67.0 & 82.8 & 74.8 & 65.4 & 61.7 & 68.2 & 66.1 & 88.0 & 77.2 & 90.0 & 86.7 \\
random (3 masks) & 67.3 & 68.3 & 76.2 & 76.0 & 61.1 & 62.2 & 66.7 & 66.9 & 56.8 & 61.2 & 72.2 & 75.2 \\
\LLMLtwo{} & 80.9 & 80.7 & 85.2 & 85.8 & 70.0 & 68.0 & 70.1 & 71.8 & 79.5 & 82.2 & 87.0 & 91.5 \\
LongLLMLingua & 74.6 & 67.6 & 81.2 & 75.2 & 65.7 & 60.1 & 68.2 & 64.1 & 69.5 & 70.6 & 85.5 & 83.1 \\
MMR (embeddings) & 78.4 & 68.3 & 84.1 & 76.1 & 66.6 & 62.3 & 71.0 & 67.1 & 89.0 & 77.3 & 91.5 & 87.2 \\
\bottomrule
\end{tabular}

%% file: generated/p4fu_ab_contrasts.tex
\begin{tabular}{lrlrlrl}
\toprule
 & \multicolumn{2}{c}{MultiRC, \texttt{gpt-6-luna}} & \multicolumn{2}{c}{MultiRC, \texttt{open-decider}} & \multicolumn{2}{c}{FEVER, \texttt{gpt-6-luna}} \\
\cmidrule(lr){2-3}\cmidrule(lr){4-5}\cmidrule(lr){6-7}
policy & policy/random & difference & policy/random & difference & policy/random & difference \\
\midrule
\multicolumn{7}{l}{\emph{hidden questions, 25\% of tokens}} \\
typed & 68.1/68.5 & \ci{\ensuremath{-}0.4}{\ensuremath{-}1.5}{+0.6} & 61.0/62.2 & \ci{\ensuremath{-}1.1}{\ensuremath{-}2.1}{\ensuremath{-}0.2} & 79.0/61.2 & \ci{+17.8}{+15.0}{+20.8} \\
agnostic & 67.6/68.6 & \ci{\ensuremath{-}1.0}{\ensuremath{-}1.8}{\ensuremath{-}0.1} & 62.2/62.2 & \ci{+0.1}{\ensuremath{-}1.0}{+1.2} & 61.6/61.2 & \ci{+0.4}{\ensuremath{-}2.4}{+3.2} \\
TF-IDF & 67.1/68.5 & \ci{\ensuremath{-}1.3}{\ensuremath{-}2.3}{\ensuremath{-}0.4} & 61.7/62.2 & \ci{\ensuremath{-}0.5}{\ensuremath{-}1.5}{+0.5} & 77.2/61.2 & \ci{+16.0}{+13.0}{+19.0} \\
\LLMLtwo{} & 80.7/68.6 & \ci{+12.1}{+11.0}{+13.3} & 68.0/62.2 & \ci{+5.9}{+4.7}{+7.0} & 82.2/61.2 & \ci{+21.0}{+18.6}{+23.2} \\
LongLLMLingua & 67.6/69.2 & \ci{\ensuremath{-}1.6}{\ensuremath{-}2.6}{\ensuremath{-}0.6} & 60.1/62.2 & \ci{\ensuremath{-}2.1}{\ensuremath{-}3.1}{\ensuremath{-}1.1} & 70.6/61.2 & \ci{+9.4}{+6.4}{+12.3} \\
MMR (embeddings) & 68.4/68.5 & \ci{\ensuremath{-}0.1}{\ensuremath{-}1.0}{+0.9} & 62.3/62.2 & \ci{+0.1}{\ensuremath{-}0.9}{+1.1} & 77.3/61.2 & \ci{+16.2}{+13.2}{+19.0} \\
\multicolumn{7}{l}{\emph{hidden questions, 50\% of tokens}} \\
typed & 76.1/76.0 & \ci{+0.1}{\ensuremath{-}1.1}{+1.2} & 66.1/66.9 & \ci{\ensuremath{-}0.8}{\ensuremath{-}1.8}{+0.2} & 85.6/75.2 & \ci{+10.4}{+7.9}{+13.0} \\
agnostic & 75.5/76.0 & \ci{\ensuremath{-}0.5}{\ensuremath{-}1.5}{+0.4} & 66.4/66.9 & \ci{\ensuremath{-}0.5}{\ensuremath{-}1.5}{+0.5} & 76.7/75.2 & \ci{+1.5}{\ensuremath{-}1.5}{+4.8} \\
TF-IDF & 74.8/76.0 & \ci{\ensuremath{-}1.2}{\ensuremath{-}2.2}{\ensuremath{-}0.2} & 66.1/66.9 & \ci{\ensuremath{-}0.8}{\ensuremath{-}1.9}{+0.2} & 86.7/75.2 & \ci{+11.5}{+9.0}{+14.1} \\
\LLMLtwo{} & 85.8/76.0 & \ci{+9.8}{+8.9}{+10.7} & 71.8/66.9 & \ci{+4.9}{+3.9}{+5.9} & 91.5/75.2 & \ci{+16.3}{+14.3}{+18.4} \\
LongLLMLingua & 75.2/76.2 & \ci{\ensuremath{-}0.9}{\ensuremath{-}1.9}{+0.2} & 64.1/66.9 & \ci{\ensuremath{-}2.7}{\ensuremath{-}3.9}{\ensuremath{-}1.6} & 83.1/75.2 & \ci{+8.0}{+4.8}{+10.9} \\
MMR (embeddings) & 76.1/76.0 & \ci{+0.1}{\ensuremath{-}0.9}{+1.1} & 67.1/66.9 & \ci{+0.2}{\ensuremath{-}0.8}{+1.1} & 87.2/75.2 & \ci{+12.0}{+9.6}{+14.4} \\
\bottomrule
\end{tabular}

%% file: generated/p4fu_c.tex
\begin{tabular}{llrrrr}
\toprule
 & & \multicolumn{3}{c}{document label recovered} & \\
\cmidrule(lr){3-5}
corpus & text & supported & refuted & balanced & ``not enough info'' \\
\midrule
FEVER & full & 93.8 & 83.5 & 88.6 & -- \\
 & typed 25\% & 67.2 & 57.3 & 62.2 & 74.6 \\
 & random 25\% & 37.0 & 37.1 & 37.1 & 81.1 \\
 & typed 50\% & 78.5 & 68.3 & 73.4 & 71.7 \\
 & random 50\% & 60.0 & 57.0 & 58.5 & 75.3 \\
Ev. Inference & full & 80.0 & 93.6 & 86.8 & -- \\
 & typed 25\% & 64.2 & 69.1 & 66.6 & 55.5 \\
 & random 25\% & 38.8 & 48.2 & 43.5 & 61.6 \\
 & typed 50\% & 71.0 & 79.1 & 75.1 & 57.7 \\
 & random 50\% & 61.6 & 75.0 & 68.3 & 40.5 \\
\bottomrule
\end{tabular}

%% file: generated/p4fu_c_paired.tex
\begin{tabular}{lllrrrrl}
\toprule
target & corpus & budget & claims & documents & typed & random & typed $-$ random \\
\midrule
document label & FEVER & 25\% & 5433 & 200 & 62.2 & 37.1 & \ci{+25.2}{+19.9}{+30.8} \\
 & FEVER & 50\% & 5433 & 200 & 73.4 & 58.5 & \ci{+14.9}{+10.3}{+19.6} \\
 & Ev. Inference & 25\% & 2915 & 792 & 66.6 & 43.5 & \ci{+23.1}{+20.2}{+25.9} \\
 & Ev. Inference & 50\% & 2915 & 792 & 75.1 & 68.3 & \ci{+6.8}{+4.5}{+9.1} \\
\midrule
annotation-derived support & FEVER & 25\% & 4528 & 200 & 83.3 & 84.3 & \ci{\ensuremath{-}1.0}{\ensuremath{-}3.9}{+2.2} \\
 & FEVER & 50\% & 4464 & 200 & 81.3 & 80.7 & \ci{+0.6}{\ensuremath{-}2.1}{+3.2} \\
 & Ev. Inference & 25\% & 2130 & 764 & 77.8 & 74.4 & \ci{+3.4}{+0.5}{+6.4} \\
 & Ev. Inference & 50\% & 2047 & 740 & 78.6 & 68.4 & \ci{+10.1}{+7.7}{+12.7} \\
\bottomrule
\end{tabular}

%% file: generated/p4_cost_table.tex
\begin{tabular}{l*{10}{r}}
\toprule
 & & \multicolumn{3}{c}{measured, per document} & \multicolumn{6}{c}{modeled, per later question} \\
\cmidrule(lr){3-5}\cmidrule(lr){6-11}
corpus & $\bar k$ & scoring & read 25\% & read full & 25\% & 50\% & full & check sep. & check joint & re-read \\
\midrule
MultiRC & 10.2 & 2,605 & 15,081 & 15,431 & 1,259 & 1,320 & 1,446 & 245 & 177 & 361 \\
FEVER & 27.2 & 2,795 & 10,460 & 10,810 & 311 & 426 & 671 & 281 & 169 & 168 \\
Evidence Inf. & 3.7 & 22,181 & 3,321 & 6,479 & 1,535 & 2,708 & 4,889 & 1,488 & 184 & 1,222 \\
LoCoMo & 149.6 & 180,510 & 64,220 & 96,742 & 6,270 & 11,783 & 22,721 & 6,218 & 146 & 5,680 \\
\midrule
\multicolumn{4}{l}{thousand input tokens per document, typed 25\%} & $k{=}1$ & $k{=}3$ & $k{=}10$ & $k{=}\bar k$ & \multicolumn{2}{r}{ratio at $\bar k$} & break-even $k$ \\
\midrule
MultiRC & \multicolumn{3}{l}{full text} & 1.4 & 4.3 & 14.5 & 14.7 & \multicolumn{2}{r}{1.00} & --- \\
 & \multicolumn{3}{l}{compact once} & 3.9 & 6.4 & 15.2 & 15.4 & \multicolumn{2}{r}{1.05} & 14.0 \\
 & \multicolumn{3}{l}{compact + joint check + re-read} & 4.4 & 8.0 & 20.6 & 20.9 & \multicolumn{2}{r}{1.42} & never \\
 & \multicolumn{3}{l}{compact + separate check + re-read} & 4.5 & 8.2 & 21.3 & 21.6 & \multicolumn{2}{r}{1.47} & never \\
FEVER & \multicolumn{3}{l}{full text} & 0.7 & 2.0 & 6.7 & 18.2 & \multicolumn{2}{r}{1.00} & --- \\
 & \multicolumn{3}{l}{compact once} & 3.1 & 3.7 & 5.9 & 11.2 & \multicolumn{2}{r}{0.62} & 7.8 \\
 & \multicolumn{3}{l}{compact + joint check + re-read} & 3.4 & 4.7 & 9.3 & 20.4 & \multicolumn{2}{r}{1.12} & 119.0 \\
 & \multicolumn{3}{l}{compact + separate check + re-read} & 3.6 & 5.1 & 10.4 & 23.4 & \multicolumn{2}{r}{1.29} & never \\
Evidence Inf. & \multicolumn{3}{l}{full text} & 4.9 & 14.7 & 48.9 & 18.0 & \multicolumn{2}{r}{1.00} & --- \\
 & \multicolumn{3}{l}{compact once} & 23.7 & 26.8 & 37.5 & 27.8 & \multicolumn{2}{r}{1.55} & 6.6 \\
 & \multicolumn{3}{l}{compact + joint check + re-read} & 25.1 & 31.0 & 51.6 & 33.0 & \multicolumn{2}{r}{1.83} & 11.4 \\
 & \multicolumn{3}{l}{compact + separate check + re-read} & 26.4 & 34.9 & 64.6 & 37.8 & \multicolumn{2}{r}{2.10} & 34.5 \\
LoCoMo & \multicolumn{3}{l}{full text} & 22.7 & 68.2 & 227.2 & 3,399.0 & \multicolumn{2}{r}{1.00} & --- \\
 & \multicolumn{3}{l}{compact once} & 186.8 & 199.3 & 243.2 & 1,118.6 & \multicolumn{2}{r}{0.33} & 11.0 \\
 & \multicolumn{3}{l}{compact + joint check + re-read} & 192.6 & 216.8 & 301.5 & 1,990.2 & \multicolumn{2}{r}{0.59} & 17.0 \\
 & \multicolumn{3}{l}{compact + separate check + re-read} & 198.7 & 235.0 & 362.2 & 2,898.5 & \multicolumn{2}{r}{0.85} & 39.7 \\
\bottomrule
\end{tabular}

%% file: generated/p4fu_d.tex
\begin{tabular}{lrrrrrr}
\toprule
 & \multicolumn{3}{c}{\texttt{gpt-6-luna}} & \multicolumn{3}{c}{\texttt{qwen3.8-flash}} \\
\cmidrule(lr){2-4}\cmidrule(lr){5-7}
memory (last checkpoint) & refused & accuracy & answered acc. & refused & accuracy & answered acc. \\
\midrule
empty, original options & 10.0 & 81.4 & 90.6 & 0.0 & 85.1 & 85.1 \\
\midrule
empty & 28.4 & 60.2 & 83.9 & 0.0 & 71.5 & 71.5 \\
full & 0.5 & 90.9 & 91.4 & --- & --- & --- \\
typed 25\% & 1.1 & 78.0 & 78.9 & 0.0 & 75.7 & 75.7 \\
agnostic 25\% & 2.3 & 81.0 & 82.8 & 0.0 & 81.7 & 81.7 \\
random 25\% & 2.8 & 76.7 & 78.8 & 0.0 & 75.4 & 75.4 \\
KEEP & 0.3 & 91.9 & 92.1 & 0.0 & 84.9 & 84.9 \\
DROP & 1.8 & 73.3 & 74.7 & 0.0 & 70.1 & 70.1 \\
\bottomrule
\end{tabular}